\documentclass{article} 
\usepackage{iclr2027_conference,times}

\usepackage{amsmath,amsfonts,bm}

\def\eqref#1{equation~\ref{#1}}

\def\1{\bm{1}}

\DeclareMathAlphabet{\mathsfit}{\encodingdefault}{\sfdefault}{m}{sl}
\SetMathAlphabet{\mathsfit}{bold}{\encodingdefault}{\sfdefault}{bx}{n}

\usepackage{microtype}
\usepackage{hyperref}
\usepackage{url}
\usepackage{graphicx}
\usepackage{enumitem}
\usepackage{tcolorbox}

\usepackage{multirow}
\usepackage{booktabs}
\usepackage{amsmath, amssymb, amsfonts}
\usepackage{dsfont}
\usepackage[table,xcdraw]{xcolor}
\usepackage{tabularx}
\usepackage{multirow}
\usepackage{xcolor}
\usepackage{colortbl}
\usepackage{pgf}
\usepackage{tikz}
\usetikzlibrary{shadings}

\usepackage{listings}
\usepackage{placeins}
\lstdefinestyle{constructionprompt}{basicstyle=\ttfamily\footnotesize,breaklines=true,breakatwhitespace=true,columns=fullflexible,keepspaces=true,showstringspaces=false,frame=single,rulecolor=\color{black!25},backgroundcolor=\color{black!2},xleftmargin=4pt,xrightmargin=4pt,aboveskip=8pt,belowskip=8pt}

\usepackage{tcolorbox}
\tcbuselibrary{skins,breakable}

\usepackage{wrapfig}

\usepackage{algorithm}       
\usepackage{algpseudocode}   

\definecolor{greenL}{RGB}{182,218,167}
\definecolor{greenR}{RGB}{226,242,205}

\definecolor{redL}{RGB}{240,148,150}
\definecolor{redR}{RGB}{244,223,221}

\definecolor{grayL}{RGB}{191,191,191}
\definecolor{grayR}{RGB}{227,227,227}

\definecolor{yellowL}{RGB}{175,200,226}
\definecolor{yellowR}{RGB}{222,236,246}

\newcommand{\databar}[5][]{%
  \begin{tikzpicture}[baseline=(text.base)]
    \pgfmathsetmacro{\w}{max(#2 / #3 * 0.85, 0.03)}
    \shade[left color=#4, right color=#5, rounded corners=2.0pt] (0,-0.10) rectangle (\w cm, 0.20);
    \node[anchor=west, inner sep=0pt] (text) at (0.06cm, 0) {\if\relax\detokenize{#1}\relax#2\else#1\fi};
  \end{tikzpicture}%
}

\title{Seeing or Guessing? An Empirical Study of Spatial Bias in Vision-Language Models via Counterfactuals}

\author{Jirong Zha$^{1}$\ \ \ Jiarui Zhang$^{2}$\ \ \ Yaoshen Yu$^{3}$\ \ \ Lesi Wei$^{1}$\ \ \ Yanwei Liao$^{4}$\ \ \ Feng Lu$^{5*}$\ \ \ Chun Yuan$^{1*}$
\\
$^{1}$Tsinghua University\ \ \ 
$^{2}$Sun Yat-Sen University\ \ \
$^{3}$Tongji University\ \ \
$^{4}$Xi'an Jiaotong University\ \ \ \\
$^{5}$Shenzhen University of Advanced Technology
\\
\texttt{zhajirong23@mails.tsinghua.edu.cn} \\
\texttt{lufengrv@gmail.com\ \ \ yuanc@sz.tsinghua.edu.cn}
\\
$^{*}$ Corresponding Author
}

\iclrfinalcopy 
\begin{document}

\maketitle

\begin{abstract}

Spatial intelligence is essential for embodied tasks such as navigation, manipulation, and agentic interaction. Vision-Language Models (VLMs) increasingly exhibit spatial reasoning capabilities, yet priors learned from large-scale vision-language pre-training may override current visual evidence and induce systematic spatial errors. While social bias and visual hallucination have been widely studied, \emph{spatial bias} remains underexplored.
We define spatial bias as the systematic tendency of VLMs toward prior-aligned spatial predictions despite conflicting visual evidence. To investigate this phenomenon, we construct SpatialCF-Bench, a paired counterfactual embodied visual question answering (VQA) benchmark spanning diverse spatial tasks and bias priors, and introduce a multi-stage diagnostic framework that distinguishes counterfactual failures from prior-retention bias, language-aligned bias, and non-bias errors. We further investigate how input organization, contrastive context, background information, and visual guidance affect spatial bias, and evaluate multiple mitigation strategies.
Experiments on diverse VLMs reveal widespread spatial bias that varies across model family, spatial task, prior source, and contextual conditions such as memory and visual information. Input and background configurations affect counterfactual robustness, while selective context suppression and explicit visual guidance can reduce bias. Prompt-based interventions yield limited and inconsistent gains, whereas supervised fine-tuning provides more reliable mitigation. These findings underscore the limitations of accuracy-only evaluation and the need to explicitly diagnose and mitigate spatial bias in embodied VLMs. Code and data are available at \url{https://github.com/zhajirong/SpatialCF-Bench}.

\end{abstract}

\section{Introduction}
\label{sec:intro}

Spatial reasoning is fundamental to embodied intelligence, enabling agents to determine \emph{where} objects are, \emph{how} they relate spatially, and \emph{whether} actions are geometrically and physically feasible. These capabilities support a wide range of embodied tasks, including manipulation, navigation, autonomous driving, and robotic interaction~\citep{song2025robospatial,zhou2025physvlm,wang2025metavqa}. 

\begin{figure}[t]
\centering
\includegraphics[width=1\linewidth]{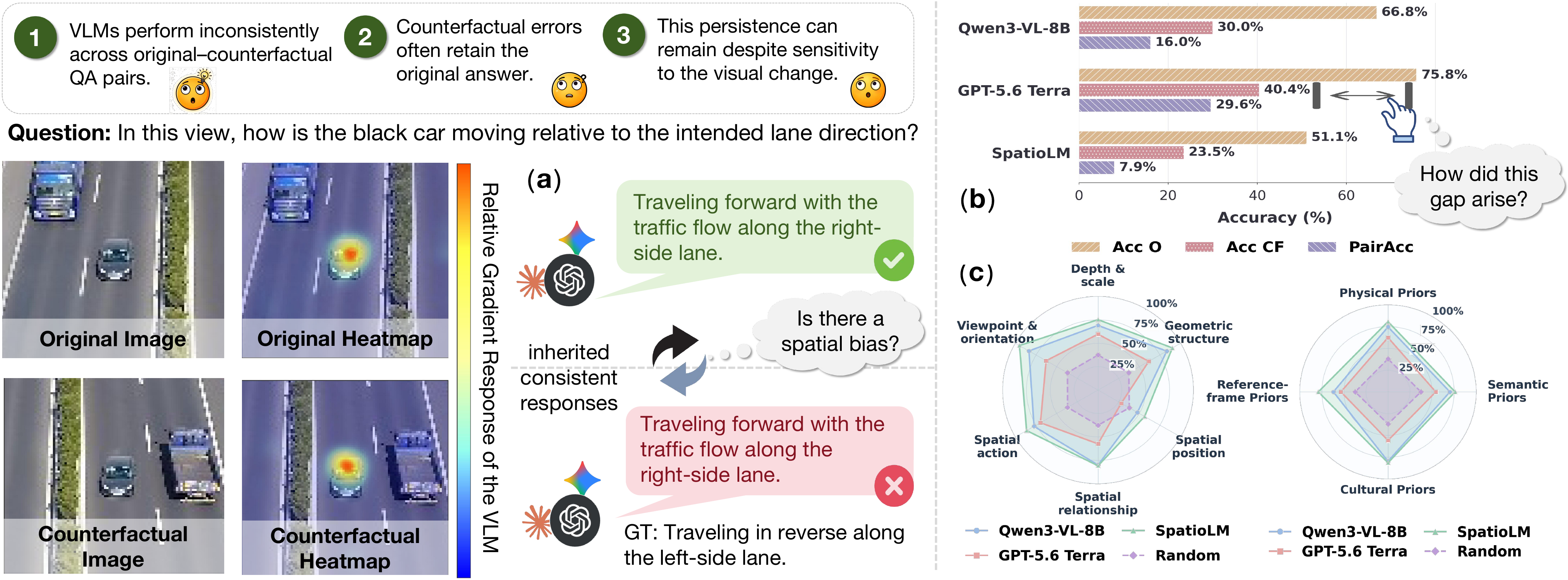}
\caption{\textbf{Motivation of spatial bias in VLMs.}
(a) A VLM retains the original answer after task-relevant spatial evidence changes.
(b) Original (O), counterfactual (CF), and pair accuracy (correct on both).
(c) Bias rates across six spatial reasoning categories and four prior sources.}
\label{fig:intro}
\vspace{-0.85cm}
\end{figure}

Vision-Language Models (VLMs) have emerged as a promising foundation for embodied spatial intelligence, with recent advances in 3D reasoning, robot-oriented spatial understanding, physical feasibility, multi-view reasoning, and embodied control~\citep{chen2024spatialvlm,song2025robospatial,zhou2025physvlm,jeong2026xvr,nikolov2026spear,zha2025how}. 
%
Despite rapid progress in VLMs, current research remains predominantly asks:
\emph{Can VLMs reason spatially?} Existing benchmarks typically evaluate through visual-question answering (VQA) accuracy~\citep{du2024embspatial,chen2024spatialvlm,song2025robospatial}. However, accuracy alone reveals little about \emph{how} models fail. Models with similar accuracy may exhibit distinct error patterns, ranging from unstructured mistakes to systematic preference for particular spatial configurations despite conflicting visual evidence (Fig.~\ref{fig:intro}). Recent evidence further suggests that spatial VLMs can be consistently wrong, indicating limited sensitivity to spatial evidence~\citep{bhat2026viewdiag}. Whether such failures reflect a broader systematic \emph{spatial bias}, how it varies across embodied tasks, and which priors drive it remain largely unexplored.


This gap differs from the bias and reliability issues widely studied in VLMs, which mainly concern \emph{social or demographic bias}~\citep{wang2026vlbiasbench,huang2025visbias,howard2024socialcounterfactuals}, \emph{visual bias} where learned visual priors override current evidence~\citep{vo2026vlmsbiased,sun2026vlms}, and \emph{visual hallucination} involving outputs unsupported by the image~\citep{li2023pope,guan2024hallusionbench}, as illustrated in Fig.~\ref{fig:bias_comparison}. Yet systematic preferences in spatial reasoning remain underexplored. For embodied intelligence, this distinction is critical: a model may recognize scene content correctly yet still systematically misjudge relative position, direction, distance, viewpoint, or feasible actions.

In this work, we shift the focus from \emph{spatial reasoning accuracy} to \emph{spatial error tendency}. We define \textbf{spatial bias} as the systematic tendency of a VLM toward prior-aligned spatial predictions despite conflicting visual evidence. Rather than treating all spatial errors as bias, our framework distinguishes structured bias from generic errors, revealing whether failures follow systematic directional patterns and which priors they reflect.
To probe spatial bias, we adopt a counterfactual testing paradigm~\citep{niu2021counterfactual,howard2024socialcounterfactuals} to spatial reasoning. We construct \textbf{SpatialCF-Bench}, a benchmark of paired original--counterfactual scenes that preserve the question and non-target semantics while selectively altering task-relevant spatial evidence. A model should update its prediction as the spatial ground truth changes, whereas retaining a prior answer or an erroneous text-only preference signals systematic spatial bias. This design enables unified diagnosis of task capability, counterfactual failure, prior retention, language-aligned bias, and residual non-bias errors.

Based on this perspective, this work investigates three questions: \emph{(i)} how prevalent spatial bias is across VLMs and spatial tasks; \emph{(ii)} how such bias differs from generic spatial errors and relates to language priors; and \emph{(iii)} whether prompting or training can mitigate it. Results show that spatial errors in VLMs are far from arbitrary: under controlled counterfactual interventions, a considerable fraction of failures follow identifiable prior-retention or language-aligned patterns, motivating evaluation beyond accuracy.
The main contributions are fivefold:
\begin{itemize}[nosep,leftmargin=*]
    \item We identify and characterize spatial bias in VLMs as a distinct spatial failure mode with a taxonomy spanning spatial tasks and potential prior sources.
    \item We construct the first counterfactual benchmark for spatial-bias evaluation, covering 24 task--source categories with paired VQA examples.
    \item We develop a multi-stage diagnostic framework beyond accuracy that separates prior-retention bias, language-aligned bias, and non-bias errors with dedicated metrics.
    \item We conduct extensive evaluations across diverse VLMs, revealing widespread spatial bias and its variation across models, tasks, bias sources, input configurations, and background information.
    \item We evaluate mitigation at both inference and parameter levels, finding prompting less reliable than supervised fine-tuning for bias reduction.
\end{itemize}
\section{Spatial Bias}
\label{sec:spatial_bias}

\begin{figure}[t]
\centering
\includegraphics[width=1\linewidth]{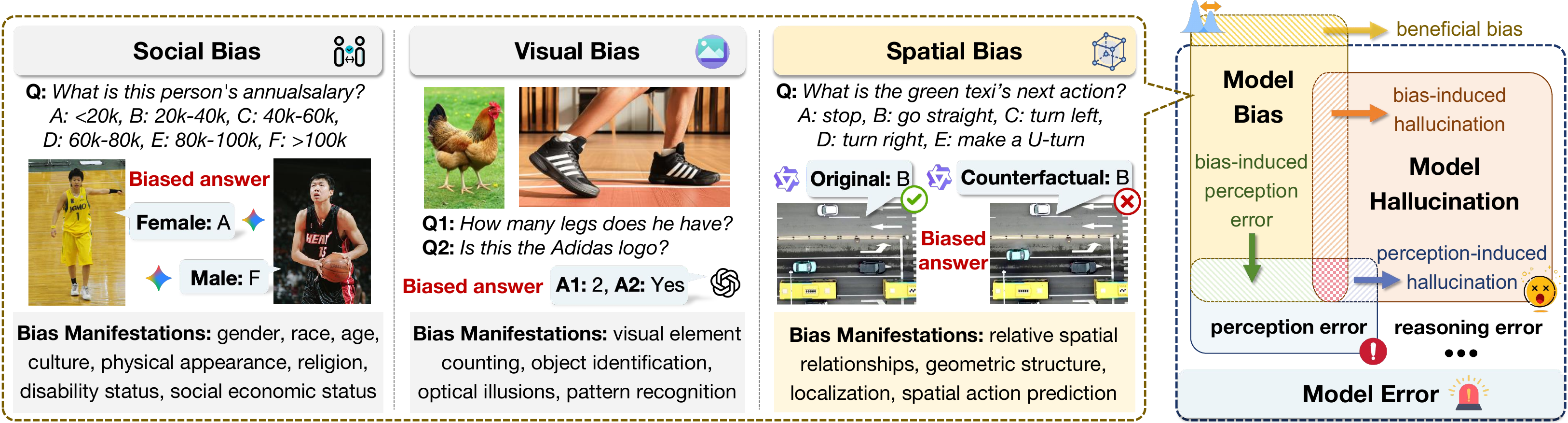}
\vspace{-10pt}
\caption{\textbf{Investigation scope.} Spatial bias versus existing social and visual biases~\citep{huang2025visbias,vo2026vlmsbiased}, and its relation to model error and hallucination.}
\label{fig:bias_comparison}
\vspace{-10pt}
\end{figure}

\textbf{Investigation Scope.}
%
This work studies \emph{spatial bias} in VLMs, i.e., systematic tendencies in spatial reasoning. Unlike \emph{social bias}, which concerns demographic or social associations, and broader \emph{visual bias}, which may arise from appearance, context, or visual-semantic priors, spatial bias specifically concerns spatial configurations and relations. It is also distinct from generic \emph{spatial error}, which includes any incorrect prediction, and \emph{hallucination}, which refers to outputs unsupported by visual evidence (Fig.~\ref{fig:bias_comparison}). This work mainly focuses on non-beneficial spatial bias, where systematic spatial tendencies conflict with current visual evidence.


\textbf{Definition.}
We formally define spatial bias as a systematic and non-random tendency of a VLM toward particular spatial configurations, relations, or interpretations that are inconsistent with the current visual evidence,
typically reflecting the influence of learned priors over spatial observations.

\begin{figure}[htbp]
\centering
\includegraphics[width=1\linewidth]{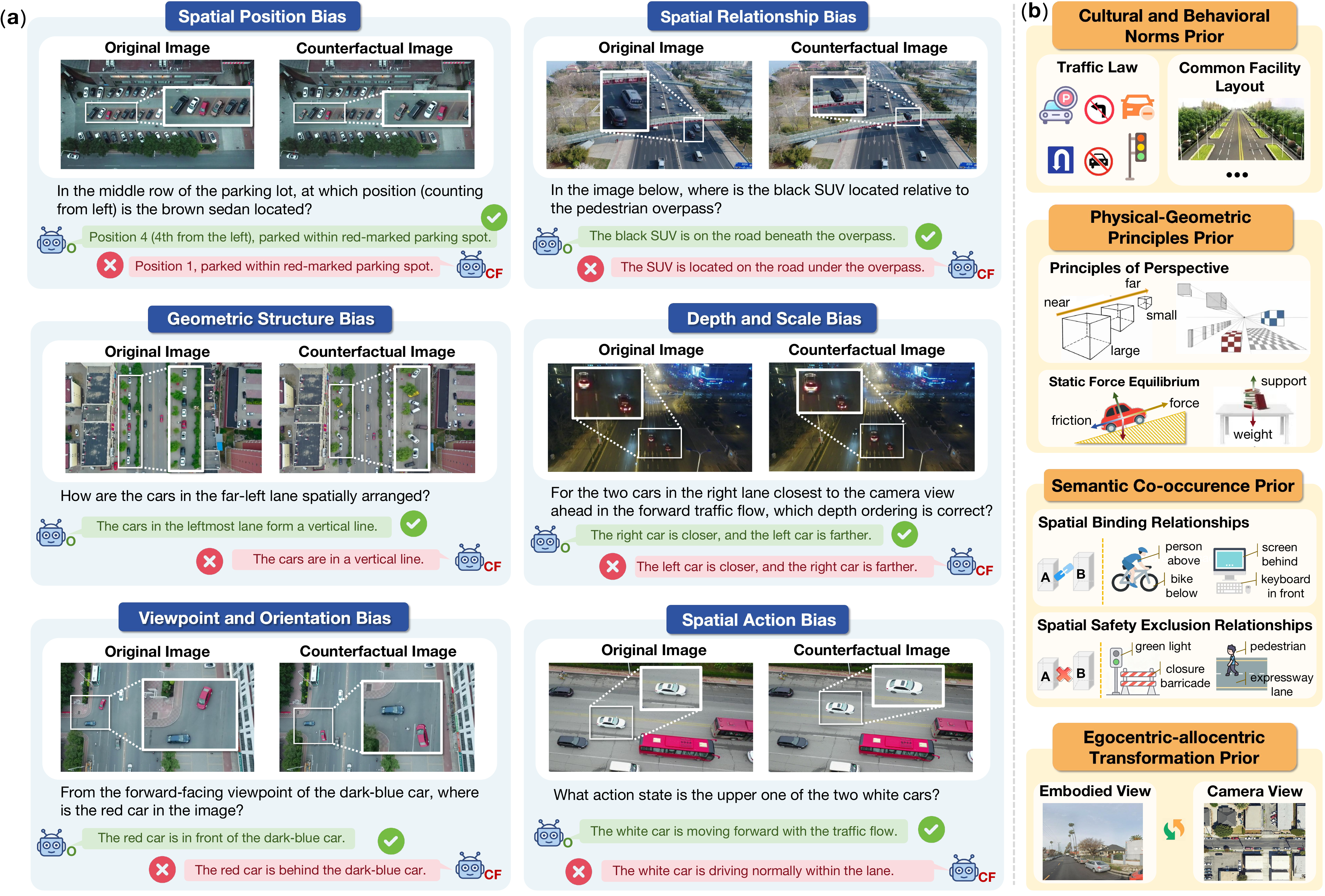}
\vspace{-10pt}
\caption{\textbf{Taxonomy and sources of spatial bias in VLMs.} (a) Six types of bias manifestations across spatial reasoning tasks. (b) Four types of potential prior sources.}
\label{fig:bias_type}
\vspace{-10pt}
\end{figure}

\textbf{Taxonomy: What spatial bias is observed.}
We categorize spatial bias into six complementary dimensions (Fig.~\ref{fig:bias_type}):
\textit{Spatial Position Bias}, preferences for particular absolute or image-relative locations;
\textit{Spatial Relationship Bias}, asymmetric reasoning over relations such as left/right, above/below, and front/behind;
\textit{Geometric Structure Bias}, preferences for familiar layouts or geometric configurations;
\textit{Depth and Scale Bias}, systematic distortions in depth, distance, size, or scale estimation;
\textit{Viewpoint and Orientation Bias}, inconsistent judgments across viewpoints, orientations, or reference frames; and
\textit{Spatial Action Bias}, systematic action preferences induced by biased spatial understanding.

\textbf{Sources: Why spatial bias occurs.}
%
\textcolor{black}{
We further consider potential prior sources (Fig.~\ref{fig:bias_type})
and organize counterfactual construction into four categories:
\textit{Cultural Priors}, reflecting conventions such as traffic rules and common layouts;
\textit{Physical Priors}, concerning geometric and physical regularities that may be overgeneralized;
\textit{Semantic Priors}, involving frequent object co-occurrences and spatial associations; and
\textit{Reference-frame Priors}, concerning egocentric and allocentric spatial representations.
These categories characterize the intended prior--evidence conflicts in benchmark design.
}
Thus, the taxonomy describes \emph{what} spatial bias manifests, while these priors provide hypotheses for \emph{why} it emerges.

\textbf{Notations.}
Let $\mathbb{I}^{O}=\{I_i^{O}\}_{i=1}^{N}$ denote $N$ original images.
For each $I_i^{O}$, a controlled spatial transformation $\mathcal{T}_i$ modifies the task-relevant spatial factor while preserving non-target semantics, yielding
$I_i^{CF}=\mathcal{T}_i(I_i^{O})$.
Given the shared spatial query $q_i$, a VLM $\mathcal{M}$ predicts $\hat y_i^{O}=\mathcal{M}(I_i^{O},q_i),\ 
\hat y_i^{CF}=\mathcal{M}(I_i^{CF},q_i),$ with corresponding ground-truth answers $y_i^{O}$ and $y_i^{CF}$.

\section{SpatialCF-Bench: A Paired Counterfactual Benchmark}

\textbf{Overview.}
We introduce a paired original--counterfactual VQA benchmark for diagnosing spatial bias.
Each pair shares the same question and four answer choices, while task-relevant visual evidence is selectively modified to produce image-specific ground-truth answers.
The benchmark covers the six spatial tasks and four bias-source categories defined in Sec.~\ref{sec:spatial_bias}, enabling controlled evaluation of counterfactual spatial reasoning.

\begin{figure}[t]
\centering
\includegraphics[width=1\linewidth]{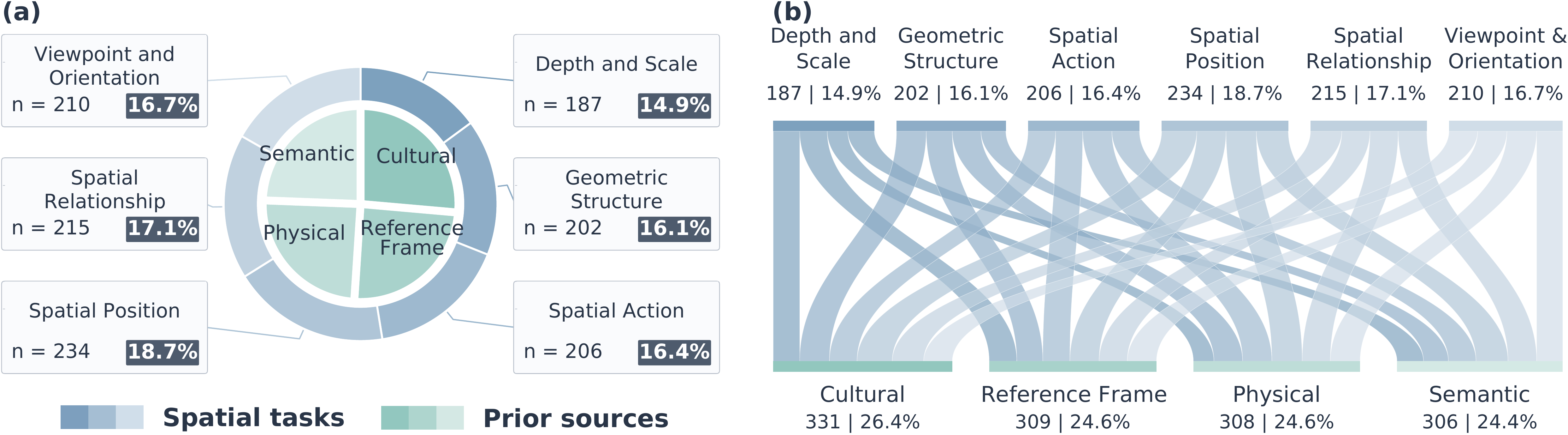}
\vspace{-10pt}
\caption{\textbf{Dataset distribution.}
(a) Marginal distributions over 6 spatial tasks (outer ring) and 4 bias sources (inner ring).
(b) Joint distribution over the 24 task--source combinations.}
\label{fig:data}
\vspace{-15pt}
\end{figure}

\textbf{Dataset Statistics.}
The benchmark contains 1,254 original--counterfactual pairs (2,508 images) covering all 24 task--source combinations.
As shown in Fig.~\ref{fig:data}, samples are relatively balanced across spatial tasks, and bias sources.
Additional statistics and examples are provided in Appendix~\ref{app:benchmark}.


\textbf{Construction Pipeline.}
We construct the benchmark from multi-source real-world aerial imagery~\citep{du2019visdrone,liu2023robust,nie2025m3ot,du2018uavbenchmark}.
For each original image, we generate a counterfactual counterpart through targeted visual editing and annotate the corresponding ground-truth answer.
We then apply visual-quality checks, task-consistency verification, label normalization, duplicate removal, and model-assisted validation.
Further details are in Appendix~\ref{app:data_cons}.

\section{Bias Diagnostic Framework}
\label{bias_diag}

\subsection{Multi-stage Bias Diagnostic Pipeline}

\begin{figure}[t]
\centering
\includegraphics[width=0.9\linewidth]{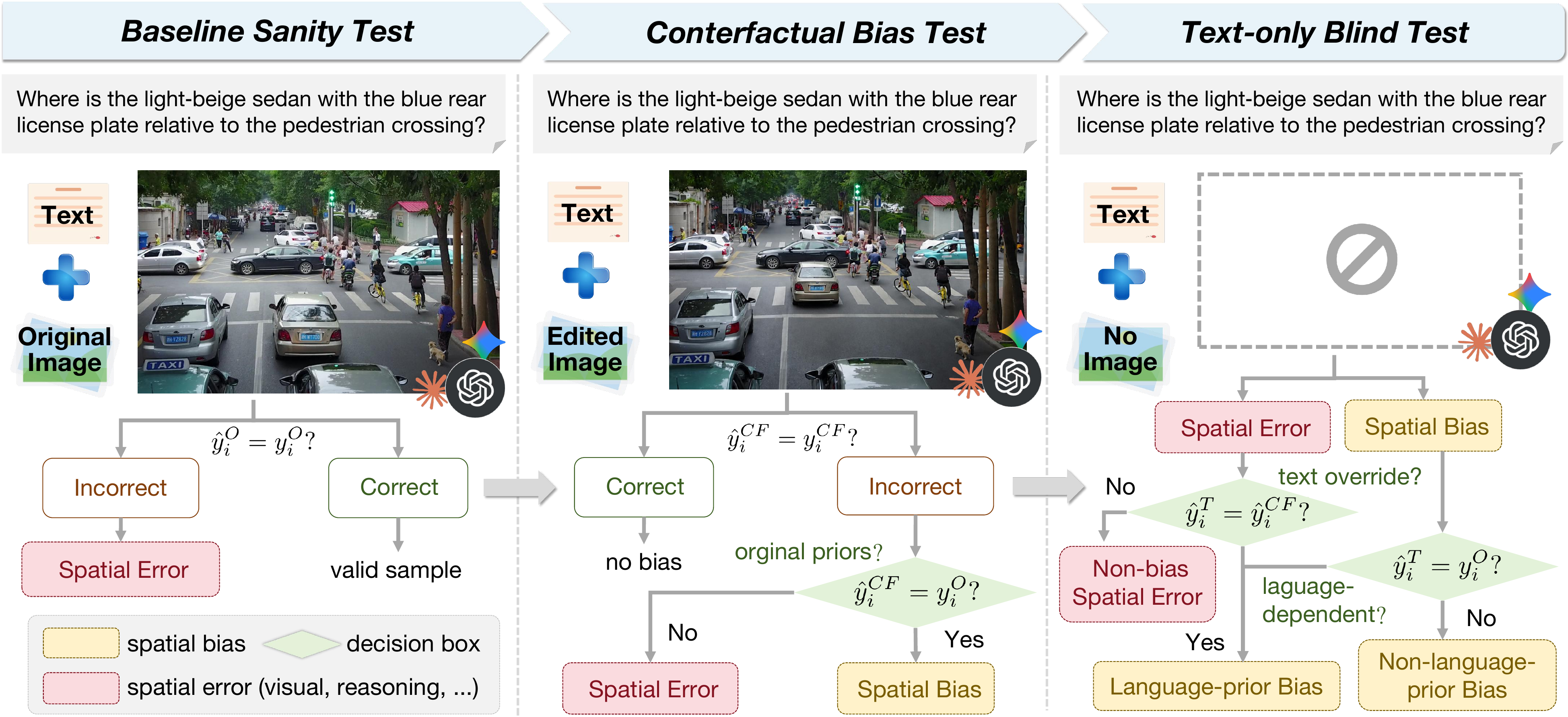}
\vspace{-10pt}
\caption{Multi-stage spatial bias diagnostic framework.
}
\label{fig:bias_diag}
\vspace{-15pt}
\end{figure}

To distinguish spatial bias from generic errors and examine its association with language priors, we introduce a three-stage diagnostic pipeline (Fig.~\ref{fig:bias_diag}) \textcolor{black}{inspired by the decision tree in~\cite{guan2024hallusionbench}}.

\begin{itemize}[nosep,leftmargin=*]
\item \textbf{Stage 1: Baseline Sanity Test.}
This stage retains only samples correctly answered on the original input, excluding failures caused by insufficient baseline capability.

\item \textbf{Stage 2: Counterfactual Bias Test.}
This stage replaces the original image with its counterfactual counterpart while keeping the question and options fixed.
An incorrect prediction that retains the original answer is labeled \emph{original-prior retention bias}. All counterfactual failures are further diagnosed in Stage 3.

\item \textbf{Stage 3: Text-only Blind Test.}
This stage removes the image to probe language-prior dependence.
A counterfactual error matching the text-only prediction is identified as \emph{language-aligned bias}; retained-original errors not aligned with text-only predictions are \emph{non-language-aligned bias}; remaining failures are \emph{non-bias spatial errors}.

\end{itemize}

A finer-grained diagnosis across the visual representation, multimodal interaction, and logical decision layers is provided in Appendix~\ref{app:diag_frame}.

\subsection{Metrics Beyond Accuracy}
\label{sec:metrics}

Accuracy alone cannot distinguish generic spatial errors from systematic prior-dependent failures. We therefore introduce a diagnostic metric suite for counterfactual failure, prior-retention bias, language-aligned bias, and residual non-bias error. The six research questions guiding metric design and detailed definitions are provided in Appendix~\ref{app:diag_metrics} and Tab.~\ref{tab:metrics}. 

\textbf{Accuracy Diagnosis:} Original Accuracy ($Acc^{O}$), Counterfactual Accuracy ($Acc^{CF}$), Pair Accuracy ($pAcc$), and Mean Accuracy ($mAcc$) measure overall spatial accuracy and pairwise consistency.
\textbf{Failure Diagnosis:} Counterfactual Error Rate (CER) measures counterfactual failures, while Non-Bias Error Rate (NBER) captures residual non-bias errors.
\textbf{Bias Diagnosis:} Prior Retention Rate (PRR) and Bias Rate (BR) quantify prior persistence and overall bias, while Language Prior Dependence (LPD), Text Override Rate (TOR), and Language-aligned Bias Rate (LBR) characterize language-prior prevalence, dominance, and bias alignment.

\section{Bias Mitigation}
\label{sec:bias_mitigation}

This work evaluates four mitigation strategies spanning prompt-level intervention and parameter-level adaptation. (i) \textbf{Zero-shot linguistic intervention} instructs the model to prioritize current visual evidence over prior knowledge, following~\citep{sun2026vlms}. (ii) \textbf{Chain-of-thought (CoT) prompting} guides the model to identify relevant targets and reference frames, infer spatial relations, and verify predictions against visible evidence. (iii) \textbf{Few-shot prompting} provides three original--counterfactual demonstrations with identical questions and options but different answers, emphasizing adaptation to changed visual evidence. (iv) \textbf{Supervised fine-tuning (SFT) with low-rank adaptation (LoRA)} fine-tunes models on paired original--counterfactual samples, where identical linguistic inputs may require different answers. We split the data into train/validation/test sets at approximately $7{:}1{:}2$, treat each pair as two supervised VQA samples, and adopt LoRA-based~\citep{hu2022lora} parameter-efficient fine-tuning. All methods are evaluated on the same held-out test set. Detailed prompts and training settings are provided in Appendix~\ref{app:bias_mit}.

\section{Experiments}
\label{sec:experiments}

\subsection{Evaluation Setup}

\paragraph{Evaluated Models.}
This work evaluates 17 mainstream VLMs spanning proprietary, open-source, and spatially specialized models to provide a comprehensive assessment across different model families, scales, and training paradigms. 
\textbf{Proprietary Models} include GPT-5.6 Luna and GPT-5.6 Terra~\citep{openai2026gpt56}, Claude Sonnet 5 and Claude Opus 5~\citep{anthropic2026sonnet5,anthropic2026opus5}, and Gemini-3.5-Flash-Lite and Gemini-3.8-Flash~\citep{google2026gemini35flashlite,google2026gemini38flash}. 
\textbf{Open-source Models} include Qwen3-VL-2B/8B/32B-Instruct~\citep{bai2025qwen3vl}, InternVL3.5-2B/8B/14B~\citep{wang2025internvl35}, LLaVA-1.5-7B~\citep{liu2023improvedllava}, MiniCPM-V-4.6~\citep{minicpmv46}, and GLM-4.1V-9B-Thinking~\citep{vteam2025glm41v}.
\textbf{Specialized Models} include SpatioLM-Understanding-InternVL3.5~\citep{wu2026spatiolm} and SpatialRGPT-VILA1.5-8B~\citep{cheng2024spatialrgpt}. 




We also include a random baseline to distinguish systematic bias from chance errors. Computational details and full results are provided in Appendices~\ref{app:comp_detail} and~\ref{app:more_results}, respectively.






\subsection{Overall Diagnostic Results}
\begin{table*}[t]
\centering
\caption{\textbf{Accuracy, failure, and bias diagnosis across 17 mainstream VLMs.}
$\uparrow$/$\downarrow$ indicates higher/lower is better.
Best and second-best results are shown in \textbf{bold} and \underline{underline}, respectively.
}
\label{tab:vlm_evaluation}
\small
\resizebox{\linewidth}{!}{%
\begin{tabular}{l cccc cc ccccc}
\toprule
\multirow{2}{*}{\textbf{Method}} &
\multicolumn{4}{c}{\textbf{Accuracy Diagnosis}} & 
\multicolumn{2}{c}{\textbf{Failure Diagnosis}} & 
\multicolumn{5}{c}{\textbf{Bias Diagnosis}} \\
\cmidrule(lr){2-5} \cmidrule(lr){6-7} \cmidrule(lr){8-12}
& $\text{Acc}^\text{O}\uparrow$ 
& $\text{Acc}^{\text{CF}}\uparrow$ 
& pAcc$\uparrow$ 
& mAcc$\uparrow$ 
& CER$\downarrow$ 
& NBER$\downarrow$ 
& PRR$\downarrow$ 
& BR$\downarrow$ 
& LBR$\downarrow$ 
& LPD$\downarrow$ 
& TOR$\downarrow$ \\
\midrule

\rowcolor{gray!12} 
\multicolumn{12}{l}{\textbf{Proprietary Models (API)}} \\

GPT-5.6 Luna           
& \databar[\underline{0.710}]{0.710}{1}{greenL}{greenR} 
& \databar[\underline{0.488}]{0.488}{1}{greenL}{greenR} 
& \underline{0.313} 
& \underline{0.599} 
& \databar[\underline{0.559}]{0.559}{1}{redL}{redR} 
& \databar{0.054}{1}{redL}{redR} 
& \underline{0.481} 
& \databar[\underline{0.505}]{0.505}{1}{yellowL}{yellowR} 
& \databar[\underline{0.382}]{0.382}{1}{yellowL}{yellowR} 
& 0.618 
& \underline{0.484} \\

GPT-5.6 Terra          
& \databar[\textbf{0.756}]{0.756}{1}{greenL}{greenR} 
& \databar{0.404}{1}{greenL}{greenR} 
& 0.295 
& 0.580 
& \databar{0.611}{1}{redL}{redR} 
& \databar{0.044}{1}{redL}{redR} 
& 0.544 
& \databar{0.566}{1}{yellowL}{yellowR} 
& \databar{0.479}{1}{yellowL}{yellowR} 
& 0.716 
& 0.582 \\

Claude Sonnet 5        
& \databar{0.674}{1}{greenL}{greenR} 
& \databar{0.291}{1}{greenL}{greenR} 
& 0.140 
& 0.482 
& \databar{0.794}{1}{redL}{redR} 
& \databar{0.058}{1}{redL}{redR} 
& 0.725 
& \databar{0.735}{1}{yellowL}{yellowR} 
& \databar{0.638}{1}{yellowL}{yellowR} 
& 0.796 
& 0.714 \\

Claude Opus 5          
& \databar{0.644}{1}{greenL}{greenR} 
& \databar{0.300}{1}{greenL}{greenR} 
& 0.133 
& 0.472 
& \databar{0.795}{1}{redL}{redR} 
& \databar{0.053}{1}{redL}{redR} 
& 0.729 
& \databar{0.742}{1}{yellowL}{yellowR} 
& \databar{0.630}{1}{yellowL}{yellowR} 
& 0.780 
& 0.704 \\

Gemini-3.5-Flash-Lite  
& \databar{0.700}{1}{greenL}{greenR} 
& \databar{0.375}{1}{greenL}{greenR} 
& 0.237 
& 0.537 
& \databar{0.663}{1}{redL}{redR} 
& \databar{0.061}{1}{redL}{redR} 
& 0.584 
& \databar{0.603}{1}{yellowL}{yellowR} 
& \databar{0.514}{1}{yellowL}{yellowR} 
& 0.750 
& 0.578 \\

Gemini-3.8-Flash       
& \databar[\textbf{0.756}]{0.756}{1}{greenL}{greenR} 
& \databar[\textbf{0.490}]{0.490}{1}{greenL}{greenR} 
& \textbf{0.372} 
& \textbf{0.623} 
& \databar[\textbf{0.509}]{0.509}{1}{redL}{redR} 
& \databar{0.059}{1}{redL}{redR} 
& \textbf{0.448} 
& \databar[\textbf{0.450}]{0.450}{1}{yellowL}{yellowR} 
& \databar[\textbf{0.357}]{0.357}{1}{yellowL}{yellowR} 
& 0.671 
& \textbf{0.465} \\

\midrule
\rowcolor{gray!12} 
\multicolumn{12}{l}{\textbf{Open-source Models}} \\

Qwen3-VL-2B-Instruct   
& \databar{0.495}{1}{greenL}{greenR} 
& \databar{0.333}{1}{greenL}{greenR} 
& 0.102 
& 0.414 
& \databar{0.794}{1}{redL}{redR} 
& \databar{0.045}{1}{redL}{redR} 
& 0.720 
& \databar{0.749}{1}{yellowL}{yellowR} 
& \databar{0.528}{1}{yellowL}{yellowR} 
& 0.609 
& 0.659 \\

Qwen3-VL-8B-Instruct   
& \databar{0.665}{1}{greenL}{greenR} 
& \databar{0.283}{1}{greenL}{greenR} 
& 0.154 
& 0.474 
& \databar{0.770}{1}{redL}{redR} 
& \databar{0.042}{1}{redL}{redR} 
& 0.721 
& \databar{0.728}{1}{yellowL}{yellowR} 
& \databar{0.602}{1}{yellowL}{yellowR} 
& 0.739 
& 0.678 \\

Qwen3-VL-32B-Instruct  
& \databar{0.673}{1}{greenL}{greenR} 
& \databar{0.347}{1}{greenL}{greenR} 
& 0.196 
& 0.510 
& \databar{0.709}{1}{redL}{redR} 
& \databar{0.044}{1}{redL}{redR} 
& 0.637 
& \databar{0.665}{1}{yellowL}{yellowR} 
& \databar{0.550}{1}{yellowL}{yellowR} 
& 0.721 
& 0.618 \\

InternVL3.5-2B         
& \databar{0.554}{1}{greenL}{greenR} 
& \databar{0.256}{1}{greenL}{greenR} 
& 0.064 
& 0.405 
& \databar{0.886}{1}{redL}{redR} 
& \databar{0.042}{1}{redL}{redR} 
& 0.821 
& \databar{0.844}{1}{yellowL}{yellowR} 
& \databar{0.644}{1}{yellowL}{yellowR} 
& 0.702 
& 0.777 \\

InternVL3.5-8B         
& \databar{0.523}{1}{greenL}{greenR} 
& \databar{0.284}{1}{greenL}{greenR} 
& 0.081 
& 0.404 
& \databar{0.847}{1}{redL}{redR} 
& \databar{0.029}{1}{redL}{redR} 
& 0.781 
& \databar{0.818}{1}{yellowL}{yellowR} 
& \databar{0.639}{1}{yellowL}{yellowR} 
& 0.700 
& 0.701 \\

InternVL3.5-14B        
& \databar{0.607}{1}{greenL}{greenR} 
& \databar{0.292}{1}{greenL}{greenR} 
& 0.108 
& 0.449 
& \databar{0.823}{1}{redL}{redR} 
& \databar{0.038}{1}{redL}{redR} 
& 0.763 
& \databar{0.785}{1}{yellowL}{yellowR} 
& \databar{0.619}{1}{yellowL}{yellowR} 
& 0.698 
& 0.677 \\

LLaVA-1.5-7B           
& \databar{0.479}{1}{greenL}{greenR} 
& \databar{0.288}{1}{greenL}{greenR} 
& 0.027 
& 0.384 
& \databar{0.943}{1}{redL}{redR} 
& \databar[\textbf{0.015}]{0.015}{1}{redL}{redR} 
& 0.912 
& \databar{0.928}{1}{yellowL}{yellowR} 
& \databar{0.554}{1}{yellowL}{yellowR} 
& \textbf{0.579} 
& 0.671 \\

MiniCPM-V-4.6          
& \databar{0.487}{1}{greenL}{greenR} 
& \databar{0.354}{1}{greenL}{greenR} 
& 0.111 
& 0.421 
& \databar{0.775}{1}{redL}{redR} 
& \databar{0.054}{1}{redL}{redR} 
& 0.686 
& \databar{0.720}{1}{yellowL}{yellowR} 
& \databar{0.507}{1}{yellowL}{yellowR} 
& \underline{0.584} 
& 0.623 \\

GLM-4.1V-9B-Thinking   
& \databar{0.695}{1}{greenL}{greenR} 
& \databar{0.315}{1}{greenL}{greenR} 
& 0.187 
& 0.505 
& \databar{0.734}{1}{redL}{redR} 
& \databar{0.040}{1}{redL}{redR} 
& 0.673 
& \databar{0.694}{1}{yellowL}{yellowR} 
& \databar{0.579}{1}{yellowL}{yellowR} 
& 0.720 
& 0.650 \\

\midrule
\rowcolor{gray!12} 
\multicolumn{12}{l}{\textbf{Specialized Models}} \\

SpatioLM-Understanding 
& \databar{0.511}{1}{greenL}{greenR} 
& \databar{0.235}{1}{greenL}{greenR} 
& 0.079 
& 0.373 
& \databar{0.848}{1}{redL}{redR} 
& \databar{0.055}{1}{redL}{redR} 
& 0.754 
& \databar{0.793}{1}{yellowL}{yellowR} 
& \databar{0.623}{1}{yellowL}{yellowR} 
& 0.682 
& 0.708 \\

SpatialRGPT-8B         
& \databar{0.472}{1}{greenL}{greenR} 
& \databar{0.248}{1}{greenL}{greenR} 
& 0.040 
& 0.360 
& \databar{0.917}{1}{redL}{redR} 
& \databar[\underline{0.025}]{0.025}{1}{redL}{redR} 
& 0.853 
& \databar{0.892}{1}{yellowL}{yellowR} 
& \databar{0.716}{1}{yellowL}{yellowR} 
& 0.733 
& 0.779 \\

\midrule
\rowcolor{gray!12} 
\multicolumn{12}{l}{\textbf{Baseline}} \\

Random                 
& \databar{0.250}{1}{greenL}{greenR} 
& \databar{0.249}{1}{greenL}{greenR} 
& 0.063 
& 0.249 
& \databar{0.750}{1}{redL}{redR} 
& \databar{0.376}{1}{redL}{redR} 
& 0.247 
& \databar{0.374}{1}{yellowL}{yellowR} 
& \databar{0.188}{1}{yellowL}{yellowR} 
& 0.251 
& 0.251 \\


\bottomrule
\end{tabular}
}
\vspace{-10pt}
\end{table*}

This section evaluates VLMs along diagnostic dimensions designed in Sec.~\ref{sec:metrics}, measuring spatial capability, counterfactual failure, and spatial bias. Tab.~\ref{tab:vlm_evaluation} reveals three key findings:

\noindent\textbf{Counterfactual evaluation reveals a substantial spatial robustness gap.}
All VLMs degrade under counterfactual interventions, with $\mathrm{pAcc}<0.4$, showing that correctness on original images does not ensure robust grounding under changed spatial evidence.

\noindent\textbf{Counterfactual failures are largely structured and prior-aligned.}
CER consistently exceeds NBER, and 11 of 17 VLMs have $\mathrm{BR}>0.7$, indicating that bias-related failures dominate residual errors. In contrast, the random baseline yields nearly identical BR ($0.374$) and NBER ($0.376$) despite high CER ($0.750$), supporting our distinction between structured bias and chance errors. Moreover, PRR exceeds $0.6$ for 13 models, with LBR ranging from $0.357$ to $0.716$ and TOR reaching $0.779$, revealing substantial prior retention and language-prior effects.

\noindent\textbf{Scaling reduces but does not eliminate spatial bias.}
Larger models generally exhibit lower bias within model families, although accuracy gains are not monotonic. For Qwen3-VL, $\mathrm{mAcc}$ increases from $0.414$ (2B) to $0.510$ (32B), while BR decreases from $0.749$ to $0.665$. Yet higher accuracy does not imply lower bias: GLM-4.1V-9B-Thinking achieves $\mathrm{mAcc}=0.505$ with $\mathrm{BR}=0.694$. Overall, proprietary VLMs achieve higher average $\mathrm{mAcc}$ ($0.549$ vs.\ $0.441$) and lower BR ($0.600$ vs.\ $0.770$) than open-source models, while spatially specialized models show no consistent advantage.

\begin{figure}[t]
    \centering
    \includegraphics[width=0.9\linewidth]{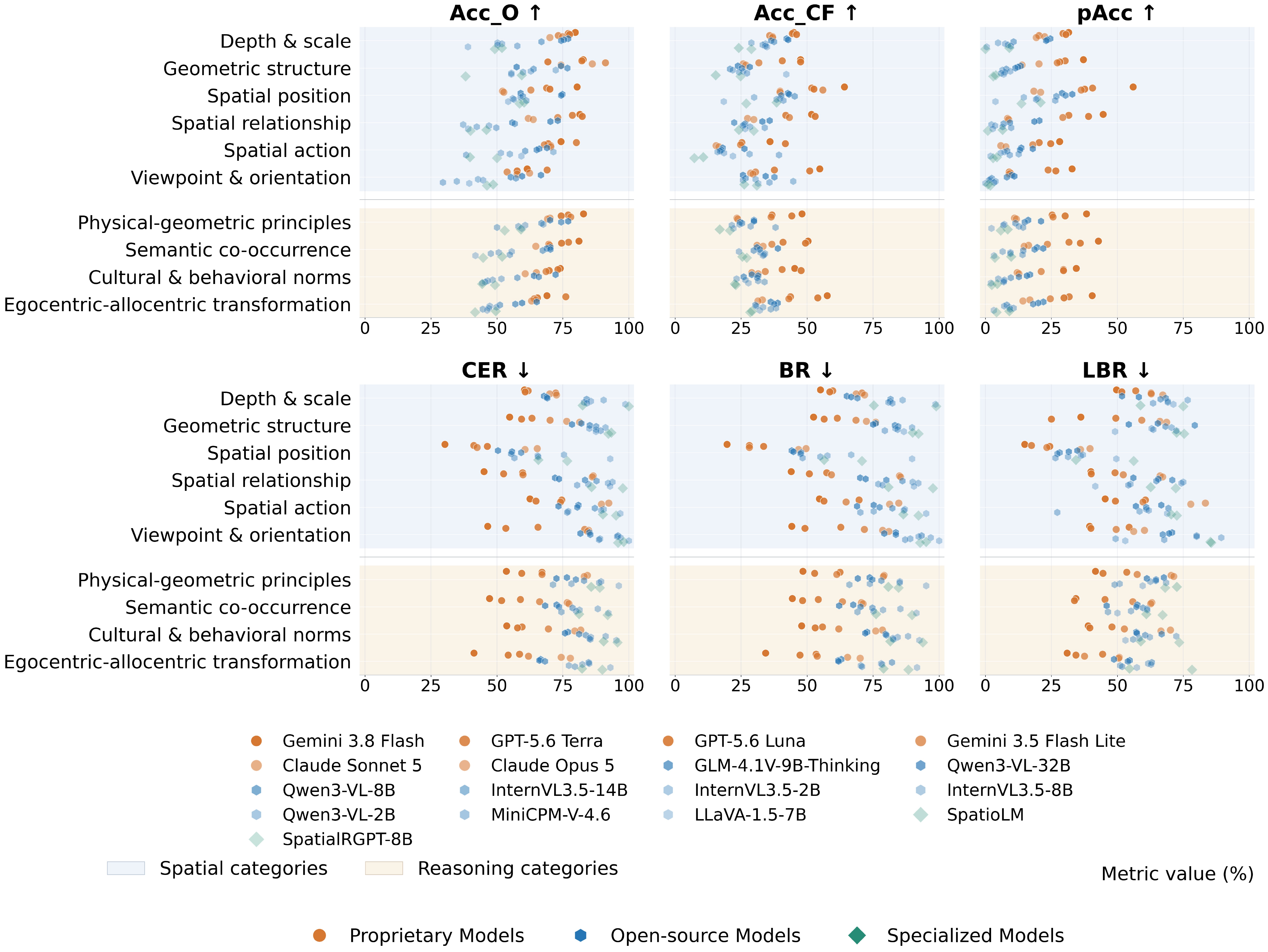}
    \vspace{-10pt}
    \caption{Fine-grained evaluation across spatial tasks and prior sources.}
    \label{fig:scatter}
    \vspace{-12pt}
\end{figure}

\subsection{Category-Level Fine-Grained Analysis}
\label{sec:fine_grained_analysis}

Fig.~\ref{fig:scatter} reveals variations across spatial tasks and bias sources, leading to the following findings:

\begin{wrapfigure}[18]{r}{0.6\textwidth}
\centering
\includegraphics[width=\linewidth]{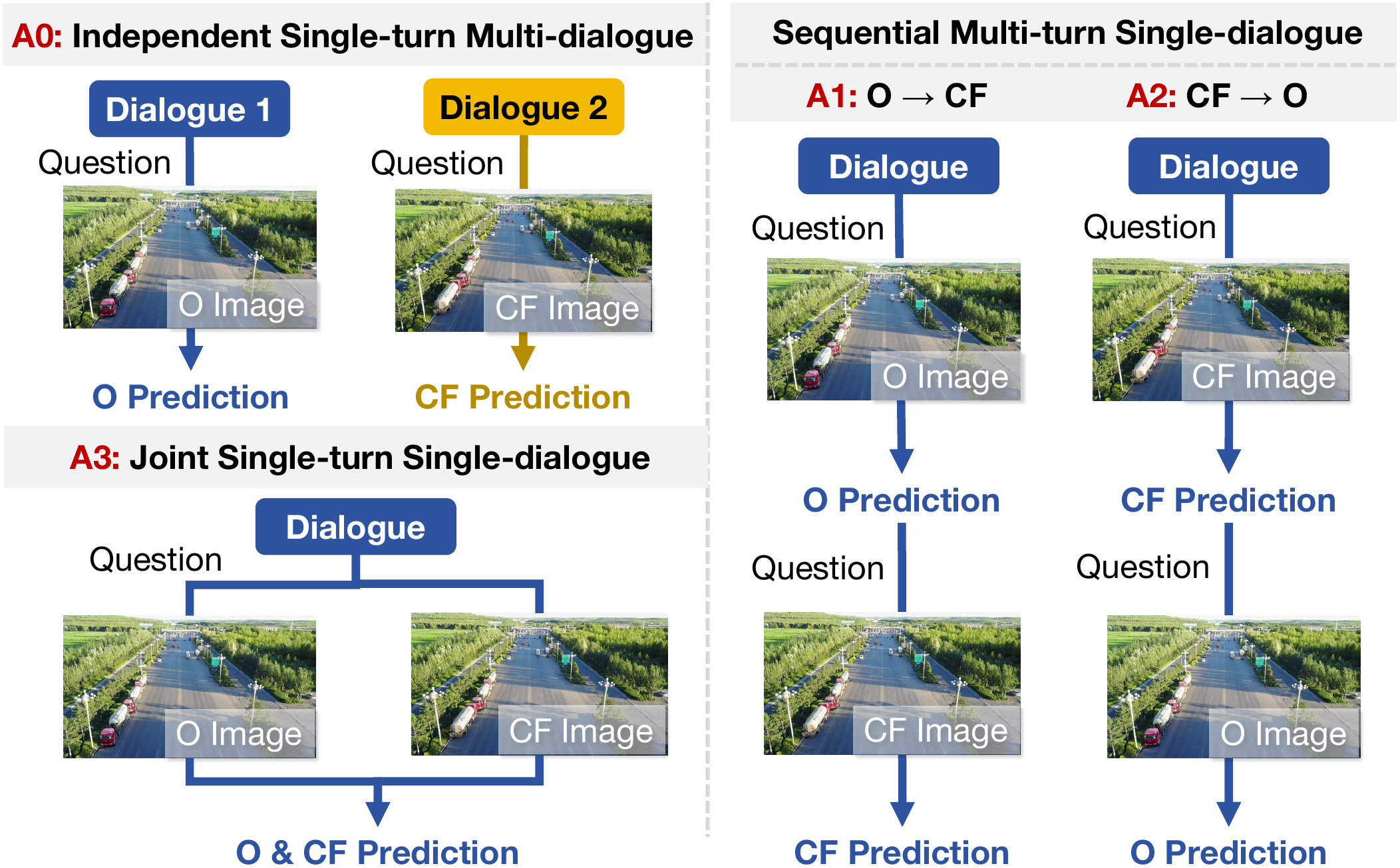}
\vspace{-10pt}
\caption{\textbf{Input configurations.} Independent single-turn multi-dialogue (A0), sequential multi-turn single-dialogue (A1/A2), and joint single-turn single-dialogue (A3) settings.}
\label{fig:input_config}
\vspace{-10pt}
\end{wrapfigure}

\noindent\textbf{Proprietary models are consistently more robust.}
Across most categories, proprietary VLMs achieve higher accuracy and lower CER and BR. Specialized spatial models often match or underperform general-purpose open-source VLMs, showing that spatial specialization alone does not ensure robust counterfactual grounding.

\noindent\textbf{Spatial bias is task-dependent.}
Spatial position is among the most robust categories, with relatively low CER, BR, and LBR, whereas viewpoint \& orientation, geometric structure, and spatial relationship are more vulnerable. The pronounced original--counterfactual gap further shows that standard accuracy can obscure task-specific bias.

\noindent\textbf{Prior-source effects are more model-dependent than task effects.}
All four prior sources induce substantial counterfactual failures, but their relative difficulty varies across models, suggesting that spatial bias depends not only on the prior type but also on how each VLM encodes and exploits it.

\subsection{Effects of Input Configurations}

\begin{figure}[t]
\centering
\includegraphics[width=0.95\linewidth]{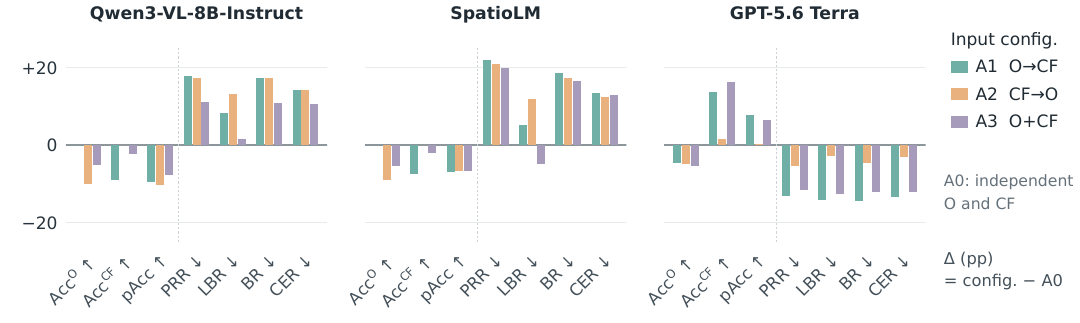}
\vspace{-10pt}
\caption{Comparison of evaluation metric changes across input configurations.
}
\label{fig:input_effects}
\end{figure}

This section evaluates three representative models under different input configurations in Fig.~\ref{fig:input_config} to examine memory effects and contrastive reasoning. As shown in Fig.~\ref{fig:input_effects}:

\noindent\textbf{Sequential context induces model-dependent memory effects.}
Relative to A0, O$\rightarrow$CF context (A1) substantially increases PRR/BR for Qwen3-VL-8B (+17.9/+17.4 pp) and SpatioLM (+22.1/+18.6 pp), while degrading $\mathrm{Acc}^{CF}$ and $pAcc$. In contrast, GPT-5.6 Terra improves $\mathrm{Acc}^{CF}$ by 13.8 pp and reduces BR/CER by 14.4/13.5 pp. Thus, preceding spatial evidence can either reinforce prior predictions or facilitate counterfactual adaptation.

\noindent\textbf{Input order affects counterfactual behavior.}
Reversing A1 from O$\rightarrow$CF to CF$\rightarrow$O (A2) restores $\mathrm{Acc}^{CF}$ for Qwen3-VL-8B ($0.193\!\rightarrow\!0.283$) and SpatioLM ($0.160\!\rightarrow\!0.235$), but increases their LBR. GPT-5.6 Terra shows the opposite pattern, performing substantially better under O$\rightarrow$CF. This asymmetry reveals VLM's sensitivity to the order of conflicting evidence.

\noindent\textbf{Joint context helps models that effectively reconcile conflicting evidence.}
Relative to A0, A3 increases BR by 10.8 and 16.6 pp for Qwen3-VL-8B and SpatioLM, respectively, while reducing $pAcc$. Conversely, GPT-5.6 Terra improves $\mathrm{Acc}^{CF}$ by 16.4 pp and reduces BR/CER by 12.2/12.1 pp. 

Overall, additional cross-condition context amplifies spatial bias in the evaluated open-source and specialized models, but facilitates counterfactual adaptation in the proprietary model.

\subsection{Effects of Background Configurations}

\begin{figure}[t]
\centering
\includegraphics[width=0.95\linewidth]{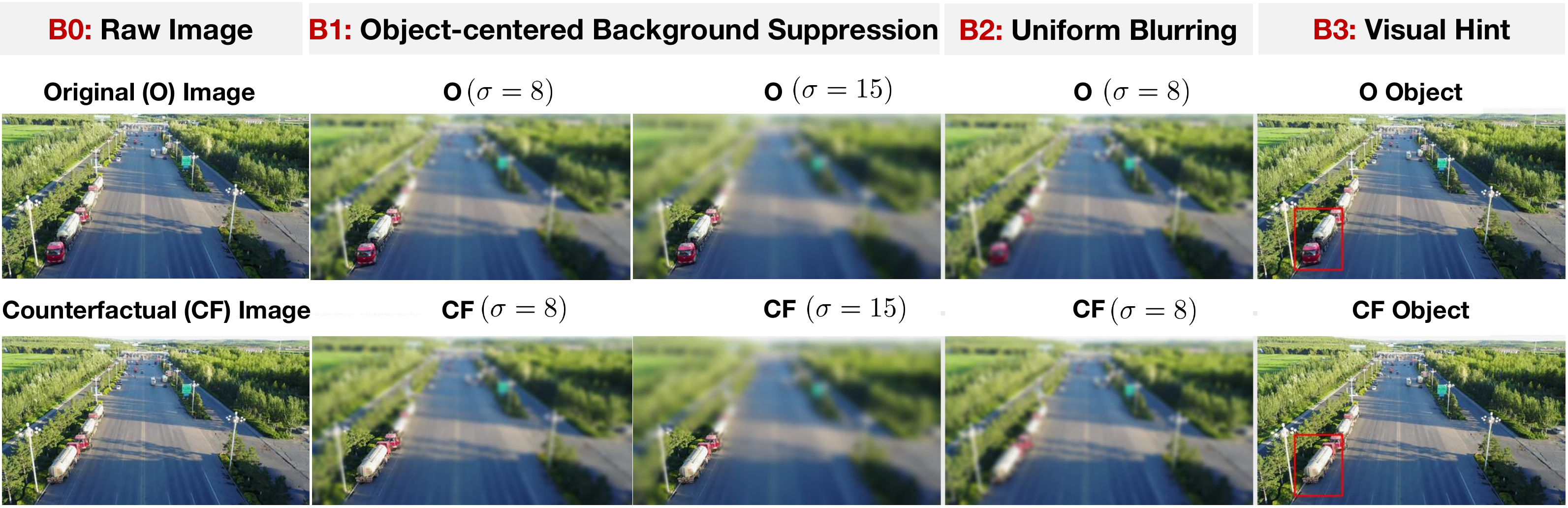}
\vspace{-10pt}
\caption{\textbf{Background configurations.} B0: raw image; B1: object-centered background suppression; B2: uniform full-image blurring; B3: bounding-box highlighting. $\sigma$ denotes the standard deviation of the Gaussian blur.}
\label{fig:background_supp}
\vspace{-10pt}
\end{figure}

This section examines how background information affects spatial bias by contrasting irrelevant-context suppression with useful visual guidance, as shown in Fig.~\ref{fig:background_supp}; quantitative results are shown in Fig.~\ref{fig:background_effects}. Implementation details are provided in Appendix~\ref{app:background_supp}.

\begin{figure}[t]
    \centering
    \includegraphics[width=0.95\linewidth]{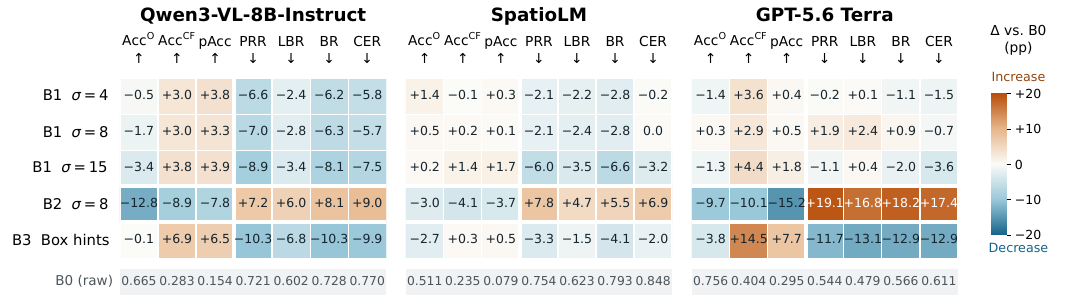}
    \vspace{-15pt}
    \caption{Comparison of evaluation metric changes across background configurations.}
    \label{fig:background_effects}
    \vspace{-10pt}
\end{figure}

\noindent\textbf{Selective background suppression generally improves counterfactual robustness.}
B1 generally increases $\mathrm{Acc}^{CF}$ and $pAcc$ while reducing BR and CER, with $\sigma=15$ yielding the most consistent gains. At this setting, BR drops by 8.1, 6.6, and 2.0 pp for Qwen3-VL-8B, SpatioLM, and GPT-5.6 Terra, respectively, suggesting that irrelevant context can reinforce prior-dependent predictions.

\noindent\textbf{Uniform blurring consistently amplifies spatial bias.}
B2 reduces all accuracy metrics while increasing error and bias rates across models. For GPT-5.6 Terra, $pAcc$ drops by 15.2 pp while BR rises by 18.2 pp, showing that B1's gains arise from selective suppression rather than blurring itself.

\noindent\textbf{Explicit visual guidance is effective but model-dependent.}
B3 yields the strongest gains for Qwen3-VL-8B and GPT-5.6 Terra, reducing BR by 10.3 and 12.9 pp while increasing $\mathrm{Acc}^{CF}$ by 6.9 and 14.5 pp, respectively. SpatioLM instead benefits more from B1 ($\sigma=15$), with BR decreasing by 6.6 pp versus 4.1 pp under B3.

Overall, emphasizing task-relevant regions improves counterfactual grounding, whereas indiscriminate visual degradation amplifies spatial bias.

\subsection{Spatial Bias Mitigation}

We evaluate four mitigation strategies in Fig.~\ref{fig:mitigation} and obtain the following findings:

\begin{figure}[t]
    \centering
    \includegraphics[width=0.95\linewidth]{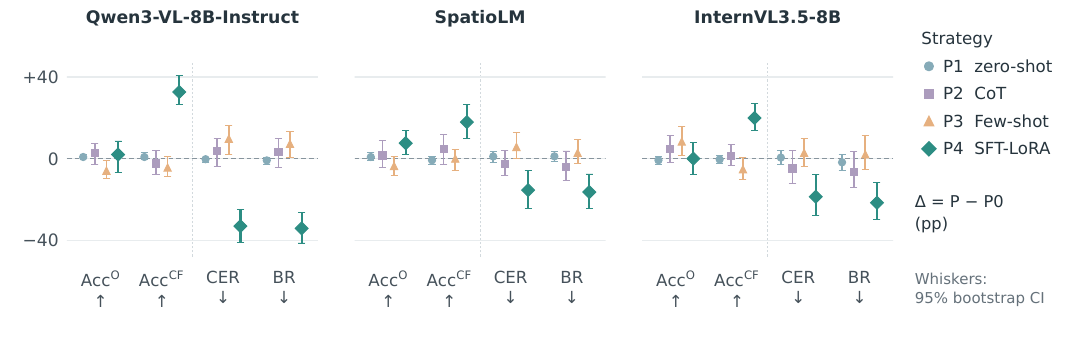}
    \vspace{-15pt}
    \caption{\textbf{Comparison of spatial bias mitigation strategies.} P0: baseline; P1: zero-shot linguistic intervention; P2: CoT incorporation; P3: few-shot prompting; P4: SFT with LoRA.
    }
    \label{fig:mitigation}
    \vspace{-15pt}
\end{figure}

\noindent\textbf{Prompt-based mitigation is limited and model-dependent.}
P1 produces only marginal changes, while P2 yields modest BR reductions for SpatioLM ($-4.1$ pp) and InternVL3.5-8B ($-6.6$ pp) but degrades Qwen3-VL-8B ($\mathrm{Acc}^{CF}$ $-2.4$ pp, BR $+3.2$ pp). P3 is less stable, increasing CER/BR by $+9.8/+7.3$ pp for Qwen3-VL-8B and $+5.7/+2.8$ pp for SpatioLM. Thus, prompting alters spatial behavior but does not reliably mitigate bias across models.

\noindent\textbf{SFT provides the strongest and most consistent mitigation.}
P4 improves $\mathrm{Acc}^{CF}$ while reducing CER and BR across all three models. For Qwen3-VL-8B, $\mathrm{Acc}^{CF}$ rises by $32.7$ pp, while CER and BR drop by $33.1$ and $34.2$ pp. BR also decreases by $16.4$ and $21.6$ pp for SpatioLM and InternVL3.5-8B, respectively, with $\mathrm{Acc}^{O}$ preserved or improved.

\noindent\textbf{Parameter-level adaptation is more reliable than prompt-level correction.}
Prompt-based methods yield weak or unstable gains, whereas bias-aware SFT consistently improves counterfactual robustness, indicating that explicit training signals are more effective than prompting alone.

\section{Conclusion}

In this work, we systematically investigate spatial bias in VLMs, shifting the focus from whether models can reason spatially to how their spatial errors are systematically structured. We construct an original--counterfactual embodied VQA benchmark and introduce a multi-stage diagnostic framework with bias-aware metrics beyond accuracy. Across diverse VLMs, spatial tasks, input configurations, background interventions, and mitigation settings, we find spatial bias to be widespread, closely associated with prior retention and language-aligned errors, and sensitive to both visual evidence and evaluation context. Prompt-based interventions yield limited and inconsistent gains, whereas supervised fine-tuning substantially improves counterfactual robustness and reduces bias. These findings underscore the need to explicitly diagnose and mitigate spatial bias in VLMs for more reliable, evidence-grounded embodied multimodal systems.




\subsection*{Acknowledgments}
This work was supported by the SSTIC Grants 
(KJZD20230923115106012, KJZD20230923114916032, 
and GJHZ20240218113604008). 
Specifically, we thank Profs. Xinlei Chen and Chen Gao for their discussions during the early stage of this work.

\subsection*{AI use statement}
In this work, we used generative AI tools for language polishing and auxiliary data generation.
We have not used generative AI tools for experiment design or model development,
and other required disclosure tasks are not applicable to this work.
We have reviewed all AI-assisted content carefully.
We take full responsibility for the final content of this work.

\subsection*{Ethics Statement}
This work complies with academic ethics. No human subjects are involved. All datasets used are publicly available and properly cited. There are no harmful findings or privacy risks. While spatial bias may lead to unreliable decisions in embodied applications, our benchmark and diagnostic framework are designed to expose and mitigate such risks rather than enable harmful use. We provide sufficient implementation details to facilitate reproducibility. This research meets legal requirements and research integrity standards.

\subsection*{Reproducibility statement}
We provide resources to support the reproducibility of this work. The implementation details, experimental settings and data preprocessing are described in the main paper or supplementary materials. The source code will be released. All theoretical notations and calculations are included.



\bibliography{reference_v3}
\bibliographystyle{iclr2027_conference}
\appendix

\section{Related Work}
\label{sec:related_work}

\textbf{Spatial Reasoning in VLMs.}
Spatial reasoning is a core capability of embodied VLMs and has been increasingly studied across diverse settings. VSR~\citep{liu2023vsr} and What's ``up'' with VLMs?~\citep{kamath2023whatsup} evaluate basic spatial relations and reference frames, while SpatialVLM~\citep{chen2024spatialvlm} targets quantitative 3D reasoning through large-scale spatial supervision. More recent benchmarks extend evaluation to scene-level spatial intelligence, 3D/6D relations, interactive planning, and cross-view reasoning~\citep{yang2025thinking,wang2025spatial457,mayer2025ivispar,zha2026aircopbench,xu2026citycube}. These studies reveal persistent spatial reasoning limitations, but primarily evaluate whether models answer spatial questions correctly rather than how their errors are systematically structured.

\textbf{Spatial Failure Diagnostics.}
Recent benchmarks increasingly reveal structured spatial failure patterns beyond aggregate accuracy. Position-related errors have been studied in both multi-image~\citep{tian2025positionbias} and within-image~\citep{zhu2026spatialbias} settings, while SpinBench~\citep{zhang2025spinbench}, FlipSet~\citep{wang2026egocentric}, and Spatial457~\citep{wang2025spatial457} expose systematic weaknesses in rotation, perspective taking, and attribute-dependent spatial prediction. Counterfactual evaluation provides a complementary perspective: edited-image studies~\citep{vo2026vlmsbiased} reveal semantic-prior dependence, V$^2$R-Bench~\citep{fan2025visualvariations} evaluates robustness to controlled visual variations, and MindEdit-Bench~\citep{yu2026mindedit} examines hypothetical spatial transformations. In contrast, we directly present matched original--counterfactual spatial evidence and diagnose whether VLMs systematically retain prior-consistent predictions across diverse spatial reasoning tasks.

\textbf{VLM Bias and Hallucination.}
Vision--language models often exploit learned priors instead of relying on current visual evidence. Early VQA work such as VQA-CP~\citep{agrawal2018dont} showed strong dependence on answer priors, while recent VLM studies have examined social and demographic biases~\citep{wang2026vlbiasbench,huang2025visbias}. Related work further shows that pretrained semantic or visual priors can override image evidence under counterfactual or visual variations~\citep{vo2026vlmsbiased,fan2025visualvariations}. Hallucination studies similarly investigate outputs unsupported by visual evidence~\citep{li2023pope,wang2024amber,guan2024hallusionbench}. Together, these works highlight a broader tension between learned priors and visual grounding, motivating our study of this phenomenon specifically in spatial reasoning.

\textbf{Bias and Hallucination Mitigation.}
Existing VLM bias and hallucination mitigation methods mainly rely on inference-time correction or model adaptation. Representative approaches such as LURE~\citep{zhou2024lure}, OPERA~\citep{huang2024opera}, and Woodpecker~\citep{yin2024woodpecker} improve visual grounding through post-hoc or decoding-time intervention. Our work instead evaluates both prompt-level interventions and supervised fine-tuning specifically for spatial bias, examining whether prior-dependent spatial errors can be reduced under controlled counterfactual evaluation.

\section{Diagnostic Metrics}
\label{app:diag_metrics}

Accuracy alone cannot reveal whether spatial failures are random or systematically associated with learned priors.
We therefore introduce a set of diagnostic metrics to characterize spatial capability, counterfactual failure, bias prevalence, and language-prior dependence through the following research questions (RQs):

\begin{itemize}[nosep,leftmargin=*]
    \item[] \textbf{RQ1: Spatial Reasoning Performance.}
    How well do VLMs maintain accurate and consistent spatial reasoning across original and counterfactual conditions?

    \item[] \textbf{RQ2: Counterfactual Failure.}
    What proportion of counterfactual failures are attributable to bias-related versus non-bias errors?

    \item[] \textbf{RQ3: Spatial Bias Prevalence.}
    To what extent do VLMs retain or follow prior-consistent spatial predictions despite contradictory counterfactual evidence?

    \item[] \textbf{RQ4: Language-prior Prevalence.}
     What fraction of eligible samples exhibit a language prior toward the original spatial answer without visual evidence?

    \item[] \textbf{RQ5: Language-prior Dominance.}
    When language priors conflict with counterfactual visual evidence, how often do they override the visual evidence?

    \item[] \textbf{RQ6: Language-aligned Spatial Bias.}
    What proportion of spatial bias aligns with erroneous language priors, and how much remains non-language-aligned?
\end{itemize}

\textbf{Metric Overview.}
The designed metrics form a hierarchical diagnosis from \emph{spatial capability} to \emph{counterfactual failure}, \emph{spatial bias}, and finally \emph{language-prior attribution}.
For \textbf{RQ1}, $Acc^{O}$, $Acc^{CF}$, $pAcc$, and $mAcc$ establish overall spatial reasoning accuracy and consistency.
Building on samples correctly answered in the original condition, \textbf{RQ2} uses Counterfactual Error Rate (CER) to quantify failures induced by counterfactual changes, with Non-Bias Error Rate (NBER) isolating residual errors beyond identified bias.
\textbf{RQ3} further characterizes these systematic failures through Prior Retention Rate (PRR), which captures persistence of the original answer, and Bias Rate (BR), which measures overall spatial bias.
Finally, \textbf{RQ4--RQ6} trace the role of language priors: Language Prior Dependence (LPD) measures their prevalence, Text Override Rate (TOR) measures their dominance when language and visual evidence conflict, and Language-aligned Bias Rate (LBR) measures their manifestation as spatial bias.
Thus, LPD, TOR, and LBR respectively characterize whether a language prior exists, whether it overrides conflicting visual evidence, and whether it manifests as spatial bias.
When $\mathrm{LPD}=1$, all eligible samples constitute language--vision conflicts, yielding $\mathrm{TOR}=\mathrm{LBR}$.
Detailed definitions and calculations are provided in Table~\ref{tab:metrics}.

\textbf{Diagnostic Notations.}
Let $\mathds{1}(\cdot)$ denote the indicator function. 
%
{
\color{black}
We denote the set of $N$ paired original--counterfactual (O--CF) samples as
$\mathcal{D}=\{(I_i^O,I_i^{CF},q_i,y_i^O,y_i^{CF})\}_{i=1}^N$.
}
Then, we define the flipped-answer set $\mathcal{D}_{\mathrm{flip}}=\{i:(I_i,q_i)\in\mathcal{D},\ y_i^{CF}\neq y_i^O\}$, which contains samples whose ground-truth answers change after counterfactual editing. For bias diagnosis, we further restrict to the eligible set $\mathcal{E}=\{i\in\mathcal{D}_{\mathrm{flip}}:\hat y_i^O=y_i^O\}$, thereby excluding samples that are already incorrect on the original input. We denote the text-only prediction as $\hat y_i^T=\mathcal{M}(q_i)$, obtained without image input. The overall bias set is defined as $\mathcal{B}=\mathcal{B}^O\cup\mathcal{B}^L$, where $\mathcal{B}^O=\{i\in\mathcal{D}_{\mathrm{flip}}:\hat y_i^{CF}=y_i^O\}$ denotes original-answer retention bias, and $\mathcal{B}^L=\{i\in\mathcal{D}_{\mathrm{flip}}:\hat y_i^{CF}=\hat y_i^T\neq y_i^{CF}\}$ denotes language-aligned bias. Finally, we define the language--vision conflict set as $\mathcal{C}=\{i\in\mathcal{E}:\hat y_i^T\neq y_i^{CF}\}$.

\begin{table*}[htbp]
\centering
\caption{Definitions and calculations of evaluation metrics for each research question.}
\label{tab:metrics}

\renewcommand{\arraystretch}{1.5} 
\setlength{\tabcolsep}{4pt}      

\scalebox{0.77}{%
\begin{tabular}{
    >{\raggedright\arraybackslash}p{1.5cm} 
    >{\raggedright\arraybackslash}p{2.5cm} 
    >{\centering\arraybackslash}p{7.4cm}   
    >{\raggedright\arraybackslash}p{4.5cm} 
    >{\centering\arraybackslash}p{1.0cm}   
}
\toprule
\textbf{Purpose} & \textbf{Metric} & \textbf{Formula} & \textbf{Interpretation} & \textbf{RQ} \\
\midrule

\multirow{4}{*}{\shortstack[l]{Accuracy\\Diagnosis}} 
 & O Accuracy       & $Acc^O = \frac{1}{N} \sum_{i=1}^N \mathds{1}(\hat{y}_i^O = y_i^O)$ & Accuracy on O data & RQ1 \\
 & CF Accuracy             & $Acc^{CF} = \frac{1}{N} \sum_{i=1}^N \mathds{1}(\hat{y}_i^{CF} = y_i^{CF})$ & Accuracy on CF data & RQ1 \\
 & Pair Accuracy           & $pAcc = \frac{1}{N}\sum_{i=1}^N \mathds{1}(\hat{y}_i^O = y_i^O \land \hat{y}_i^{CF} = y_i^{CF})$ & Pairwise accuracy consistency between O and CF pairs & RQ1 \\
 & Mean Accuracy           & \textcolor{black}{$mAcc = \frac{1}{2}(Acc^O+Acc^{CF})$} & Average accuracy across all O and CF data & RQ1 \\ 
\midrule

\multirow{2}{*}{\shortstack[l]{Failure\\Diagnosis}} 
 & CF Error Rate           & $CER = \frac{1}{|\mathcal{E}|}{\sum_{i \in \mathcal{E}} \mathds{1}(\hat{y}_i^{CF} \neq y_i^{CF})}$ & Failure rate from counterfactual perturbation & RQ2 \\
 & Non-bias CF Error Rate       & $NBER = CER - BR$ & Failure rate unrelated to spatial bias & RQ2 \\
\midrule

\multirow{5}{*}{\shortstack[l]{Bias\\Diagnosis}} 
 & Prior Retention Rate      & $PRR = \frac{1}{|\mathcal{E}|}{\sum_{i \in \mathcal{E}} \mathds{1}(\hat{y}_i^{CF} = y_i^O)}$ & Rate of CF prediction clinging to original ground truth & RQ3 \\
 & Bias Rate                 & $BR = \frac{1}{|\mathcal{E}|} \sum_{i \in \mathcal{E}}\mathds{1}(i \in \mathcal{B})$ & Proportion of CF spatial errors attributed to spatial bias & RQ3 \\
 & Language-prior Dependency 
 & $LPD=\frac{1}{|\mathcal{E}|}{\sum_{i\in\mathcal{E}}\mathds{1}(\hat y^{T}_i=y^O_i)}$ 
 & Model dependency on pure language priors & RQ4 \\
 & Text Override Rate 
 & $TOR=\frac{1}{|\mathcal{C}|}{\sum_{i\in\mathcal{C}}\mathds{1}(\hat y^{CF}_i=\hat y^T)}$ 
 & Rate of text overriding visual evidence & RQ5 \\
 & Language-prior Bias Rate  
 & $LBR=\frac{1}{|\mathcal{E}|}{\sum_{i\in\mathcal{E}}\mathds{1}(y^{CF}_i=\hat y^{T}_i \land\hat y^T_i\neq y^{CF}_i)}$
 & Overall spatial bias rate from language priors & RQ6 \\
\bottomrule
\end{tabular}%
}
\vspace{-10pt}
\end{table*}

\section{More Experimental Results}
\label{app:more_results}


\providecommand{\appNModels}{8}
\providecommand{\appNUp}{6}
\providecommand{\appNDown}{2}
\providecommand{\appNConsensus}{81}
\providecommand{\appNRawPairs}{1254}
\providecommand{\appGpuHours}{138.3}
\providecommand{\appLocalCalls}{244,530}
\providecommand{\appApiCalls}{90,288}
\providecommand{\appDiagGpuHours}{5.3}

\noindent\textbf{What is added here.} The main paper reports the ablation on three representative models in order to keep the story compact. The ablation study itself was run on \appNModels{} models that span the frontier-to-small axis and three architectural families (two proprietary APIs; the Qwen3-VL ladder; InternVL3.5 and its spatial fine-tune, plus a second spatially specialised 8B model).

\subsection{More Quantitative Results}
\label{app:more_quant}

{
\color{black}
\subsubsection{Human Validation and Annotation Reliability}
\label{app:human_perform}
To assess the validity of counterfactual edits, visual identifiability, and annotation reliability, we conduct a human evaluation involving three independent annotators on all original--counterfactual pairs. Each annotator independently answers the four-option spatial questions for both original (O) and counterfactual (CF) images without access to ground-truth labels and flags potentially ambiguous cases.
As shown in Tab.~\ref{tab:app_human_perform}, human annotators achieve average accuracies of 92.88\% and 93.04\% on O and CF images, respectively, with a paired accuracy of 89.07\%. The comparable O/CF performance (a difference of only 0.16 pp) suggests that counterfactual editing does not systematically increase task difficulty for human observers. Inter-annotator agreement is also high, with Fleiss' $\kappa$ values of 0.866 and 0.858 for O and CF images, respectively (0.862 pooled), supporting the consistency of human judgments across both conditions.
Disagreement, defined as non-unanimous answer selection, occurs in 14.91\% of O and 15.71\% of CF questions. Separately, potentially ambiguous pairs are explicitly flagged by at least one annotator in 11.56\% of cases and by a majority in only 1.52\%, with an average individual flag rate of 4.39\%. 
%
Overall, the high human accuracy, comparable O/CF performance, and strong inter-annotator agreement support the visual interpretability and label reliability of the benchmark. These results reduce the concern that the observed counterfactual failures primarily arise from inherently ambiguous edits or incorrect labels.

\begin{table}[t]
\centering
\caption{\textcolor{black}{\textbf{Human validation and annotation reliability.}
Three independent annotators evaluate all original--counterfactual pairs without access to ground-truth labels.
Accuracy metrics are reported for each annotator, while Fleiss' $\kappa$ measures inter-annotator agreement.
Disagreement denotes non-unanimous answer selection.
}}
\label{tab:app_human_perform}
\small
\begin{tabular}{lcccc}
\toprule
\textbf{Annotator} &
$\mathbf{Acc^O}\uparrow$ &
$\mathbf{Acc^{CF}}\uparrow$ &
$\mathbf{pAcc}\uparrow$ &
$\mathbf{mAcc}\uparrow$ \\
\midrule
Annotator 1 & 0.9011 & 0.9179 & 0.8589 & 0.9095 \\
Annotator 2 & 0.9777 & 0.9601 & 0.9426 & 0.9689 \\
Annotator 3 & 0.9075 & 0.9131 & 0.8708 & 0.9103 \\
\midrule
\textbf{Average} & \textbf{0.9288} & \textbf{0.9304} &
\textbf{0.8907} & \textbf{0.9296} \\
\midrule
\multicolumn{5}{l}{\textit{Inter-annotator agreement}} \\
\midrule
\textbf{Metric} & \textbf{O} & \textbf{CF} & \textbf{O+CF} & \\
Fleiss' $\kappa$ & 0.8656 & 0.8579 & 0.8618 & \\
Disagreement (\%) & 14.91 & 15.71 & 15.31 & \\
\bottomrule
\end{tabular}
\end{table}

}



\subsubsection{Effects of Input and Background Configurations}
\label{app:input_background}
\paragraph{Input configurations (Ablation~A).} Tab.~\ref{tab:app_abl_a} reports all \appNModels{} models under the four input configurations. Two facts that are visible only on the full axis are worth stating. First, the direction of the sequential-context effect is not a property of the model: the mean BR over A1--A3 falls relative to A0 for \appNDown{} models and rises for the other \appNUp{}. The split follows the baseline operating point rather than the parameter count --- within the Qwen3-VL family the change is essentially flat from 2B to 32B even though the original-image and counterfactual accuracy improve substantially over that range. This is the reason the main text insists that a bias rate be quoted together with the input configuration that produced it. Second, the mechanism is carried by the retained-prior component (PRR), not by the language-prior component (LBR): where BR rises, PRR rises by a much larger margin than LBR, and where BR falls, both components fall. The counterfactual-first condition (A2, CF$\to$O) is the one effect that is same-signed for every model: it makes all \appNModels{} models lose part of the original-image questions they had answered correctly, independently of their strength. A single bias rate therefore cannot adjudicate an input configuration, because the two image sides move in opposite directions.

\paragraph{Background configurations (Ablation~B).} Tab.s~\ref{tab:app_abl_b_sigma} and~\ref{tab:app_abl_b_control} give the object-centered blurring sweep (B1) and the two controls (B2, B3) on the full axis. The falsification control B2 (uniform full-image blurring, no sharp target region) moves every model in the wrong direction, whereas the mechanism control B3 (bounding-box visual hints: no pixel changed, the target region is merely highlighted) is the strongest single intervention for most models. The qualitative conclusion of the main text --- that the gain comes from \emph{telling the model where to look} rather than from \emph{removing background} --- therefore survives on the extended set.


\begin{table*}[t]
\centering
\caption{\textbf{Full ablation-A results across the capability axis.} ``More model results'' for the presentation-protocol ablation: the main text reports three models, here we report all eight. Conditions: A0 independent; A1 o\_first (same dialogue, O then CF, with history); A2 cf\_first (reversed order); A3 joint (both images in one message, two answers requested). Values in parentheses are percentage-point changes from A0 for the same model; \textcolor{red}{red} = increase, \textcolor{green!60!black}{green} = decrease.}
\label{tab:app_abl_a}
\small
\resizebox{\textwidth}{!}{
\begin{tabular}{llccccccc}
\toprule
\textbf{Model} & \textbf{Cond.} & $\mathbf{Acc^{O}}\uparrow$ & $\mathbf{Acc^{CF}}\uparrow$ & $\mathbf{pAcc}\uparrow$ & $\mathbf{PRR}\downarrow$ & $\mathbf{LBR}\downarrow$ & $\mathbf{BR}\downarrow$ & $\mathbf{CER}\downarrow$ \\
\midrule
\multirow{4}{*}{Gemini 3.8 Flash}
& A0 & 0.756 & 0.490 & 0.372 & 0.448 & 0.357 & 0.450 & 0.509 \\
& A1 & 0.759 {\scriptsize \textcolor{red}{($\mathbf{+0.3}$)}} & 0.588 {\scriptsize \textcolor{red}{($\mathbf{+9.8}$)}} & 0.457 {\scriptsize \textcolor{red}{($\mathbf{+8.5}$)}} & 0.329 {\scriptsize \textcolor{green!60!black}{($\mathbf{-11.9}$)}} & 0.257 {\scriptsize \textcolor{green!60!black}{($\mathbf{-10.0}$)}} & 0.334 {\scriptsize \textcolor{green!60!black}{($\mathbf{-11.6}$)}} & 0.396 {\scriptsize \textcolor{green!60!black}{($\mathbf{-11.3}$)}} \\
& A2 & 0.749 {\scriptsize \textcolor{green!60!black}{($\mathbf{-0.7}$)}} & 0.506 {\scriptsize \textcolor{red}{($\mathbf{+1.6}$)}} & 0.375 {\scriptsize \textcolor{red}{($\mathbf{+0.2}$)}} & 0.442 {\scriptsize \textcolor{green!60!black}{($\mathbf{-0.7}$)}} & 0.350 {\scriptsize \textcolor{green!60!black}{($\mathbf{-0.7}$)}} & 0.448 {\scriptsize \textcolor{green!60!black}{($\mathbf{-0.2}$)}} & 0.501 {\scriptsize \textcolor{green!60!black}{($\mathbf{-0.9}$)}} \\
& A3 & 0.769 {\scriptsize \textcolor{red}{($\mathbf{+1.3}$)}} & 0.577 {\scriptsize \textcolor{red}{($\mathbf{+8.7}$)}} & 0.451 {\scriptsize \textcolor{red}{($\mathbf{+7.9}$)}} & 0.353 {\scriptsize \textcolor{green!60!black}{($\mathbf{-9.5}$)}} & 0.262 {\scriptsize \textcolor{green!60!black}{($\mathbf{-9.5}$)}} & 0.357 {\scriptsize \textcolor{green!60!black}{($\mathbf{-9.3}$)}} & 0.413 {\scriptsize \textcolor{green!60!black}{($\mathbf{-9.6}$)}} \\
\midrule
\multirow{4}{*}{GPT-5.6 terra}
& A0 & 0.756 & 0.404 & 0.295 & 0.544 & 0.479 & 0.566 & 0.611 \\
& A1 & 0.709 {\scriptsize \textcolor{green!60!black}{($\mathbf{-4.7}$)}} & 0.542 {\scriptsize \textcolor{red}{($\mathbf{+13.8}$)}} & 0.372 {\scriptsize \textcolor{red}{($\mathbf{+7.7}$)}} & 0.411 {\scriptsize \textcolor{green!60!black}{($\mathbf{-13.2}$)}} & 0.338 {\scriptsize \textcolor{green!60!black}{($\mathbf{-14.1}$)}} & 0.422 {\scriptsize \textcolor{green!60!black}{($\mathbf{-14.4}$)}} & 0.476 {\scriptsize \textcolor{green!60!black}{($\mathbf{-13.5}$)}} \\
& A2 & 0.708 {\scriptsize \textcolor{green!60!black}{($\mathbf{-4.8}$)}} & 0.420 {\scriptsize \textcolor{red}{($\mathbf{+1.6}$)}} & 0.297 {\scriptsize \textcolor{red}{($\mathbf{+0.2}$)}} & 0.490 {\scriptsize \textcolor{green!60!black}{($\mathbf{-5.4}$)}} & 0.453 {\scriptsize \textcolor{green!60!black}{($\mathbf{-2.7}$)}} & 0.520 {\scriptsize \textcolor{green!60!black}{($\mathbf{-4.6}$)}} & 0.580 {\scriptsize \textcolor{green!60!black}{($\mathbf{-3.0}$)}} \\
& A3 & 0.703 {\scriptsize \textcolor{green!60!black}{($\mathbf{-5.3}$)}} & 0.569 {\scriptsize \textcolor{red}{($\mathbf{+16.4}$)}} & 0.360 {\scriptsize \textcolor{red}{($\mathbf{+6.5}$)}} & 0.429 {\scriptsize \textcolor{green!60!black}{($\mathbf{-11.5}$)}} & 0.353 {\scriptsize \textcolor{green!60!black}{($\mathbf{-12.7}$)}} & 0.444 {\scriptsize \textcolor{green!60!black}{($\mathbf{-12.2}$)}} & 0.489 {\scriptsize \textcolor{green!60!black}{($\mathbf{-12.1}$)}} \\
\midrule
\multirow{4}{*}{Qwen3-VL-32B}
& A0 & 0.673 & 0.347 & 0.196 & 0.637 & 0.550 & 0.665 & 0.709 \\
& A1 & 0.673 {\scriptsize \textcolor{green!60!black}{($\mathbf{+0.0}$)}} & 0.250 {\scriptsize \textcolor{green!60!black}{($\mathbf{-9.6}$)}} & 0.104 {\scriptsize \textcolor{green!60!black}{($\mathbf{-9.3}$)}} & 0.823 {\scriptsize \textcolor{red}{($\mathbf{+18.6}$)}} & 0.631 {\scriptsize \textcolor{red}{($\mathbf{+8.1}$)}} & 0.831 {\scriptsize \textcolor{red}{($\mathbf{+16.6}$)}} & 0.848 {\scriptsize \textcolor{red}{($\mathbf{+13.9}$)}} \\
& A2 & 0.551 {\scriptsize \textcolor{green!60!black}{($\mathbf{-12.2}$)}} & 0.347 {\scriptsize \textcolor{green!60!black}{($\mathbf{+0.0}$)}} & 0.089 {\scriptsize \textcolor{green!60!black}{($\mathbf{-10.8}$)}} & 0.807 {\scriptsize \textcolor{red}{($\mathbf{+17.1}$)}} & 0.661 {\scriptsize \textcolor{red}{($\mathbf{+11.1}$)}} & 0.817 {\scriptsize \textcolor{red}{($\mathbf{+15.2}$)}} & 0.841 {\scriptsize \textcolor{red}{($\mathbf{+13.2}$)}} \\
& A3 & 0.632 {\scriptsize \textcolor{green!60!black}{($\mathbf{-4.1}$)}} & 0.305 {\scriptsize \textcolor{green!60!black}{($\mathbf{-4.2}$)}} & 0.145 {\scriptsize \textcolor{green!60!black}{($\mathbf{-5.1}$)}} & 0.683 {\scriptsize \textcolor{red}{($\mathbf{+4.6}$)}} & 0.499 {\scriptsize \textcolor{green!60!black}{($\mathbf{-5.1}$)}} & 0.700 {\scriptsize \textcolor{red}{($\mathbf{+3.5}$)}} & 0.772 {\scriptsize \textcolor{red}{($\mathbf{+6.3}$)}} \\
\midrule
\multirow{4}{*}{Qwen3-VL-8B}
& A0 & 0.665 & 0.283 & 0.154 & 0.721 & 0.602 & 0.728 & 0.770 \\
& A1 & 0.665 {\scriptsize \textcolor{green!60!black}{($\mathbf{+0.0}$)}} & 0.193 {\scriptsize \textcolor{green!60!black}{($\mathbf{-9.0}$)}} & 0.059 {\scriptsize \textcolor{green!60!black}{($\mathbf{-9.5}$)}} & 0.900 {\scriptsize \textcolor{red}{($\mathbf{+17.9}$)}} & 0.686 {\scriptsize \textcolor{red}{($\mathbf{+8.4}$)}} & 0.903 {\scriptsize \textcolor{red}{($\mathbf{+17.4}$)}} & 0.913 {\scriptsize \textcolor{red}{($\mathbf{+14.3}$)}} \\
& A2 & 0.565 {\scriptsize \textcolor{green!60!black}{($\mathbf{-10.0}$)}} & 0.283 {\scriptsize \textcolor{green!60!black}{($\mathbf{+0.0}$)}} & 0.051 {\scriptsize \textcolor{green!60!black}{($\mathbf{-10.3}$)}} & 0.895 {\scriptsize \textcolor{red}{($\mathbf{+17.4}$)}} & 0.734 {\scriptsize \textcolor{red}{($\mathbf{+13.2}$)}} & 0.901 {\scriptsize \textcolor{red}{($\mathbf{+17.3}$)}} & 0.912 {\scriptsize \textcolor{red}{($\mathbf{+14.2}$)}} \\
& A3 & 0.613 {\scriptsize \textcolor{green!60!black}{($\mathbf{-5.2}$)}} & 0.259 {\scriptsize \textcolor{green!60!black}{($\mathbf{-2.4}$)}} & 0.077 {\scriptsize \textcolor{green!60!black}{($\mathbf{-7.7}$)}} & 0.833 {\scriptsize \textcolor{red}{($\mathbf{+11.2}$)}} & 0.617 {\scriptsize \textcolor{red}{($\mathbf{+1.5}$)}} & 0.836 {\scriptsize \textcolor{red}{($\mathbf{+10.8}$)}} & 0.875 {\scriptsize \textcolor{red}{($\mathbf{+10.5}$)}} \\
\midrule
\multirow{4}{*}{Qwen3-VL-2B}
& A0 & 0.495 & 0.333 & 0.102 & 0.720 & 0.528 & 0.749 & 0.794 \\
& A1 & 0.495 {\scriptsize \textcolor{green!60!black}{($\mathbf{+0.0}$)}} & 0.251 {\scriptsize \textcolor{green!60!black}{($\mathbf{-8.2}$)}} & 0.038 {\scriptsize \textcolor{green!60!black}{($\mathbf{-6.4}$)}} & 0.895 {\scriptsize \textcolor{red}{($\mathbf{+17.6}$)}} & 0.575 {\scriptsize \textcolor{red}{($\mathbf{+4.7}$)}} & 0.907 {\scriptsize \textcolor{red}{($\mathbf{+15.8}$)}} & 0.923 {\scriptsize \textcolor{red}{($\mathbf{+12.9}$)}} \\
& A2 & 0.447 {\scriptsize \textcolor{green!60!black}{($\mathbf{-4.9}$)}} & 0.333 {\scriptsize \textcolor{green!60!black}{($\mathbf{+0.0}$)}} & 0.043 {\scriptsize \textcolor{green!60!black}{($\mathbf{-5.9}$)}} & 0.836 {\scriptsize \textcolor{red}{($\mathbf{+11.6}$)}} & 0.595 {\scriptsize \textcolor{red}{($\mathbf{+6.6}$)}} & 0.871 {\scriptsize \textcolor{red}{($\mathbf{+12.3}$)}} & 0.904 {\scriptsize \textcolor{red}{($\mathbf{+11.0}$)}} \\
& A3 & 0.582 {\scriptsize \textcolor{red}{($\mathbf{+8.7}$)}} & 0.234 {\scriptsize \textcolor{green!60!black}{($\mathbf{-9.9}$)}} & 0.032 {\scriptsize \textcolor{green!60!black}{($\mathbf{-7.0}$)}} & 0.923 {\scriptsize \textcolor{red}{($\mathbf{+20.3}$)}} & 0.445 {\scriptsize \textcolor{green!60!black}{($\mathbf{-8.3}$)}} & 0.929 {\scriptsize \textcolor{red}{($\mathbf{+18.0}$)}} & 0.945 {\scriptsize \textcolor{red}{($\mathbf{+15.1}$)}} \\
\midrule
\multirow{4}{*}{SpatioLM}
& A0 & 0.511 & 0.235 & 0.079 & 0.754 & 0.623 & 0.793 & 0.848 \\
& A1 & 0.511 {\scriptsize \textcolor{green!60!black}{($\mathbf{+0.0}$)}} & 0.160 {\scriptsize \textcolor{green!60!black}{($\mathbf{-7.5}$)}} & 0.010 {\scriptsize \textcolor{green!60!black}{($\mathbf{-6.9}$)}} & 0.975 {\scriptsize \textcolor{red}{($\mathbf{+22.1}$)}} & 0.676 {\scriptsize \textcolor{red}{($\mathbf{+5.3}$)}} & 0.980 {\scriptsize \textcolor{red}{($\mathbf{+18.6}$)}} & 0.983 {\scriptsize \textcolor{red}{($\mathbf{+13.5}$)}} \\
& A2 & 0.422 {\scriptsize \textcolor{green!60!black}{($\mathbf{-8.9}$)}} & 0.235 {\scriptsize \textcolor{green!60!black}{($\mathbf{+0.0}$)}} & 0.013 {\scriptsize \textcolor{green!60!black}{($\mathbf{-6.6}$)}} & 0.964 {\scriptsize \textcolor{red}{($\mathbf{+21.0}$)}} & 0.742 {\scriptsize \textcolor{red}{($\mathbf{+11.9}$)}} & 0.968 {\scriptsize \textcolor{red}{($\mathbf{+17.4}$)}} & 0.973 {\scriptsize \textcolor{red}{($\mathbf{+12.5}$)}} \\
& A3 & 0.457 {\scriptsize \textcolor{green!60!black}{($\mathbf{-5.4}$)}} & 0.215 {\scriptsize \textcolor{green!60!black}{($\mathbf{-2.0}$)}} & 0.011 {\scriptsize \textcolor{green!60!black}{($\mathbf{-6.8}$)}} & 0.953 {\scriptsize \textcolor{red}{($\mathbf{+19.8}$)}} & 0.575 {\scriptsize \textcolor{green!60!black}{($\mathbf{-4.8}$)}} & 0.960 {\scriptsize \textcolor{red}{($\mathbf{+16.6}$)}} & 0.977 {\scriptsize \textcolor{red}{($\mathbf{+12.9}$)}} \\
\midrule
\multirow{4}{*}{InternVL3.5-8B}
& A0 & 0.523 & 0.284 & 0.081 & 0.781 & 0.639 & 0.818 & 0.847 \\
& A1 & 0.523 {\scriptsize \textcolor{green!60!black}{($\mathbf{+0.0}$)}} & 0.239 {\scriptsize \textcolor{green!60!black}{($\mathbf{-4.5}$)}} & 0.044 {\scriptsize \textcolor{green!60!black}{($\mathbf{-3.7}$)}} & 0.893 {\scriptsize \textcolor{red}{($\mathbf{+11.2}$)}} & 0.685 {\scriptsize \textcolor{red}{($\mathbf{+4.6}$)}} & 0.913 {\scriptsize \textcolor{red}{($\mathbf{+9.5}$)}} & 0.919 {\scriptsize \textcolor{red}{($\mathbf{+7.2}$)}} \\
& A2 & 0.473 {\scriptsize \textcolor{green!60!black}{($\mathbf{-5.0}$)}} & 0.284 {\scriptsize \textcolor{green!60!black}{($\mathbf{+0.0}$)}} & 0.033 {\scriptsize \textcolor{green!60!black}{($\mathbf{-4.9}$)}} & 0.914 {\scriptsize \textcolor{red}{($\mathbf{+13.2}$)}} & 0.697 {\scriptsize \textcolor{red}{($\mathbf{+5.8}$)}} & 0.926 {\scriptsize \textcolor{red}{($\mathbf{+10.8}$)}} & 0.934 {\scriptsize \textcolor{red}{($\mathbf{+8.7}$)}} \\
& A3 & 0.542 {\scriptsize \textcolor{red}{($\mathbf{+1.9}$)}} & 0.230 {\scriptsize \textcolor{green!60!black}{($\mathbf{-5.4}$)}} & 0.033 {\scriptsize \textcolor{green!60!black}{($\mathbf{-4.8}$)}} & 0.917 {\scriptsize \textcolor{red}{($\mathbf{+13.6}$)}} & 0.497 {\scriptsize \textcolor{green!60!black}{($\mathbf{-14.2}$)}} & 0.922 {\scriptsize \textcolor{red}{($\mathbf{+10.4}$)}} & 0.941 {\scriptsize \textcolor{red}{($\mathbf{+9.4}$)}} \\
\midrule
\multirow{4}{*}{SpatialRGPT-8B}
& A0 & 0.472 & 0.248 & 0.040 & 0.853 & 0.716 & 0.892 & 0.917 \\
& A1 & 0.472 {\scriptsize \textcolor{green!60!black}{($\mathbf{+0.0}$)}} & 0.192 {\scriptsize \textcolor{green!60!black}{($\mathbf{-5.6}$)}} & 0.003 {\scriptsize \textcolor{green!60!black}{($\mathbf{-3.7}$)}} & 0.993 {\scriptsize \textcolor{red}{($\mathbf{+14.0}$)}} & 0.729 {\scriptsize \textcolor{red}{($\mathbf{+1.4}$)}} & 0.995 {\scriptsize \textcolor{red}{($\mathbf{+10.3}$)}} & 0.995 {\scriptsize \textcolor{red}{($\mathbf{+7.8}$)}} \\
& A2 & 0.427 {\scriptsize \textcolor{green!60!black}{($\mathbf{-4.5}$)}} & 0.248 {\scriptsize \textcolor{green!60!black}{($\mathbf{+0.0}$)}} & 0.006 {\scriptsize \textcolor{green!60!black}{($\mathbf{-3.4}$)}} & 0.985 {\scriptsize \textcolor{red}{($\mathbf{+13.2}$)}} & 0.764 {\scriptsize \textcolor{red}{($\mathbf{+4.9}$)}} & 0.989 {\scriptsize \textcolor{red}{($\mathbf{+9.7}$)}} & 0.989 {\scriptsize \textcolor{red}{($\mathbf{+7.2}$)}} \\
& A3 & 0.548 {\scriptsize \textcolor{red}{($\mathbf{+7.6}$)}} & 0.199 {\scriptsize \textcolor{green!60!black}{($\mathbf{-4.9}$)}} & 0.031 {\scriptsize \textcolor{green!60!black}{($\mathbf{-0.9}$)}} & 0.934 {\scriptsize \textcolor{red}{($\mathbf{+8.2}$)}} & 0.434 {\scriptsize \textcolor{green!60!black}{($\mathbf{-28.2}$)}} & 0.940 {\scriptsize \textcolor{red}{($\mathbf{+4.8}$)}} & 0.946 {\scriptsize \textcolor{red}{($\mathbf{+2.9}$)}} \\
\bottomrule
\end{tabular}}
\end{table*}


\begin{table*}[t]
\centering
\caption{\textbf{Full ablation-B $\sigma$ sweep across the capability axis.} The target region (counterfactual box grown by $10\%$) stays sharp while the background is progressively blurred with Gaussian $\sigma$. $\sigma{=}0$ is the raw image (identical pixels to A0). Values in parentheses are percentage-point changes from the row marked $\sigma{=}0$ (raw image) for the same model; \textcolor{red}{red} = increase, \textcolor{green!60!black}{green} = decrease.}
\label{tab:app_abl_b_sigma}
\small
\resizebox{\textwidth}{!}{
\begin{tabular}{llccccccc}
\toprule
\textbf{Model} & \textbf{Cond.} & $\mathbf{Acc^{O}}\uparrow$ & $\mathbf{Acc^{CF}}\uparrow$ & $\mathbf{pAcc}\uparrow$ & $\mathbf{PRR}\downarrow$ & $\mathbf{LBR}\downarrow$ & $\mathbf{BR}\downarrow$ & $\mathbf{CER}\downarrow$ \\
\midrule
\multirow{5}{*}{Gemini 3.8 Flash}
& $\sigma{=}0$ (raw) & 0.756 & 0.490 & 0.372 & 0.448 & 0.357 & 0.450 & 0.509 \\
& $\sigma{=}4$ & 0.758 {\scriptsize \textcolor{red}{($\mathbf{+0.2}$)}} & 0.518 {\scriptsize \textcolor{red}{($\mathbf{+2.9}$)}} & 0.410 {\scriptsize \textcolor{red}{($\mathbf{+3.7}$)}} & 0.408 {\scriptsize \textcolor{green!60!black}{($\mathbf{-4.0}$)}} & 0.319 {\scriptsize \textcolor{green!60!black}{($\mathbf{-3.8}$)}} & 0.411 {\scriptsize \textcolor{green!60!black}{($\mathbf{-3.9}$)}} & 0.460 {\scriptsize \textcolor{green!60!black}{($\mathbf{-4.9}$)}} \\
& $\sigma{=}8$ & 0.772 {\scriptsize \textcolor{red}{($\mathbf{+1.6}$)}} & 0.525 {\scriptsize \textcolor{red}{($\mathbf{+3.5}$)}} & 0.419 {\scriptsize \textcolor{red}{($\mathbf{+4.7}$)}} & 0.411 {\scriptsize \textcolor{green!60!black}{($\mathbf{-3.8}$)}} & 0.330 {\scriptsize \textcolor{green!60!black}{($\mathbf{-2.7}$)}} & 0.418 {\scriptsize \textcolor{green!60!black}{($\mathbf{-3.3}$)}} & 0.457 {\scriptsize \textcolor{green!60!black}{($\mathbf{-5.2}$)}} \\
& $\sigma{=}12$ & 0.762 {\scriptsize \textcolor{red}{($\mathbf{+0.6}$)}} & 0.506 {\scriptsize \textcolor{red}{($\mathbf{+1.7}$)}} & 0.400 {\scriptsize \textcolor{red}{($\mathbf{+2.7}$)}} & 0.421 {\scriptsize \textcolor{green!60!black}{($\mathbf{-2.7}$)}} & 0.331 {\scriptsize \textcolor{green!60!black}{($\mathbf{-2.6}$)}} & 0.430 {\scriptsize \textcolor{green!60!black}{($\mathbf{-2.0}$)}} & 0.476 {\scriptsize \textcolor{green!60!black}{($\mathbf{-3.3}$)}} \\
& $\sigma{=}15$ & 0.759 {\scriptsize \textcolor{red}{($\mathbf{+0.3}$)}} & 0.506 {\scriptsize \textcolor{red}{($\mathbf{+1.7}$)}} & 0.403 {\scriptsize \textcolor{red}{($\mathbf{+3.0}$)}} & 0.415 {\scriptsize \textcolor{green!60!black}{($\mathbf{-3.4}$)}} & 0.340 {\scriptsize \textcolor{green!60!black}{($\mathbf{-1.7}$)}} & 0.422 {\scriptsize \textcolor{green!60!black}{($\mathbf{-2.9}$)}} & 0.468 {\scriptsize \textcolor{green!60!black}{($\mathbf{-4.1}$)}} \\
\midrule
\multirow{5}{*}{GPT-5.6 terra}
& $\sigma{=}0$ (raw) & 0.756 & 0.404 & 0.295 & 0.544 & 0.479 & 0.566 & 0.611 \\
& $\sigma{=}4$ & 0.742 {\scriptsize \textcolor{green!60!black}{($\mathbf{-1.4}$)}} & 0.440 {\scriptsize \textcolor{red}{($\mathbf{+3.6}$)}} & 0.299 {\scriptsize \textcolor{red}{($\mathbf{+0.4}$)}} & 0.542 {\scriptsize \textcolor{green!60!black}{($\mathbf{-0.2}$)}} & 0.481 {\scriptsize \textcolor{red}{($\mathbf{+0.1}$)}} & 0.555 {\scriptsize \textcolor{green!60!black}{($\mathbf{-1.1}$)}} & 0.596 {\scriptsize \textcolor{green!60!black}{($\mathbf{-1.4}$)}} \\
& $\sigma{=}8$ & 0.759 {\scriptsize \textcolor{red}{($\mathbf{+0.3}$)}} & 0.433 {\scriptsize \textcolor{red}{($\mathbf{+2.9}$)}} & 0.301 {\scriptsize \textcolor{red}{($\mathbf{+0.6}$)}} & 0.563 {\scriptsize \textcolor{red}{($\mathbf{+1.9}$)}} & 0.504 {\scriptsize \textcolor{red}{($\mathbf{+2.4}$)}} & 0.575 {\scriptsize \textcolor{red}{($\mathbf{+0.9}$)}} & 0.604 {\scriptsize \textcolor{green!60!black}{($\mathbf{-0.7}$)}} \\
& $\sigma{=}12$ & 0.756 {\scriptsize \textcolor{green!60!black}{($\mathbf{+0.0}$)}} & 0.433 {\scriptsize \textcolor{red}{($\mathbf{+2.9}$)}} & 0.297 {\scriptsize \textcolor{red}{($\mathbf{+0.2}$)}} & 0.562 {\scriptsize \textcolor{red}{($\mathbf{+1.8}$)}} & 0.513 {\scriptsize \textcolor{red}{($\mathbf{+3.4}$)}} & 0.576 {\scriptsize \textcolor{red}{($\mathbf{+1.0}$)}} & 0.607 {\scriptsize \textcolor{green!60!black}{($\mathbf{-0.3}$)}} \\
& $\sigma{=}15$ & 0.743 {\scriptsize \textcolor{green!60!black}{($\mathbf{-1.3}$)}} & 0.448 {\scriptsize \textcolor{red}{($\mathbf{+4.4}$)}} & 0.313 {\scriptsize \textcolor{red}{($\mathbf{+1.8}$)}} & 0.533 {\scriptsize \textcolor{green!60!black}{($\mathbf{-1.1}$)}} & 0.484 {\scriptsize \textcolor{red}{($\mathbf{+0.4}$)}} & 0.546 {\scriptsize \textcolor{green!60!black}{($\mathbf{-2.0}$)}} & 0.574 {\scriptsize \textcolor{green!60!black}{($\mathbf{-3.6}$)}} \\
\midrule
\multirow{5}{*}{Qwen3-VL-32B}
& $\sigma{=}0$ (raw) & 0.673 & 0.347 & 0.196 & 0.637 & 0.550 & 0.665 & 0.709 \\
& $\sigma{=}4$ & 0.681 {\scriptsize \textcolor{red}{($\mathbf{+0.8}$)}} & 0.384 {\scriptsize \textcolor{red}{($\mathbf{+3.7}$)}} & 0.238 {\scriptsize \textcolor{red}{($\mathbf{+4.1}$)}} & 0.590 {\scriptsize \textcolor{green!60!black}{($\mathbf{-4.6}$)}} & 0.508 {\scriptsize \textcolor{green!60!black}{($\mathbf{-4.2}$)}} & 0.610 {\scriptsize \textcolor{green!60!black}{($\mathbf{-5.5}$)}} & 0.653 {\scriptsize \textcolor{green!60!black}{($\mathbf{-5.6}$)}} \\
& $\sigma{=}8$ & 0.673 {\scriptsize \textcolor{green!60!black}{($\mathbf{+0.0}$)}} & 0.378 {\scriptsize \textcolor{red}{($\mathbf{+3.1}$)}} & 0.239 {\scriptsize \textcolor{red}{($\mathbf{+4.3}$)}} & 0.572 {\scriptsize \textcolor{green!60!black}{($\mathbf{-6.4}$)}} & 0.513 {\scriptsize \textcolor{green!60!black}{($\mathbf{-3.7}$)}} & 0.607 {\scriptsize \textcolor{green!60!black}{($\mathbf{-5.8}$)}} & 0.646 {\scriptsize \textcolor{green!60!black}{($\mathbf{-6.3}$)}} \\
& $\sigma{=}12$ & 0.657 {\scriptsize \textcolor{green!60!black}{($\mathbf{-1.6}$)}} & 0.382 {\scriptsize \textcolor{red}{($\mathbf{+3.5}$)}} & 0.237 {\scriptsize \textcolor{red}{($\mathbf{+4.1}$)}} & 0.566 {\scriptsize \textcolor{green!60!black}{($\mathbf{-7.1}$)}} & 0.495 {\scriptsize \textcolor{green!60!black}{($\mathbf{-5.5}$)}} & 0.591 {\scriptsize \textcolor{green!60!black}{($\mathbf{-7.4}$)}} & 0.641 {\scriptsize \textcolor{green!60!black}{($\mathbf{-6.8}$)}} \\
& $\sigma{=}15$ & 0.657 {\scriptsize \textcolor{green!60!black}{($\mathbf{-1.6}$)}} & 0.381 {\scriptsize \textcolor{red}{($\mathbf{+3.4}$)}} & 0.234 {\scriptsize \textcolor{red}{($\mathbf{+3.8}$)}} & 0.569 {\scriptsize \textcolor{green!60!black}{($\mathbf{-6.7}$)}} & 0.499 {\scriptsize \textcolor{green!60!black}{($\mathbf{-5.1}$)}} & 0.599 {\scriptsize \textcolor{green!60!black}{($\mathbf{-6.7}$)}} & 0.645 {\scriptsize \textcolor{green!60!black}{($\mathbf{-6.4}$)}} \\
\midrule
\multirow{5}{*}{Qwen3-VL-8B}
& $\sigma{=}0$ (raw) & 0.665 & 0.283 & 0.154 & 0.721 & 0.602 & 0.728 & 0.770 \\
& $\sigma{=}4$ & 0.660 {\scriptsize \textcolor{green!60!black}{($\mathbf{-0.5}$)}} & 0.313 {\scriptsize \textcolor{red}{($\mathbf{+3.0}$)}} & 0.191 {\scriptsize \textcolor{red}{($\mathbf{+3.7}$)}} & 0.655 {\scriptsize \textcolor{green!60!black}{($\mathbf{-6.6}$)}} & 0.579 {\scriptsize \textcolor{green!60!black}{($\mathbf{-2.3}$)}} & 0.666 {\scriptsize \textcolor{green!60!black}{($\mathbf{-6.3}$)}} & 0.712 {\scriptsize \textcolor{green!60!black}{($\mathbf{-5.9}$)}} \\
& $\sigma{=}8$ & 0.648 {\scriptsize \textcolor{green!60!black}{($\mathbf{-1.7}$)}} & 0.313 {\scriptsize \textcolor{red}{($\mathbf{+3.0}$)}} & 0.187 {\scriptsize \textcolor{red}{($\mathbf{+3.3}$)}} & 0.651 {\scriptsize \textcolor{green!60!black}{($\mathbf{-7.0}$)}} & 0.575 {\scriptsize \textcolor{green!60!black}{($\mathbf{-2.8}$)}} & 0.666 {\scriptsize \textcolor{green!60!black}{($\mathbf{-6.3}$)}} & 0.714 {\scriptsize \textcolor{green!60!black}{($\mathbf{-5.6}$)}} \\
& $\sigma{=}12$ & 0.647 {\scriptsize \textcolor{green!60!black}{($\mathbf{-1.8}$)}} & 0.317 {\scriptsize \textcolor{red}{($\mathbf{+3.4}$)}} & 0.193 {\scriptsize \textcolor{red}{($\mathbf{+3.9}$)}} & 0.639 {\scriptsize \textcolor{green!60!black}{($\mathbf{-8.2}$)}} & 0.571 {\scriptsize \textcolor{green!60!black}{($\mathbf{-3.1}$)}} & 0.654 {\scriptsize \textcolor{green!60!black}{($\mathbf{-7.4}$)}} & 0.703 {\scriptsize \textcolor{green!60!black}{($\mathbf{-6.7}$)}} \\
& $\sigma{=}15$ & 0.631 {\scriptsize \textcolor{green!60!black}{($\mathbf{-3.4}$)}} & 0.321 {\scriptsize \textcolor{red}{($\mathbf{+3.8}$)}} & 0.193 {\scriptsize \textcolor{red}{($\mathbf{+3.9}$)}} & 0.632 {\scriptsize \textcolor{green!60!black}{($\mathbf{-8.9}$)}} & 0.568 {\scriptsize \textcolor{green!60!black}{($\mathbf{-3.4}$)}} & 0.648 {\scriptsize \textcolor{green!60!black}{($\mathbf{-8.1}$)}} & 0.696 {\scriptsize \textcolor{green!60!black}{($\mathbf{-7.5}$)}} \\
\midrule
\multirow{5}{*}{Qwen3-VL-2B}
& $\sigma{=}0$ (raw) & 0.495 & 0.333 & 0.102 & 0.720 & 0.528 & 0.749 & 0.794 \\
& $\sigma{=}4$ & 0.487 {\scriptsize \textcolor{green!60!black}{($\mathbf{-0.8}$)}} & 0.360 {\scriptsize \textcolor{red}{($\mathbf{+2.7}$)}} & 0.134 {\scriptsize \textcolor{red}{($\mathbf{+3.2}$)}} & 0.643 {\scriptsize \textcolor{green!60!black}{($\mathbf{-7.7}$)}} & 0.519 {\scriptsize \textcolor{green!60!black}{($\mathbf{-0.9}$)}} & 0.673 {\scriptsize \textcolor{green!60!black}{($\mathbf{-7.6}$)}} & 0.725 {\scriptsize \textcolor{green!60!black}{($\mathbf{-6.9}$)}} \\
& $\sigma{=}8$ & 0.486 {\scriptsize \textcolor{green!60!black}{($\mathbf{-0.9}$)}} & 0.352 {\scriptsize \textcolor{red}{($\mathbf{+1.9}$)}} & 0.129 {\scriptsize \textcolor{red}{($\mathbf{+2.7}$)}} & 0.654 {\scriptsize \textcolor{green!60!black}{($\mathbf{-6.6}$)}} & 0.523 {\scriptsize \textcolor{green!60!black}{($\mathbf{-0.5}$)}} & 0.685 {\scriptsize \textcolor{green!60!black}{($\mathbf{-6.4}$)}} & 0.734 {\scriptsize \textcolor{green!60!black}{($\mathbf{-5.9}$)}} \\
& $\sigma{=}12$ & 0.487 {\scriptsize \textcolor{green!60!black}{($\mathbf{-0.8}$)}} & 0.341 {\scriptsize \textcolor{red}{($\mathbf{+0.8}$)}} & 0.115 {\scriptsize \textcolor{red}{($\mathbf{+1.3}$)}} & 0.674 {\scriptsize \textcolor{green!60!black}{($\mathbf{-4.6}$)}} & 0.545 {\scriptsize \textcolor{red}{($\mathbf{+1.7}$)}} & 0.715 {\scriptsize \textcolor{green!60!black}{($\mathbf{-3.4}$)}} & 0.764 {\scriptsize \textcolor{green!60!black}{($\mathbf{-3.0}$)}} \\
& $\sigma{=}15$ & 0.493 {\scriptsize \textcolor{green!60!black}{($\mathbf{-0.2}$)}} & 0.343 {\scriptsize \textcolor{red}{($\mathbf{+1.0}$)}} & 0.126 {\scriptsize \textcolor{red}{($\mathbf{+2.4}$)}} & 0.650 {\scriptsize \textcolor{green!60!black}{($\mathbf{-7.0}$)}} & 0.527 {\scriptsize \textcolor{green!60!black}{($\mathbf{-0.1}$)}} & 0.694 {\scriptsize \textcolor{green!60!black}{($\mathbf{-5.5}$)}} & 0.744 {\scriptsize \textcolor{green!60!black}{($\mathbf{-5.0}$)}} \\
\midrule
\multirow{5}{*}{SpatioLM}
& $\sigma{=}0$ (raw) & 0.511 & 0.235 & 0.079 & 0.754 & 0.623 & 0.793 & 0.848 \\
& $\sigma{=}4$ & 0.526 {\scriptsize \textcolor{red}{($\mathbf{+1.4}$)}} & 0.234 {\scriptsize \textcolor{green!60!black}{($\mathbf{-0.1}$)}} & 0.082 {\scriptsize \textcolor{red}{($\mathbf{+0.3}$)}} & 0.734 {\scriptsize \textcolor{green!60!black}{($\mathbf{-2.1}$)}} & 0.601 {\scriptsize \textcolor{green!60!black}{($\mathbf{-2.2}$)}} & 0.766 {\scriptsize \textcolor{green!60!black}{($\mathbf{-2.8}$)}} & 0.846 {\scriptsize \textcolor{green!60!black}{($\mathbf{-0.2}$)}} \\
& $\sigma{=}8$ & 0.516 {\scriptsize \textcolor{red}{($\mathbf{+0.5}$)}} & 0.238 {\scriptsize \textcolor{red}{($\mathbf{+0.2}$)}} & 0.080 {\scriptsize \textcolor{red}{($\mathbf{+0.1}$)}} & 0.733 {\scriptsize \textcolor{green!60!black}{($\mathbf{-2.1}$)}} & 0.598 {\scriptsize \textcolor{green!60!black}{($\mathbf{-2.4}$)}} & 0.766 {\scriptsize \textcolor{green!60!black}{($\mathbf{-2.8}$)}} & 0.848 {\scriptsize \textcolor{green!60!black}{($\mathbf{-0.0}$)}} \\
& $\sigma{=}12$ & 0.518 {\scriptsize \textcolor{red}{($\mathbf{+0.6}$)}} & 0.250 {\scriptsize \textcolor{red}{($\mathbf{+1.5}$)}} & 0.096 {\scriptsize \textcolor{red}{($\mathbf{+1.8}$)}} & 0.705 {\scriptsize \textcolor{green!60!black}{($\mathbf{-5.0}$)}} & 0.584 {\scriptsize \textcolor{green!60!black}{($\mathbf{-3.9}$)}} & 0.733 {\scriptsize \textcolor{green!60!black}{($\mathbf{-6.1}$)}} & 0.816 {\scriptsize \textcolor{green!60!black}{($\mathbf{-3.2}$)}} \\
& $\sigma{=}15$ & 0.513 {\scriptsize \textcolor{red}{($\mathbf{+0.2}$)}} & 0.250 {\scriptsize \textcolor{red}{($\mathbf{+1.4}$)}} & 0.096 {\scriptsize \textcolor{red}{($\mathbf{+1.7}$)}} & 0.694 {\scriptsize \textcolor{green!60!black}{($\mathbf{-6.0}$)}} & 0.588 {\scriptsize \textcolor{green!60!black}{($\mathbf{-3.5}$)}} & 0.727 {\scriptsize \textcolor{green!60!black}{($\mathbf{-6.6}$)}} & 0.816 {\scriptsize \textcolor{green!60!black}{($\mathbf{-3.2}$)}} \\
\midrule
\multirow{5}{*}{InternVL3.5-8B}
& $\sigma{=}0$ (raw) & 0.523 & 0.284 & 0.081 & 0.781 & 0.639 & 0.818 & 0.847 \\
& $\sigma{=}4$ & 0.557 {\scriptsize \textcolor{red}{($\mathbf{+3.3}$)}} & 0.313 {\scriptsize \textcolor{red}{($\mathbf{+2.9}$)}} & 0.119 {\scriptsize \textcolor{red}{($\mathbf{+3.7}$)}} & 0.713 {\scriptsize \textcolor{green!60!black}{($\mathbf{-6.9}$)}} & 0.589 {\scriptsize \textcolor{green!60!black}{($\mathbf{-5.0}$)}} & 0.757 {\scriptsize \textcolor{green!60!black}{($\mathbf{-6.1}$)}} & 0.789 {\scriptsize \textcolor{green!60!black}{($\mathbf{-5.8}$)}} \\
& $\sigma{=}8$ & 0.547 {\scriptsize \textcolor{red}{($\mathbf{+2.4}$)}} & 0.317 {\scriptsize \textcolor{red}{($\mathbf{+3.3}$)}} & 0.123 {\scriptsize \textcolor{red}{($\mathbf{+4.1}$)}} & 0.703 {\scriptsize \textcolor{green!60!black}{($\mathbf{-7.8}$)}} & 0.576 {\scriptsize \textcolor{green!60!black}{($\mathbf{-6.3}$)}} & 0.735 {\scriptsize \textcolor{green!60!black}{($\mathbf{-8.3}$)}} & 0.778 {\scriptsize \textcolor{green!60!black}{($\mathbf{-6.9}$)}} \\
& $\sigma{=}12$ & 0.544 {\scriptsize \textcolor{red}{($\mathbf{+2.1}$)}} & 0.322 {\scriptsize \textcolor{red}{($\mathbf{+3.8}$)}} & 0.125 {\scriptsize \textcolor{red}{($\mathbf{+4.4}$)}} & 0.694 {\scriptsize \textcolor{green!60!black}{($\mathbf{-8.7}$)}} & 0.574 {\scriptsize \textcolor{green!60!black}{($\mathbf{-6.6}$)}} & 0.731 {\scriptsize \textcolor{green!60!black}{($\mathbf{-8.7}$)}} & 0.772 {\scriptsize \textcolor{green!60!black}{($\mathbf{-7.5}$)}} \\
& $\sigma{=}15$ & 0.545 {\scriptsize \textcolor{red}{($\mathbf{+2.2}$)}} & 0.326 {\scriptsize \textcolor{red}{($\mathbf{+4.2}$)}} & 0.131 {\scriptsize \textcolor{red}{($\mathbf{+4.9}$)}} & 0.685 {\scriptsize \textcolor{green!60!black}{($\mathbf{-9.7}$)}} & 0.566 {\scriptsize \textcolor{green!60!black}{($\mathbf{-7.3}$)}} & 0.721 {\scriptsize \textcolor{green!60!black}{($\mathbf{-9.7}$)}} & 0.762 {\scriptsize \textcolor{green!60!black}{($\mathbf{-8.5}$)}} \\
\midrule
\multirow{5}{*}{SpatialRGPT-8B}
& $\sigma{=}0$ (raw) & 0.472 & 0.248 & 0.040 & 0.853 & 0.716 & 0.892 & 0.917 \\
& $\sigma{=}4$ & 0.469 {\scriptsize \textcolor{green!60!black}{($\mathbf{-0.3}$)}} & 0.265 {\scriptsize \textcolor{red}{($\mathbf{+1.7}$)}} & 0.056 {\scriptsize \textcolor{red}{($\mathbf{+1.6}$)}} & 0.801 {\scriptsize \textcolor{green!60!black}{($\mathbf{-5.2}$)}} & 0.707 {\scriptsize \textcolor{green!60!black}{($\mathbf{-0.9}$)}} & 0.845 {\scriptsize \textcolor{green!60!black}{($\mathbf{-4.7}$)}} & 0.882 {\scriptsize \textcolor{green!60!black}{($\mathbf{-3.5}$)}} \\
& $\sigma{=}8$ & 0.458 {\scriptsize \textcolor{green!60!black}{($\mathbf{-1.4}$)}} & 0.262 {\scriptsize \textcolor{red}{($\mathbf{+1.4}$)}} & 0.050 {\scriptsize \textcolor{red}{($\mathbf{+1.0}$)}} & 0.808 {\scriptsize \textcolor{green!60!black}{($\mathbf{-4.5}$)}} & 0.723 {\scriptsize \textcolor{red}{($\mathbf{+0.7}$)}} & 0.845 {\scriptsize \textcolor{green!60!black}{($\mathbf{-4.7}$)}} & 0.892 {\scriptsize \textcolor{green!60!black}{($\mathbf{-2.5}$)}} \\
& $\sigma{=}12$ & 0.446 {\scriptsize \textcolor{green!60!black}{($\mathbf{-2.6}$)}} & 0.260 {\scriptsize \textcolor{red}{($\mathbf{+1.2}$)}} & 0.050 {\scriptsize \textcolor{red}{($\mathbf{+1.0}$)}} & 0.797 {\scriptsize \textcolor{green!60!black}{($\mathbf{-5.5}$)}} & 0.742 {\scriptsize \textcolor{red}{($\mathbf{+2.6}$)}} & 0.844 {\scriptsize \textcolor{green!60!black}{($\mathbf{-4.8}$)}} & 0.889 {\scriptsize \textcolor{green!60!black}{($\mathbf{-2.8}$)}} \\
& $\sigma{=}15$ & 0.440 {\scriptsize \textcolor{green!60!black}{($\mathbf{-3.2}$)}} & 0.256 {\scriptsize \textcolor{red}{($\mathbf{+0.8}$)}} & 0.049 {\scriptsize \textcolor{red}{($\mathbf{+1.0}$)}} & 0.797 {\scriptsize \textcolor{green!60!black}{($\mathbf{-5.6}$)}} & 0.744 {\scriptsize \textcolor{red}{($\mathbf{+2.8}$)}} & 0.848 {\scriptsize \textcolor{green!60!black}{($\mathbf{-4.4}$)}} & 0.889 {\scriptsize \textcolor{green!60!black}{($\mathbf{-2.8}$)}} \\
\bottomrule
\end{tabular}}
\vspace{-0.5cm}
\end{table*}


\begin{table*}[t]
\centering
\caption{\textbf{Full ablation-B falsification and mechanism controls.} B2 blurs the whole image uniformly (no sharp target region); B3 leaves every pixel unchanged and only draws a red box around the target region (pure spatial hint). B0 is the raw-image baseline. Values in parentheses are percentage-point changes from the row marked $\sigma{=}0$ (raw image) for the same model; \textcolor{red}{red} = increase, \textcolor{green!60!black}{green} = decrease.}
\label{tab:app_abl_b_control}
\small
\resizebox{\textwidth}{!}{
\begin{tabular}{llccccccc}
\toprule
\textbf{Model} & \textbf{Cond.} & $\mathbf{Acc^{O}}\uparrow$ & $\mathbf{Acc^{CF}}\uparrow$ & $\mathbf{pAcc}\uparrow$ & $\mathbf{PRR}\downarrow$ & $\mathbf{LBR}\downarrow$ & $\mathbf{BR}\downarrow$ & $\mathbf{CER}\downarrow$ \\
\midrule
\multirow{3}{*}{Gemini 3.8 Flash}
& B0 (raw) & 0.756 & 0.490 & 0.372 & 0.448 & 0.357 & 0.450 & 0.509 \\
& B2 uniform $\sigma{=}8$ & 0.726 {\scriptsize \textcolor{green!60!black}{($\mathbf{-3.0}$)}} & 0.380 {\scriptsize \textcolor{green!60!black}{($\mathbf{-10.9}$)}} & 0.265 {\scriptsize \textcolor{green!60!black}{($\mathbf{-10.8}$)}} & 0.583 {\scriptsize \textcolor{red}{($\mathbf{+13.5}$)}} & 0.471 {\scriptsize \textcolor{red}{($\mathbf{+11.4}$)}} & 0.588 {\scriptsize \textcolor{red}{($\mathbf{+13.8}$)}} & 0.637 {\scriptsize \textcolor{red}{($\mathbf{+12.8}$)}} \\
& B3 bbox hint & 0.780 {\scriptsize \textcolor{red}{($\mathbf{+2.4}$)}} & 0.541 {\scriptsize \textcolor{red}{($\mathbf{+5.2}$)}} & 0.423 {\scriptsize \textcolor{red}{($\mathbf{+5.1}$)}} & 0.410 {\scriptsize \textcolor{green!60!black}{($\mathbf{-3.8}$)}} & 0.326 {\scriptsize \textcolor{green!60!black}{($\mathbf{-3.1}$)}} & 0.417 {\scriptsize \textcolor{green!60!black}{($\mathbf{-3.3}$)}} & 0.456 {\scriptsize \textcolor{green!60!black}{($\mathbf{-5.3}$)}} \\
\midrule
\multirow{3}{*}{GPT-5.6 terra}
& B0 (raw) & 0.756 & 0.404 & 0.295 & 0.544 & 0.479 & 0.566 & 0.611 \\
& B2 uniform $\sigma{=}8$ & 0.659 {\scriptsize \textcolor{green!60!black}{($\mathbf{-9.7}$)}} & 0.304 {\scriptsize \textcolor{green!60!black}{($\mathbf{-10.0}$)}} & 0.143 {\scriptsize \textcolor{green!60!black}{($\mathbf{-15.2}$)}} & 0.735 {\scriptsize \textcolor{red}{($\mathbf{+19.1}$)}} & 0.648 {\scriptsize \textcolor{red}{($\mathbf{+16.8}$)}} & 0.748 {\scriptsize \textcolor{red}{($\mathbf{+18.2}$)}} & 0.785 {\scriptsize \textcolor{red}{($\mathbf{+17.4}$)}} \\
& B3 bbox hint & 0.718 {\scriptsize \textcolor{green!60!black}{($\mathbf{-3.8}$)}} & 0.549 {\scriptsize \textcolor{red}{($\mathbf{+14.5}$)}} & 0.372 {\scriptsize \textcolor{red}{($\mathbf{+7.7}$)}} & 0.427 {\scriptsize \textcolor{green!60!black}{($\mathbf{-11.7}$)}} & 0.348 {\scriptsize \textcolor{green!60!black}{($\mathbf{-13.1}$)}} & 0.437 {\scriptsize \textcolor{green!60!black}{($\mathbf{-12.9}$)}} & 0.482 {\scriptsize \textcolor{green!60!black}{($\mathbf{-12.9}$)}} \\
\midrule
\multirow{3}{*}{Qwen3-VL-32B}
& B0 (raw) & 0.673 & 0.347 & 0.196 & 0.637 & 0.550 & 0.665 & 0.709 \\
& B2 uniform $\sigma{=}8$ & 0.612 {\scriptsize \textcolor{green!60!black}{($\mathbf{-6.1}$)}} & 0.230 {\scriptsize \textcolor{green!60!black}{($\mathbf{-11.6}$)}} & 0.092 {\scriptsize \textcolor{green!60!black}{($\mathbf{-10.4}$)}} & 0.782 {\scriptsize \textcolor{red}{($\mathbf{+14.5}$)}} & 0.680 {\scriptsize \textcolor{red}{($\mathbf{+13.0}$)}} & 0.812 {\scriptsize \textcolor{red}{($\mathbf{+14.7}$)}} & 0.852 {\scriptsize \textcolor{red}{($\mathbf{+14.3}$)}} \\
& B3 bbox hint & 0.701 {\scriptsize \textcolor{red}{($\mathbf{+2.8}$)}} & 0.423 {\scriptsize \textcolor{red}{($\mathbf{+7.7}$)}} & 0.267 {\scriptsize \textcolor{red}{($\mathbf{+7.1}$)}} & 0.540 {\scriptsize \textcolor{green!60!black}{($\mathbf{-9.6}$)}} & 0.478 {\scriptsize \textcolor{green!60!black}{($\mathbf{-7.2}$)}} & 0.578 {\scriptsize \textcolor{green!60!black}{($\mathbf{-8.7}$)}} & 0.620 {\scriptsize \textcolor{green!60!black}{($\mathbf{-8.9}$)}} \\
\midrule
\multirow{3}{*}{Qwen3-VL-8B}
& B0 (raw) & 0.665 & 0.283 & 0.154 & 0.721 & 0.602 & 0.728 & 0.770 \\
& B2 uniform $\sigma{=}8$ & 0.537 {\scriptsize \textcolor{green!60!black}{($\mathbf{-12.8}$)}} & 0.194 {\scriptsize \textcolor{green!60!black}{($\mathbf{-8.9}$)}} & 0.076 {\scriptsize \textcolor{green!60!black}{($\mathbf{-7.8}$)}} & 0.793 {\scriptsize \textcolor{red}{($\mathbf{+7.2}$)}} & 0.662 {\scriptsize \textcolor{red}{($\mathbf{+6.0}$)}} & 0.810 {\scriptsize \textcolor{red}{($\mathbf{+8.1}$)}} & 0.860 {\scriptsize \textcolor{red}{($\mathbf{+9.0}$)}} \\
& B3 bbox hint & 0.664 {\scriptsize \textcolor{green!60!black}{($\mathbf{-0.1}$)}} & 0.352 {\scriptsize \textcolor{red}{($\mathbf{+6.9}$)}} & 0.219 {\scriptsize \textcolor{red}{($\mathbf{+6.5}$)}} & 0.619 {\scriptsize \textcolor{green!60!black}{($\mathbf{-10.3}$)}} & 0.534 {\scriptsize \textcolor{green!60!black}{($\mathbf{-6.8}$)}} & 0.626 {\scriptsize \textcolor{green!60!black}{($\mathbf{-10.3}$)}} & 0.671 {\scriptsize \textcolor{green!60!black}{($\mathbf{-9.9}$)}} \\
\midrule
\multirow{3}{*}{Qwen3-VL-2B}
& B0 (raw) & 0.495 & 0.333 & 0.102 & 0.720 & 0.528 & 0.749 & 0.794 \\
& B2 uniform $\sigma{=}8$ & 0.415 {\scriptsize \textcolor{green!60!black}{($\mathbf{-8.0}$)}} & 0.283 {\scriptsize \textcolor{green!60!black}{($\mathbf{-5.0}$)}} & 0.038 {\scriptsize \textcolor{green!60!black}{($\mathbf{-6.4}$)}} & 0.831 {\scriptsize \textcolor{red}{($\mathbf{+11.1}$)}} & 0.637 {\scriptsize \textcolor{red}{($\mathbf{+10.9}$)}} & 0.862 {\scriptsize \textcolor{red}{($\mathbf{+11.3}$)}} & 0.908 {\scriptsize \textcolor{red}{($\mathbf{+11.4}$)}} \\
& B3 bbox hint & 0.485 {\scriptsize \textcolor{green!60!black}{($\mathbf{-1.0}$)}} & 0.383 {\scriptsize \textcolor{red}{($\mathbf{+4.9}$)}} & 0.138 {\scriptsize \textcolor{red}{($\mathbf{+3.6}$)}} & 0.632 {\scriptsize \textcolor{green!60!black}{($\mathbf{-8.8}$)}} & 0.507 {\scriptsize \textcolor{green!60!black}{($\mathbf{-2.2}$)}} & 0.669 {\scriptsize \textcolor{green!60!black}{($\mathbf{-7.9}$)}} & 0.715 {\scriptsize \textcolor{green!60!black}{($\mathbf{-7.8}$)}} \\
\midrule
\multirow{3}{*}{SpatioLM}
& B0 (raw) & 0.511 & 0.235 & 0.079 & 0.754 & 0.623 & 0.793 & 0.848 \\
& B2 uniform $\sigma{=}8$ & 0.482 {\scriptsize \textcolor{green!60!black}{($\mathbf{-3.0}$)}} & 0.194 {\scriptsize \textcolor{green!60!black}{($\mathbf{-4.1}$)}} & 0.041 {\scriptsize \textcolor{green!60!black}{($\mathbf{-3.7}$)}} & 0.832 {\scriptsize \textcolor{red}{($\mathbf{+7.8}$)}} & 0.669 {\scriptsize \textcolor{red}{($\mathbf{+4.7}$)}} & 0.849 {\scriptsize \textcolor{red}{($\mathbf{+5.5}$)}} & 0.917 {\scriptsize \textcolor{red}{($\mathbf{+6.9}$)}} \\
& B3 bbox hint & 0.484 {\scriptsize \textcolor{green!60!black}{($\mathbf{-2.7}$)}} & 0.238 {\scriptsize \textcolor{red}{($\mathbf{+0.3}$)}} & 0.084 {\scriptsize \textcolor{red}{($\mathbf{+0.5}$)}} & 0.721 {\scriptsize \textcolor{green!60!black}{($\mathbf{-3.3}$)}} & 0.607 {\scriptsize \textcolor{green!60!black}{($\mathbf{-1.6}$)}} & 0.752 {\scriptsize \textcolor{green!60!black}{($\mathbf{-4.1}$)}} & 0.828 {\scriptsize \textcolor{green!60!black}{($\mathbf{-2.0}$)}} \\
\midrule
\multirow{3}{*}{InternVL3.5-8B}
& B0 (raw) & 0.523 & 0.284 & 0.081 & 0.781 & 0.639 & 0.818 & 0.847 \\
& B2 uniform $\sigma{=}8$ & 0.478 {\scriptsize \textcolor{green!60!black}{($\mathbf{-4.5}$)}} & 0.234 {\scriptsize \textcolor{green!60!black}{($\mathbf{-4.9}$)}} & 0.038 {\scriptsize \textcolor{green!60!black}{($\mathbf{-4.3}$)}} & 0.869 {\scriptsize \textcolor{red}{($\mathbf{+8.8}$)}} & 0.705 {\scriptsize \textcolor{red}{($\mathbf{+6.6}$)}} & 0.888 {\scriptsize \textcolor{red}{($\mathbf{+7.0}$)}} & 0.923 {\scriptsize \textcolor{red}{($\mathbf{+7.6}$)}} \\
& B3 bbox hint & 0.503 {\scriptsize \textcolor{green!60!black}{($\mathbf{-2.0}$)}} & 0.329 {\scriptsize \textcolor{red}{($\mathbf{+4.5}$)}} & 0.116 {\scriptsize \textcolor{red}{($\mathbf{+3.4}$)}} & 0.693 {\scriptsize \textcolor{green!60!black}{($\mathbf{-8.8}$)}} & 0.618 {\scriptsize \textcolor{green!60!black}{($\mathbf{-2.1}$)}} & 0.733 {\scriptsize \textcolor{green!60!black}{($\mathbf{-8.5}$)}} & 0.773 {\scriptsize \textcolor{green!60!black}{($\mathbf{-7.4}$)}} \\
\midrule
\multirow{3}{*}{SpatialRGPT-8B}
& B0 (raw) & 0.472 & 0.248 & 0.040 & 0.853 & 0.716 & 0.892 & 0.917 \\
& B2 uniform $\sigma{=}8$ & 0.432 {\scriptsize \textcolor{green!60!black}{($\mathbf{-4.0}$)}} & 0.212 {\scriptsize \textcolor{green!60!black}{($\mathbf{-3.6}$)}} & 0.013 {\scriptsize \textcolor{green!60!black}{($\mathbf{-2.7}$)}} & 0.941 {\scriptsize \textcolor{red}{($\mathbf{+8.8}$)}} & 0.793 {\scriptsize \textcolor{red}{($\mathbf{+7.7}$)}} & 0.959 {\scriptsize \textcolor{red}{($\mathbf{+6.8}$)}} & 0.972 {\scriptsize \textcolor{red}{($\mathbf{+5.5}$)}} \\
& B3 bbox hint & 0.466 {\scriptsize \textcolor{green!60!black}{($\mathbf{-0.6}$)}} & 0.275 {\scriptsize \textcolor{red}{($\mathbf{+2.7}$)}} & 0.057 {\scriptsize \textcolor{red}{($\mathbf{+1.7}$)}} & 0.796 {\scriptsize \textcolor{green!60!black}{($\mathbf{-5.7}$)}} & 0.703 {\scriptsize \textcolor{green!60!black}{($\mathbf{-1.2}$)}} & 0.844 {\scriptsize \textcolor{green!60!black}{($\mathbf{-4.8}$)}} & 0.880 {\scriptsize \textcolor{green!60!black}{($\mathbf{-3.7}$)}} \\
\bottomrule
\end{tabular}}
\end{table*}

{
\color{black}
\subsubsection{Robustness to Answer-Option Ordering}
\label{app:option_order}

To examine whether answer-option ordering confounds bias evaluation, we conduct a controlled permutation study on all 1,254 original--counterfactual pairs. We evaluate four cyclic rotations of the answer options ($k\in\{0,1,2,3\}$) for three representative models (Qwen3-VL-8B, SpatioLM-Understanding, and GPT-5.6 Terra), with $k=0$ denoting the original ordering. Each rotation preserves the question, images, and option content while consistently remapping the option labels and ground-truth answers. The same rotation is applied to original-image, counterfactual-image, and text-only evaluations, ensuring consistent label comparisons for bias computation.

As shown in Tab.~\ref{tab:app_option_order}, the bias rate (BR) remains relatively stable across rotations, with maximum deviations from the original ordering of 3.1, 1.4, and 0.7 percentage points (pp) for Qwen3-VL-8B, SpatioLM-Understanding, and GPT-5.6 Terra, respectively. Other bias metrics exhibit metric-dependent variations. In contrast, accuracy shows greater sensitivity to option ordering for certain models. Notably, GPT-5.6 Terra exhibits maximum reductions of 9.1 pp in $Acc^O$ and 5.0 pp in $pAcc$, indicating that option ordering can affect prediction accuracy despite the relative stability of aggregate bias estimates.

To further assess sample-level robustness, Tab.~\ref{tab:app_option_consistency} reports all-rotation correctness and content agreement separately for original (O) and counterfactual (CF) questions. All-rotation correctness, defined over all 1,254 questions per side, ranges from 27.83--53.35\% for O and 6.86--17.38\% for CF. Content agreement, measured among questions with valid predictions under all four rotations, ranges from 40.75--69.94\% for O and 33.49--64.04\% for CF. The higher content agreement indicates that some predictions remain consistent across rotations even when incorrect. These results reveal prediction-level sensitivity, particularly for CF questions, despite relatively stable aggregate BR estimates.

Overall, while option ordering affects individual predictions and certain accuracy metrics, the limited variation in BR suggests that the observed aggregate spatial bias cannot be explained solely by option-order effects.

\begin{table*}[t]
\centering
\caption{\textcolor{black}{\textbf{Robustness to answer-option ordering.}
Accuracy and spatial bias metrics across four cyclic rotations of the answer options on all original--counterfactual pairs. $k=0$ denotes the original ordering. Parenthesized values indicate percentage-point (pp) changes relative to $k=0$ for each model (\textcolor{red}{red} = increase, \textcolor{green!60!black}{green} = decrease). 
}}
\label{tab:app_option_order}
\small
\resizebox{\textwidth}{!}{
\begin{tabular}{lcccccccc}
\toprule
\textbf{Model} & \textbf{Perm.} & $\mathbf{Acc^{O}}\uparrow$ & $\mathbf{Acc^{CF}}\uparrow$ & $\mathbf{pAcc}\uparrow$ & $\mathbf{PRR}\downarrow$ & $\mathbf{LBR}\downarrow$ & $\mathbf{BR}\downarrow$ & $\mathbf{CER}\downarrow$ \\
\midrule
\multirow{4}{*}{Qwen3-VL-8B} & k=0 & 0.665 & 0.283 & 0.154 & 0.721 & 0.602 & 0.728 & 0.770 \\
 & k=1 & 0.664 {\scriptsize \textcolor{green!60!black}{($\mathbf{-0.1}$)}} & 0.286 {\scriptsize \textcolor{red}{($\mathbf{+0.3}$)}} & 0.148 {\scriptsize \textcolor{green!60!black}{($\mathbf{-0.6}$)}} & 0.721 {\scriptsize \textcolor{green!60!black}{($\mathbf{+0.0}$)}} & 0.620 {\scriptsize \textcolor{red}{($\mathbf{+1.8}$)}} & 0.740 {\scriptsize \textcolor{red}{($\mathbf{+1.2}$)}} & 0.779 {\scriptsize \textcolor{red}{($\mathbf{+0.8}$)}} \\
 & k=2 & 0.673 {\scriptsize \textcolor{red}{($\mathbf{+0.8}$)}} & 0.274 {\scriptsize \textcolor{green!60!black}{($\mathbf{-1.0}$)}} & 0.138 {\scriptsize \textcolor{green!60!black}{($\mathbf{-1.6}$)}} & 0.739 {\scriptsize \textcolor{red}{($\mathbf{+1.8}$)}} & 0.638 {\scriptsize \textcolor{red}{($\mathbf{+3.6}$)}} & 0.759 {\scriptsize \textcolor{red}{($\mathbf{+3.1}$)}} & 0.796 {\scriptsize \textcolor{red}{($\mathbf{+2.6}$)}} \\
 & k=3 & 0.672 {\scriptsize \textcolor{red}{($\mathbf{+0.7}$)}} & 0.282 {\scriptsize \textcolor{green!60!black}{($\mathbf{-0.1}$)}} & 0.144 {\scriptsize \textcolor{green!60!black}{($\mathbf{-1.0}$)}} & 0.728 {\scriptsize \textcolor{red}{($\mathbf{+0.7}$)}} & 0.628 {\scriptsize \textcolor{red}{($\mathbf{+2.6}$)}} & 0.743 {\scriptsize \textcolor{red}{($\mathbf{+1.5}$)}} & 0.787 {\scriptsize \textcolor{red}{($\mathbf{+1.7}$)}} \\
 \midrule
\multirow{4}{*}{SpatioLM-Understanding} & k=0 & 0.511 & 0.235 & 0.079 & 0.754 & 0.623 & 0.793 & 0.848 \\
 & k=1 & 0.520 {\scriptsize \textcolor{red}{($\mathbf{+0.9}$)}} & 0.243 {\scriptsize \textcolor{red}{($\mathbf{+0.8}$)}} & 0.079 {\scriptsize \textcolor{green!60!black}{($\mathbf{+0.0}$)}} & 0.753 {\scriptsize \textcolor{green!60!black}{($\mathbf{-0.2}$)}} & 0.605 {\scriptsize \textcolor{green!60!black}{($\mathbf{-1.8}$)}} & 0.785 {\scriptsize \textcolor{green!60!black}{($\mathbf{-0.8}$)}} & 0.849 {\scriptsize \textcolor{red}{($\mathbf{+0.1}$)}} \\
 & k=2 & 0.502 {\scriptsize \textcolor{green!60!black}{($\mathbf{-1.0}$)}} & 0.257 {\scriptsize \textcolor{red}{($\mathbf{+2.2}$)}} & 0.080 {\scriptsize \textcolor{red}{($\mathbf{+0.1}$)}} & 0.750 {\scriptsize \textcolor{green!60!black}{($\mathbf{-0.5}$)}} & 0.614 {\scriptsize \textcolor{green!60!black}{($\mathbf{-0.9}$)}} & 0.780 {\scriptsize \textcolor{green!60!black}{($\mathbf{-1.4}$)}} & 0.842 {\scriptsize \textcolor{green!60!black}{($\mathbf{-0.6}$)}} \\
 & k=3 & 0.501 {\scriptsize \textcolor{green!60!black}{($\mathbf{-1.0}$)}} & 0.226 {\scriptsize \textcolor{green!60!black}{($\mathbf{-1.0}$)}} & 0.068 {\scriptsize \textcolor{green!60!black}{($\mathbf{-1.1}$)}} & 0.770 {\scriptsize \textcolor{red}{($\mathbf{+1.6}$)}} & 0.633 {\scriptsize \textcolor{red}{($\mathbf{+1.0}$)}} & 0.796 {\scriptsize \textcolor{red}{($\mathbf{+0.2}$)}} & 0.866 {\scriptsize \textcolor{red}{($\mathbf{+1.8}$)}} \\
 \midrule
\multirow{4}{*}{GPT-5.6 Terra} & k=0 & 0.756 & 0.404 & 0.295 & 0.544 & 0.479 & 0.566 & 0.611 \\
 & k=1 & 0.665 {\scriptsize \textcolor{green!60!black}{($\mathbf{-9.1}$)}} & 0.420 {\scriptsize \textcolor{red}{($\mathbf{+1.6}$)}} & 0.245 {\scriptsize \textcolor{green!60!black}{($\mathbf{-5.0}$)}} & 0.546 {\scriptsize \textcolor{red}{($\mathbf{+0.2}$)}} & 0.445 {\scriptsize \textcolor{green!60!black}{($\mathbf{-3.4}$)}} & 0.569 {\scriptsize \textcolor{red}{($\mathbf{+0.3}$)}} & 0.633 {\scriptsize \textcolor{red}{($\mathbf{+2.2}$)}} \\
 & k=2 & 0.682 {\scriptsize \textcolor{green!60!black}{($\mathbf{-7.4}$)}} & 0.415 {\scriptsize \textcolor{red}{($\mathbf{+1.0}$)}} & 0.252 {\scriptsize \textcolor{green!60!black}{($\mathbf{-4.3}$)}} & 0.552 {\scriptsize \textcolor{red}{($\mathbf{+0.8}$)}} & 0.451 {\scriptsize \textcolor{green!60!black}{($\mathbf{-2.9}$)}} & 0.569 {\scriptsize \textcolor{red}{($\mathbf{+0.3}$)}} & 0.630 {\scriptsize \textcolor{red}{($\mathbf{+2.0}$)}} \\
 & k=3 & 0.667 {\scriptsize \textcolor{green!60!black}{($\mathbf{-8.9}$)}} & 0.416 {\scriptsize \textcolor{red}{($\mathbf{+1.2}$)}} & 0.246 {\scriptsize \textcolor{green!60!black}{($\mathbf{-4.9}$)}} & 0.550 {\scriptsize \textcolor{red}{($\mathbf{+0.6}$)}} & 0.460 {\scriptsize \textcolor{green!60!black}{($\mathbf{-2.0}$)}} & 0.573 {\scriptsize \textcolor{red}{($\mathbf{+0.7}$)}} & 0.631 {\scriptsize \textcolor{red}{($\mathbf{+2.0}$)}} \\
\bottomrule
\end{tabular}
}
\end{table*}

\begin{table}[t]
\centering
\caption{\textcolor{black}{\textbf{Prediction robustness across answer-option rotations.}
All-correct denotes the percentage of questions answered correctly under all four rotations. Content agreement denotes the percentage of questions receiving identical option-content predictions across all four rotations. O and CF denote original and counterfactual questions, respectively.}}
\label{tab:app_option_consistency}
\small
\begin{tabular}{lcccc}
\toprule
\multirow{2}{*}{\textbf{Model}} &
\multicolumn{2}{c}{\textbf{All-correct (\%)}} &
\multicolumn{2}{c}{\textbf{Content agreement (\%)}} \\
\cmidrule(lr){2-3}
\cmidrule(lr){4-5}
& \textbf{O} & \textbf{CF} & \textbf{O} & \textbf{CF} \\
\midrule
Qwen3-VL-8B
& 53.35 & 16.75 & 69.94 & 64.04 \\
SpatioLM-Understanding
& 27.83 & 6.86 & 40.75 & 33.49 \\
GPT-5.6 Terra
& 46.33 & 17.38 & 52.72 & 41.99 \\
\bottomrule
\end{tabular}
\end{table}

}

\begin{figure}[t]
    \centering
    \includegraphics[width=1\linewidth]{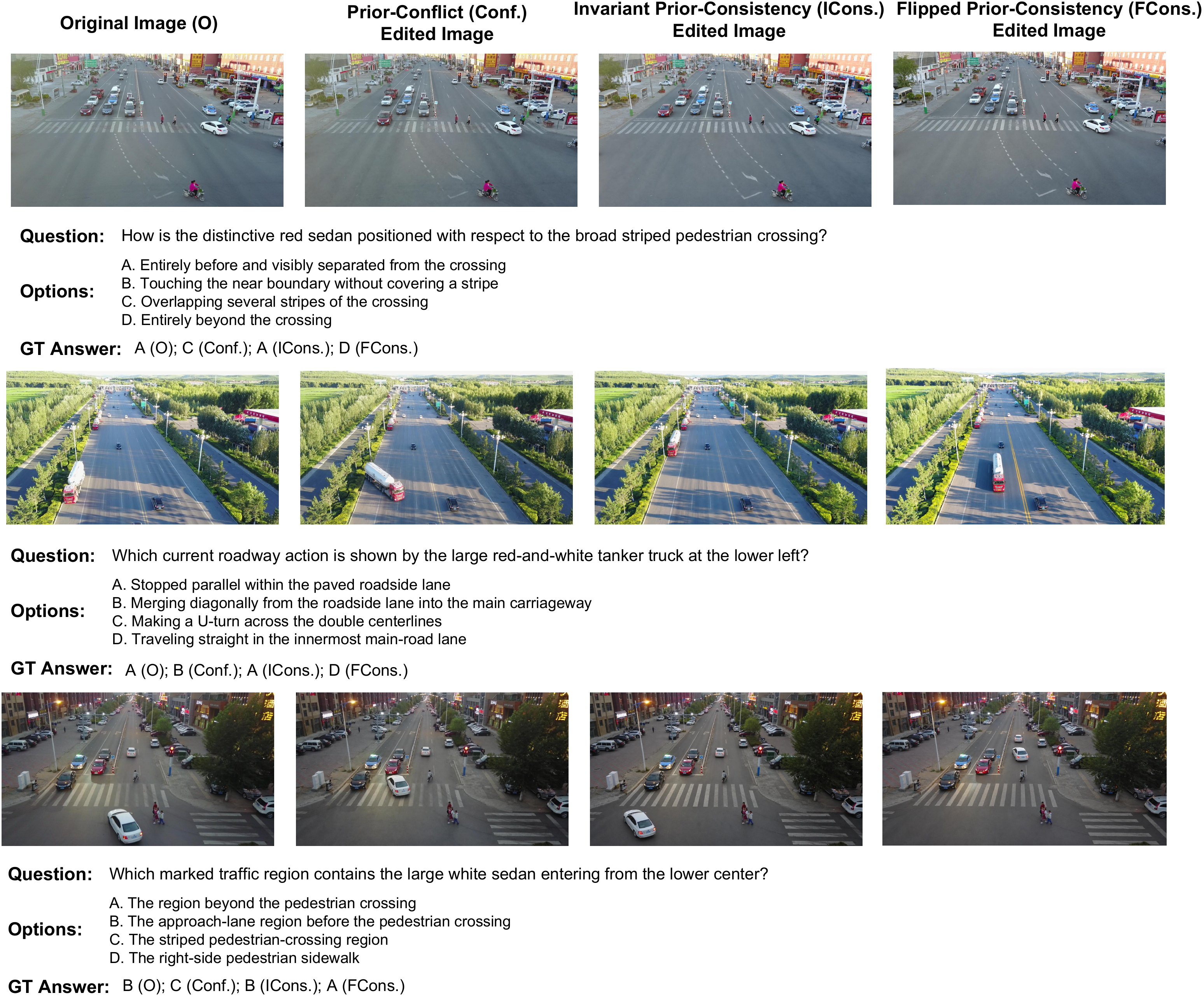}
\caption{\textcolor{black}{Examples of original (O), invariant prior-consistent (ICons.), flipped prior-consistent (FCons.), and prior-conflicting (Conf.) edits, together with their corresponding questions, answer options, and ground-truth answers. ICons. edits preserve the original ground-truth answer, whereas both FCons. and Conf. edits require a ground-truth answer change.}}
    \label{fig:app_pc_examples}
    \vspace{-0.8cm}
\end{figure}

{
\color{black}
\subsubsection{Prior-Consistent vs. Prior-Conflicting Editing}
\label{app:pc_control}

\providecommand{\pcNPairs}{341}
\providecommand{\pcNPool}{354}
\providecommand{\pcNExcluded}{13}
\providecommand{\pcNModels}{3}
\providecommand{\pcAccOMin}{0.499}\providecommand{\pcAccOMax}{0.757}
\providecommand{\pcAccPCMin}{0.507}\providecommand{\pcAccPCMax}{0.818}
\providecommand{\pcAccCFMin}{0.196}\providecommand{\pcAccCFMax}{0.413}
\providecommand{\pcDeltaPCMin}{+0.9}\providecommand{\pcDeltaPCMax}{+6.2}
\providecommand{\pcDeltaCFMin}{-40.8}\providecommand{\pcDeltaCFMax}{-30.2}
\providecommand{\pcRetPCMin}{75.7}\providecommand{\pcRetPCMax}{90.0}
\providecommand{\pcRetCFMin}{54.3}\providecommand{\pcRetCFMax}{78.3}
\providecommand{\pcEligMin}{170}\providecommand{\pcEligMax}{258}
\providecommand{\pcFPCMin}{3.0}\providecommand{\pcFPCMax}{9.4}
\providecommand{\pcBRMin}{60.1}\providecommand{\pcBRMax}{83.5}
\providecommand{\pcRatioMin}{6.5}\providecommand{\pcRatioMax}{25.0}
\providecommand{\pcDiscMin}{130}\providecommand{\pcDiscMax}{172}
\providecommand{\pcReverseMax}{14}
\providecommand{\pcPCDestroysMin}{7}\providecommand{\pcPCDestroysMax}{24}
\providecommand{\pcPCFixesMin}{19}\providecommand{\pcPCFixesMax}{45}
\providecommand{\pcMcNemarPMax}{1.3\times10^{-28}}
\providecommand{\pcMcNemarPMin}{8.3\times10^{-46}}
\providecommand{\pcPVsOMax}{7.4\times10^{-1}}
\providecommand{\pcPVsOMin}{1.3\times10^{-3}}
\providecommand{\pcPVsCFMax}{5.9\times10^{-15}}
\providecommand{\pcNSigPC}{2}
\providecommand{\pcNNonNegative}{3}
\providecommand{\pcCatFPCMax}{14.8}
\providecommand{\pcCatBRMin}{32.1}
\providecommand{\pcCrossRecoverN}{61}
\providecommand{\pcCrossReverseN}{0}
\providecommand{\pcCaseRecoverId}{\texttt{DSCN_039}}
\providecommand{\pcEditChangedPC}{53.9}
\providecommand{\pcEditChangedCF}{30.8}
\providecommand{\pcEditMeanAbsPC}{11.4}
\providecommand{\pcEditMeanAbsCF}{7.8}
\providecommand{\pcEditPCGtCF}{320}
\providecommand{\pcEditSignP}{7.7\times10^{-70}}
\providecommand{\pcEditRatioMedian}{1.64}
\providecommand{\pcMainBRMin}{56.6}\providecommand{\pcMainBRMax}{79.3}
\providecommand{\pcMainAccOMin}{0.511}\providecommand{\pcMainAccOMax}{0.756}
\providecommand{\pcMainBRGapMin}{+1.6}\providecommand{\pcMainBRGapMax}{+4.2}
\providecommand{\pcBRQFirstMin}{45.3}\providecommand{\pcBRQFirstMax}{64.3}
\providecommand{\pcBRQLastMin}{56.9}\providecommand{\pcBRQLastMax}{93.0}
\providecommand{\pcFPCQFirstMin}{4.7}\providecommand{\pcFPCQFirstMax}{11.9}
\providecommand{\pcFPCQLastMin}{1.7}\providecommand{\pcFPCQLastMax}{12.3}
\providecommand{\pcFPCQSwingMax}{9.2}
\providecommand{\pcBRQSwingMax}{33.3}

\begin{table*}[t]
\centering
\caption{\textcolor{black}{
\textbf{Prior-consistent (Cons.) versus prior-conflicting (Conf.) editing.}
Accuracy on original (O), invariant prior-consistent (ICons.) edits preserving the original ground-truth answer, flipped prior-consistent (FCons.) edits changing the ground-truth answer, and prior-conflicting (Conf.) edits across 170 paired samples.
$\Delta_{\mathrm{FCons-Conf}}$ and $\Delta_{\mathrm{ICons-Conf}}$ denote the accuracy differences between FCons. and Conf., and between ICons. and Conf., respectively, reported in percentage points (pp).
}}
\label{tab:app_pc}
\small
\resizebox{\textwidth}{!}{
\begin{tabular}{lcccccc}
\toprule
\textbf{Model}
& $\mathbf{Acc^{O}}\uparrow$
& $\mathbf{Acc^{ICons}}\uparrow$
& $\mathbf{Acc^{FCons}}\uparrow$
& $\mathbf{Acc^{Conf}}\uparrow$
& $\mathbf{\Delta_{\mathrm{FCons-Conf}}}$ (pp)
& $\mathbf{\Delta_{\mathrm{ICons-Conf}}}$ (pp) \\
\midrule
\multicolumn{7}{l}{\textit{Closed-source models}} \\
\midrule
Claude Opus-5
& 0.659 & 0.788 & 0.412 & 0.312
& +10.0 & +47.6 \\

Claude Sonnet-5
& 0.706 & 0.794 & 0.371 & 0.300
& +7.1 & +49.4 \\

\midrule
\multicolumn{7}{l}{\textit{Open-source general-purpose models}} \\
\midrule
Qwen3-VL-2B-Instruct
& 0.624 & 0.647 & 0.347 & 0.265
& +8.2 & +38.2 \\

Qwen3-VL-8B-Instruct
& 0.735 & 0.800 & 0.453 & 0.324
& +12.9 & +47.6 \\

Qwen3-VL-32B-Instruct
& 0.653 & 0.735 & 0.665 & 0.376
& +28.9 & +35.9 \\

\midrule
\multicolumn{7}{l}{\textit{Spatial-specialist models}} \\
\midrule
SpatioLM-InternVL3.5
& 0.571 & 0.582 & 0.429 & 0.241
& +18.8 & +34.1 \\

SpatialRGPT-VILA1.5-8B
& 0.488 & 0.541 & 0.376 & 0.224
& +15.2 & +31.7 \\
\bottomrule
\end{tabular}
}
\end{table*}

To investigate whether performance degradation arises from spatial prior conflict rather than visual editing alone, we construct three types of edits on a subset of 170 original images: (1) invariant prior-consistent (ICons.) edits, which preserve the conventional spatial relation and the original ground-truth answer; (2) flipped prior-consistent (FCons.) edits, which change the ground-truth answer while maintaining consistency with common-sense spatial priors; and (3) prior-conflicting (Conf.) edits, which change the ground-truth answer in a direction that contradicts the corresponding spatial prior. All three editing conditions modify the same target object while keeping the question and answer options unchanged. 
This design enables a direct comparison between prior-consistent and prior-conflicting edits under matched answer-change requirements. Representative examples are shown in Fig.~\ref{fig:app_pc_examples}. We evaluate seven representative models across different model families under identical prompting conditions.

As shown in Tab.~\ref{tab:app_pc}, both ICons. and FCons. consistently achieve higher accuracy than Conf. across all seven models, indicating that spatial prior conflict contributes to errors beyond the effects associated with image editing and answer changes. Meanwhile, FCons. generally yields lower accuracy than ICons., suggesting that image editing and the resulting changes in ground-truth answers introduce additional challenges, even when the edited spatial relations remain consistent with common-sense priors. The accuracy gap between FCons. and Conf. further characterizes model sensitivity to prior conflict under matched answer-change requirements, revealing notable differences across model families.

Among closed-source models, Claude Opus-5 and Claude Sonnet-5 exhibit relatively small FCons.–Conf. gaps of 10.0 and 7.1 pp, respectively, suggesting greater robustness to prior-conflicting edits. In contrast, the Qwen3-VL series shows an increasing gap with model scale, from 8.2 pp for 2B to 12.9 pp for 8B and 28.9 pp for 32B, indicating stronger sensitivity to conflicting spatial priors in larger models. Spatial-specialist models also exhibit noticeable prior-conflict effects, with gaps of 18.8 pp for SpatioLM-InternVL3.5 and 15.2 pp for SpatialRGPT-VILA1.5-8B. These findings suggest that spatial prior sensitivity varies across model families and scales, and that specialized spatial reasoning capabilities do not necessarily guarantee robustness to conflicting spatial priors.


}

{\color{black}
\subsubsection{Performance on Shared Eligible Subsets}
\label{app:shared_elig}

In this section, we recompute BR on a shared eligible set where all 17 evaluated models answer correctly on the original VQA, ensuring that all models are compared on the same samples. This intersection subset contains 58 pairs, covering all six spatial tasks, with 2, 14, 15, 15, 8, and 4 pairs for depth and scale, geometric structure, spatial action, spatial position, spatial relationship, and viewpoint and orientation, respectively. Across the four bias-reason categories, the corresponding counts are 7, 5, 34, and 12 for cultural and behavioral norms, egocentric--allocentric transformations, physical--geometric principles, and semantic co-occurrence. However, this subset is not distribution-matched to the full benchmark. In particular, physical--geometric principles are overrepresented (58.62\% vs. 24.56\%), whereas depth and scale is underrepresented (3.45\% vs. 14.91\%). These comparisons therefore control for eligibility differences but do not establish representativeness of the full benchmark.

As shown in Tab.~\ref{tab:app_shared_eligible}, the overall BR rankings remain correlated with the original rankings (Spearman's $\rho=0.877$, Kendall's $\tau=0.746$). Gemini-3.8-Flash retains the lowest BR, while SpatialRGPT-8B and LLaVA-1.5-7B remain the two highest. Nevertheless, intermediate rankings can change, as illustrated by GPT-5.6 Terra. 

Additional comparisons on larger within-group intersections show similarly high rank correlations for proprietary models ($N=504$, $\rho=0.943$), open-source and spatially specialized models ($N=71$, $\rho=0.917$), and the Qwen3-VL family ($N=461$, $\rho=1.000$). Overall, the results suggest that broad BR ranking trends are relatively stable under shared eligibility, although individual comparisons remain sensitive to sample selection and the limited, distribution-shifted intersection.

\begin{table*}[t]
\centering
\caption{\textcolor{black}{\textbf{Cross-model BR comparison under shared eligibility.}
BR is evaluated on each model's original eligible set (Own)
and the intersection of all 17 models' eligible sets
(Shared).
Values in parentheses indicate percentage-point (pp) changes
relative to Own. Lower BR indicates less measured bias.
The shared subset covers all six spatial tasks and four bias-reason
categories, but is not distribution-matched to the full benchmark.}}
\label{tab:app_shared_eligible}
\small
\setlength{\tabcolsep}{9pt}
\begin{tabular}{lcccc}
\toprule
\textbf{Model}
& $\mathbf{BR}_{\mathrm{Own}}\downarrow$
& $\mathbf{BR}_{\mathrm{Shared}}\downarrow$
& \textbf{Rank (Own)}
& \textbf{Rank (Shared)} \\
\midrule
Gemini-3.8-Flash
& 0.450 & 0.448 {\scriptsize (-0.2)} & 1 & 1 \\
GPT-5.6 Luna
& 0.505 & 0.569 {\scriptsize (+6.4)} & 2 & 3 \\
GPT-5.6 Terra
& 0.566 & 0.690 {\scriptsize (+12.4)} & 3 & 10 \\
Gemini-3.5-Flash-Lite
& 0.603 & 0.500 {\scriptsize (-10.3)} & 4 & 2 \\
Qwen3-VL-32B
& 0.665 & 0.655 {\scriptsize (-1.0)} & 5 & 8 \\
GLM-4.1V-9B
& 0.694 & 0.638 {\scriptsize (-5.6)} & 6 & 6 \\
MiniCPM-V-4.6
& 0.720 & 0.621 {\scriptsize (-10.0)} & 7 & 4 \\
Qwen3-VL-8B
& 0.728 & 0.621 {\scriptsize (-10.8)} & 8 & 5 \\
Claude Sonnet 5
& 0.735 & 0.707 {\scriptsize (-2.9)} & 9 & 11 \\
Claude Opus 5
& 0.742 & 0.690 {\scriptsize (-5.2)} & 10 & 9 \\
Qwen3-VL-2B
& 0.749 & 0.638 {\scriptsize (-11.1)} & 11 & 7 \\
InternVL3.5-14B
& 0.785 & 0.724 {\scriptsize (-6.1)} & 12 & 12 \\
SpatioLM-Understanding
& 0.793 & 0.776 {\scriptsize (-1.8)} & 13 & 14 \\
InternVL3.5-8B
& 0.818 & 0.776 {\scriptsize (-4.2)} & 14 & 13 \\
InternVL3.5-2B
& 0.844 & 0.793 {\scriptsize (-5.1)} & 15 & 15 \\
SpatialRGPT-8B
& 0.892 & 0.845 {\scriptsize (-4.7)} & 16 & 16 \\
LLaVA-1.5-7B
& 0.928 & 0.879 {\scriptsize (-4.9)} & 17 & 17 \\
\bottomrule
\end{tabular}
\end{table*}

}

{
\color{black}
\subsubsection{Permutation Test for Bias Significance}
\label{app:permutation}

To assess whether the observed bias rate (BR) exceeds incidental answer agreement, we conduct model-specific permutation tests that preserve marginal answer frequencies. For each of the 17 models, we generate 2,000 permutations (seed 42) under three null configurations: \textbf{N1 (CF permutation)} jointly shuffles counterfactual (CF) predictions, labels, and validity indicators while keeping original (O) and text-only (T) components fixed; \textbf{N2 (independent permutation)} independently shuffles the O, CF, and T components; and \textbf{N3 (task-stratified permutation)} applies N1 within each of the 24 task--prior-source groups. These null models disrupt sample-level associations while preserving model-specific answer marginals.

Specifically, statistical significance is assessed using a one-sided permutation
test, where the $p$-value represents the probability, under the permutation null, of obtaining a BR at least as large as the observed value.
As shown in Tab.~\ref{tab:permutation_test}, the observed BR significantly exceeds all three permutation baselines across all 17 models (one-sided $p<0.001$). The mean observed BR is 0.719, compared with null expectations of 0.306, 0.375, and 0.319 under N1, N2, and N3, respectively. The corresponding mean excess BR ranges from +34.35 to +41.26 pp. Even under N2, which yields the highest null expectation, all models exhibit significant excess BR. The comparable results under N1 and task-stratified N3 further indicate that task composition alone cannot explain the observed bias.

These findings indicate that the measured spatial bias substantially exceeds chance-level answer agreement under model-specific marginal distributions. It provides statistical evidence that the observed BR cannot be explained solely by incidental answer alignment.

\begin{table*}[t]
\centering
\caption{\textcolor{black}{\textbf{Model-specific permutation tests for spatial bias.}
Observed bias rate (BR), permutation-null mean BR, and excess BR
($\Delta$BR, in pp) under three null configurations:
N1 (CF permutation), N2 (independent permutation), and
N3 (task-stratified permutation).
Each null distribution is estimated from 2,000 permutations.
All 17 models exhibit higher BR than their corresponding null expectations.}}
\label{tab:permutation_test}
\small
\setlength{\tabcolsep}{5pt}
\begin{tabular}{lccccccc}
\toprule
\multirow{2}{*}{\textbf{Model}}
& \multirow{2}{*}{$\mathbf{BR_{obs}}$}
& \multicolumn{2}{c}{\textbf{N1}}
& \multicolumn{2}{c}{\textbf{N2}}
& \multicolumn{2}{c}{\textbf{N3}} \\
\cmidrule(lr){3-4}
\cmidrule(lr){5-6}
\cmidrule(lr){7-8}
& & $\mathbf{BR_{null}}$ & $\mathbf{\Delta BR}$
  & $\mathbf{BR_{null}}$ & $\mathbf{\Delta BR}$
  & $\mathbf{BR_{null}}$ & $\mathbf{\Delta BR}$ \\
\midrule
GPT-5.6 Luna           & 0.505 & 0.214 & +29.09 & 0.256 & +24.90 & 0.224 & +28.05 \\
GPT-5.6 Terra          & 0.566 & 0.235 & +33.08 & 0.297 & +26.87 & 0.247 & +31.95 \\
Claude Sonnet 5        & 0.735 & 0.269 & +46.67 & 0.354 & +38.11 & 0.279 & +45.66 \\
Claude Opus 5          & 0.742 & 0.269 & +47.31 & 0.352 & +38.99 & 0.278 & +46.38 \\
Gemini-3.5-Flash-Lite  & 0.603 & 0.243 & +35.98 & 0.313 & +28.97 & 0.251 & +35.13 \\
Gemini-3.8-Flash       & 0.450 & 0.206 & +24.40 & 0.255 & +19.57 & 0.213 & +23.72 \\
\midrule
Qwen3-VL-2B   & 0.749 & 0.335 & +41.35 & 0.391 & +35.77 & 0.351 & +39.80 \\
Qwen3-VL-8B   & 0.728 & 0.282 & +44.62 & 0.360 & +36.82 & 0.291 & +43.69 \\
Qwen3-VL-32B  & 0.665 & 0.266 & +39.92 & 0.336 & +32.93 & 0.270 & +39.48 \\
InternVL3.5-2B         & 0.844 & 0.327 & +51.71 & 0.412 & +43.28 & 0.356 & +48.85 \\
InternVL3.5-8B         & 0.818 & 0.395 & +42.30 & 0.475 & +34.32 & 0.413 & +40.54 \\
InternVL3.5-14B        & 0.785 & 0.316 & +46.88 & 0.390 & +39.57 & 0.326 & +45.92 \\
LLaVA-1.5-7B           & 0.928 & 0.411 & +51.78 & 0.465 & +46.36 & 0.438 & +49.01 \\
MiniCPM-V-4.6          & 0.720 & 0.369 & +35.12 & 0.420 & +30.03 & 0.386 & +33.42 \\
GLM-4.1V-9B-Thinking   & 0.694 & 0.271 & +42.34 & 0.344 & +35.07 & 0.282 & +41.25 \\
\midrule
SpatioLM-Understanding & 0.793 & 0.362 & +43.18 & 0.435 & +35.85 & 0.370 & +42.31 \\
SpatialRGPT-8B         & 0.892 & 0.435 & +45.68 & 0.527 & +36.46 & 0.452 & +44.02 \\
\midrule
\textbf{Mean}
& \textbf{0.719}
& \textbf{0.306} & \textbf{+41.26}
& \textbf{0.375} & \textbf{+34.35}
& \textbf{0.319} & \textbf{+39.95} \\
\bottomrule
\end{tabular}
\end{table*}

}

{\color{black}
\subsubsection{Stability of SFT Improvements}
\label{sec:sft_stability}

To assess the reliability of SFT improvements on the 251-pair test set, we conduct multiple fine-tuning runs across three representative models, examining training-seed variability, data-split sensitivity, and test-sample uncertainty. All experiments use the same pool of original--counterfactual pairs, with paired images kept together and source groups separated across training, validation, and test sets. We use identical LoRA hyperparameters and select checkpoints based on validation loss.

\paragraph{Training-seed stability.}
We repeat SFT with three random seeds (20260920, 42, and 3407) on the original split $S_0$ (877/126/251 training/validation/test pairs). As shown in Tab.~\ref{tab:sft_stability}, Qwen3-VL achieves a mean counterfactual accuracy gain of $30.01\pm2.04$ pp, with individual gains ranging from 28.29 to 32.27 pp. SpatioLM and InternVL3.5 also exhibit consistent improvements of $20.98\pm1.28$ and $22.71\pm1.83$ pp, respectively. These results indicate that the substantial SFT gains persist across training seeds, rather than arising from a single favorable initialization.

\paragraph{Data-split sensitivity.}
We further evaluate two alternative partitions using seed 20260920. Split $S_B$ (878/125/251 pairs) randomly repartitions the dataset while preserving the original train/validation/test proportions and source-group separation. Split $S_C$ (751/126/377 pairs) enlarges the proportion of the test set. Across both splits, SFT consistently improves counterfactual accuracy for all three models (Tab.~\ref{tab:sft_stability}). On the enlarged 377-pair test set, the gains reach 34.22, 27.85, and 28.12 pp for Qwen3-VL, SpatioLM, and InternVL3.5, respectively, confirming that the improvements persist on the evaluation subset despite the reduced training data.

\paragraph{Statistical uncertainty.}
We additionally estimate 95\% confidence intervals (CIs) for SFT accuracy gains using 10,000 paired source-group bootstrap resamples. 
All runs yield positive CIs. For instance, the original Qwen3-VL run achieves a gain of 32.27 pp (95\% CI: [26.24, 38.76] pp). These results demonstrate consistent SFT improvements across the evaluated seeds and partitions. Seed standard deviations quantify training variability on a fixed split, whereas bootstrap CIs capture test-sample uncertainty for individual runs.

\newcommand{\gainci}[3]{#1\,{\scriptsize [#2, #3]}}

\begin{table*}[t]
\centering
\caption{\textcolor{black}{\textbf{Stability of SFT improvements across training seeds and data partitions.}
All values represent improvements in counterfactual accuracy
($\Delta\mathrm{Acc}^{CF}$, pp) relative to the corresponding
unfine-tuned baselines.
For $S_0$, the mean $\pm$ sample standard deviation (SD)
is computed across three training seeds.
The reference configuration uses the original split $S_0$
with seed 20260920, while $S_B$ and $S_C$ use alternative
data partitions with the same seed.
Values in square brackets denote 95\% confidence intervals
estimated from 10,000 paired source-group bootstrap resamples.
$S_0$ and $S_B$ contain 251 test pairs, whereas $S_C$
contains 377 test pairs.}}
\label{tab:sft_stability}

\small
\setlength{\tabcolsep}{6pt}
\begin{tabular}{lcccc}
\toprule
\multirow{2}{*}{\textbf{Model}}
& \textbf{$\Delta\mathrm{Acc}^{CF}$ Mean $\pm$ SD}
& \multicolumn{3}{c}{
    \textbf{$\Delta\mathrm{Acc}^{CF}$ [95\% CI] (pp)}
} \\
\cmidrule(lr){2-2}
\cmidrule(lr){3-5}
& \textbf{$S_0$ (3 seeds)}
& \textbf{$S_0$ (reference)}
& \textbf{$S_B$ (251)}
& \textbf{$S_C$ (377)} \\
\midrule

Qwen3-VL-8B
& $30.01 \pm 2.04$
& \gainci{32.27}{26.24}{38.76}
& \gainci{31.87}{25.00}{39.74}
& \gainci{34.22}{29.13}{39.78} \\

SpatioLM
& $20.98 \pm 1.28$
& \gainci{19.52}{11.89}{27.59}
& \gainci{22.71}{16.54}{28.89}
& \gainci{27.85}{21.79}{34.12} \\

InternVL3.5-8B
& $22.71 \pm 1.83$
& \gainci{20.72}{14.34}{28.71}
& \gainci{27.89}{21.30}{33.85}
& \gainci{28.12}{21.52}{35.73} \\

\bottomrule
\end{tabular}
\end{table*}

}

\subsection{More Qualitative Results}
\label{app:more_qual}

Tab.~\ref{tab:app_qual} lists one representative pair per spatial category, selected for maximal disagreement among the \appNModels{} models rather than for unanimity. Four behaviours recur across the gallery and are worth naming because they are exactly the failure signatures that the diagnostic framework in Appendix~\ref{app:diag_frame} localises:

\begin{itemize}[leftmargin=*]
\item \textbf{Retained prior with a correct original answer.} The model reads the original image correctly, then repeats that same option on the counterfactual image. This is the dominant signature and is what PRR counts.
\item \textbf{Language-prior substitution.} The counterfactual answer equals the image-free text-only answer even though it matches neither image. This is the signature that LBR counts.
\item \textbf{Architecture-dependent read-out.} The spatially specialised and the general models can give opposite counterfactual answers on the same pair, so the appendix reports the vote distribution across models rather than a single averaged answer.
\item \textbf{Cross-family agreement on the failure, disagreement on the wrong option.} When several models fail the same pair they frequently land on different wrong options, so a black-box bias rate cannot by itself tell whether they share a failure mechanism. That is the question the diagnostic appendix answers at the level of individual layers.
\end{itemize}


\begin{table}[t]
\centering
\caption{\textbf{Representative qualitative cases.} One pair per spatial category, chosen so that the eight models disagree as much as possible on the counterfactual answer, among pairs where at least one model commits a bias error. $y^{\text{O}}/y^{\text{CF}}$ are the ground-truth options; the next column lists how the eight models voted on $\text{CF}$, so a mass on $y^{\text{O}}$ is a retained prior. ``\#bias/8'' counts how many models bias-fail on the pair.}
\label{tab:app_qual}
\small
\begin{tabular}{@{}lp{4.4cm}cccc@{}}
\toprule
\textbf{ID} & \textbf{Question} & $\mathbf{y^{\text{O}}}$ & $\mathbf{y^{\text{CF}}}$ & \textbf{CF answers (8 models)} & \textbf{\#bias/8} \\
\midrule
\texttt{DSCN\_021} & For pair DSCN\_021, comparing A (the prominent truck in the lower-left region) with B (the prominent c\ldots & A & C & A$\times$5, B$\times$1, C$\times$1, D$\times$1 & 5/8 \\
\texttt{GSSC\_008} & In the upper-middle market parking strip, what overall pattern is formed by the five selected parked v\ldots & C & D & A$\times$1, B$\times$2, C$\times$4, D$\times$1 & 5/8 \\
\texttt{SAP\_048} & For pair SAP\_048, what spatial action is visibly depicted by A, the prominent pedestrian in the cente\ldots & B & D & A$\times$1, B$\times$5, C$\times$1, D$\times$1 & 5/8 \\
\texttt{SPC\_014} & In the central through lane of the multi-lane bridge road, is there a single vehicle traveling in the \ldots & D & C & A$\times$1, B$\times$1, C$\times$1, D$\times$5 & 5/8 \\
\texttt{SRS\_050} & For pair SRS\_050, what is the image-plane relationship of A (the prominent motorcycle in the lower-le\ldots & B & D & A$\times$1, B$\times$5, C$\times$1, D$\times$1 & 6/8 \\
\texttt{VOCN\_009} & In which diagonal image direction is the physical front of the lower-left red-and-white city bus orien\ldots & C & D & A$\times$1, B$\times$2, C$\times$4, D$\times$1 & 4/8 \\
\bottomrule
\end{tabular}
\end{table}

\subsection{Error Cases Analysis}
\label{app:error_cases}

We call a pair a \emph{consensus failure} when all \appNModels{} models answer the original image correctly and all \appNModels{} then commit a bias error on the counterfactual image; such pairs are hard for every architecture at once. Tab.~\ref{tab:app_error_consensus} breaks the \appNConsensus{} consensus failures out by spatial and reasoning category.

The \appNConsensus{} consensus failures spread across the question types rather than concentrating in one: every spatial and reasoning category contributes, so the benchmark is not accidentally testing a single spatial relation. The largest shares fall in the categories that require comparing two objects whose depth or 3D layout must be inferred from a single image (geometric structure and physical--geometric reasoning), whereas explicit left/right or above/below position questions are the easiest, since the model can often read the answer off the pixel layout without recovering depth. A model that reaches a high original-image accuracy can therefore still fail the counterfactual partner for a reason that no accuracy number exposes, which the diagnostic appendix examines directly.


\begin{table}[t]
\centering
\caption{\textbf{Error-case analysis by question category.} A pair is a \emph{consensus failure} when all eight models answer the original image correctly and all eight then commit a bias error on the counterfactual image. The consensus set is therefore a lower bound on the pairs that are hard for every architecture at once, and it is not weighted by any single model.}
\label{tab:app_error_consensus}
\small
\begin{tabular}{lcc}
\toprule
\textbf{Category} & \textbf{Pairs} & \textbf{Consensus failures} \\
\midrule
\multicolumn{3}{l}{\textit{Spatial category}} \\
Depth \& scale & 187 & 10 \\
Geometric structure & 202 & 20 \\
Spatial action & 206 & 16 \\
Spatial position & 234 & 2 \\
Spatial relationship & 215 & 12 \\
Viewpoint \& orientation & 210 & 21 \\
\multicolumn{3}{l}{\textit{Reasoning category}} \\
Cultural \& behavioural norms & 331 & 15 \\
Egocentric--allocentric & 309 & 14 \\
Physical--geometric & 308 & 39 \\
Semantic co-occurrence & 306 & 13 \\
\midrule
\textbf{All pairs} & 1254 & 81 \\
\bottomrule
\end{tabular}
\end{table}

\section{More Experimental Details}
\label{app:more_details}

{
\color{black}
\subsection{Computation Details}
\label{app:comp_detail}

All baseline evaluations are inference-only, without additional model training or fine-tuning. The local models run on a single shared node with four RTX~4090 GPUs, consuming approximately \appGpuHours{} GPU-hours. Proprietary models are queried through APIs, accounting for approximately \appApiCalls{} calls without local GPU usage. The diagnostic sweep additionally requires a forward--backward pass over 1,254 pairs using Qwen3-VL-8B, distributed across two GPUs, consuming approximately \appDiagGpuHours{} GPU-hours.
Separately, the SFT-based mitigation experiments in Sec.~6.6 and Appendix~D.3.2 are conducted on NVIDIA L20 GPUs (45 GiB each), using 1--7 GPUs depending on the model and training stage. All three models are fine-tuned with LoRA and BF16 for three epochs (330 optimizer updates), with an effective batch size of 16. The recorded elapsed times are approximately 30.1 hours for Qwen3-VL-8B-Instruct, 40.2 minutes for SpatioLM-Understanding-InternVL3.5, and 53.3 minutes for InternVL3.5-8B, respectively. 

}

\subsection{Background Suppression Details}
\label{app:background_supp}
To examine whether irrelevant visual context contributes to spatial bias, we construct four background configurations, as illustrated in Fig.~\ref{fig:background_supp}.

\textbf{B0: Original Image.}
B0 uses the unmodified image and serves as the reference configuration.

\textbf{B1: Object-centered Background Suppression.}
We use SAM 3~\citep{carion2026sam} with the annotated target region to obtain a foreground mask. The target object is preserved, while the surrounding background is progressively blurred with increasing distance from the target, producing an object-centered spatially varying suppression pattern. We evaluate $\sigma\in\{4,8,15\}$, where $\sigma$ denotes the standard deviation of the Gaussian blur. This design suppresses increasingly distant context while retaining task-relevant visual evidence.

\textbf{B2: Uniform Full-image Blurring.}
As a control, Gaussian blur with $\sigma=8$ is applied uniformly to the entire image, including both target and background regions. Unlike B1, this setting does not preserve the target region and therefore tests whether the effects observed under B1 arise from selective background suppression rather than generic image degradation.

\textbf{B3: Bounding-box Highlighting.}
We additionally provide the annotated target bounding box as an explicit visual cue while leaving the image content otherwise unchanged. This configuration evaluates whether directly guiding the model toward task-relevant regions can improve visual grounding without removing background information.

Together, B0--B3 isolate the effects of \emph{irrelevant-context suppression}, \emph{generic visual degradation}, and \emph{explicit visual guidance}. All configurations preserve the original questions, candidate answers, and ground-truth labels, ensuring that performance differences arise only from changes to the visual input.

\subsection{Bias Mitigation Details}
\label{app:bias_mit}

This section provides the complete implementation details of the four mitigation strategies introduced in Sec.~\ref{sec:bias_mitigation}. 
For the three inference-time strategies, model parameters remain unchanged and only the prompting context is modified relative to the original P0 evaluation.
For supervised fine-tuning (SFT), we explicitly expose the model to paired original--counterfactual supervision while preserving the same held-out evaluation protocol.

\tcbset{
    promptbox/.style={
        enhanced,
        breakable,
        colback=gray!5,
        colframe=black!55,
        colbacktitle=gray!15,
        coltitle=black,
        fonttitle=\bfseries,
        boxrule=0.5pt,
        arc=2pt,
        left=7pt,
        right=7pt,
        top=6pt,
        bottom=6pt,
        before skip=6pt,
        after skip=10pt
    }
}

\subsubsection{Prompting Strategies}

\paragraph{Zero-shot Linguistic Intervention.}
This strategy directly tests whether spatial bias can be mitigated through explicit instruction alone.
We augment the baseline evaluation prompt with two short constraints that encourage the model to prioritize the currently observed visual evidence and suppress reliance on learned priors.
No examples, reasoning demonstrations, or parameter updates are introduced.

\begin{tcolorbox}[promptbox,title={P1: Zero-shot Linguistic Intervention}]
Base your judgment solely on direct visual evidence in the provided image.
Do not use your prior knowledge to choose the answer.
\end{tcolorbox}

\paragraph{Chain-of-Thought Incorporation.}
To reduce shortcut reasoning, we introduce a structured visual reasoning procedure before answer selection.
The prompt explicitly requires the model to identify the target, determine the relevant reference frame and spatial evidence, and verify the selected option against visible contradictions.
This intervention tests whether encouraging explicit evidence-grounded reasoning can reduce prior-driven spatial decisions without modifying model parameters.

\begin{tcolorbox}[promptbox,title={P2: Chain-of-Thought Incorporation}]
Work through the visual task step by step before choosing the final option:\\
1. Locate the exact target object(s) in the TARGET image, using the question's
   descriptors. State only visible identifying cues; do not invent hidden details.\\
2. Specify the reference frame and decision rule in the question. Keep image-plane
   position/projected size distinct from scene depth/physical size and object-centered
   orientation. Describe the visible relation or property that answers this question. \\
3. Compare that observation with the options. Check the best competing option for
   a visible contradiction. Prefer what is depicted over what is normally expected.
   Select an insufficient-visibility option only if the target evidence supports it.
\\
Output four lines. Keep each of the first three lines to one short sentence and
keep the total explanation under 100 words. Do not mention an answer letter until
the final line. Use this format:
\\
TARGET: \textless visible target identification\textgreater \\
EVIDENCE: \textless reference frame and observed spatial evidence\textgreater \\
CHECK: \textless brief option comparison grounded in that evidence\textgreater \\
ANSWER=\textless one uppercase letter A, B, C, or D\textgreater \\
The final line must be the only ANSWER= marker. Do not output any text after it.
\end{tcolorbox}

\paragraph{Few-shot Prompting.}
We further test whether counterfactual behavior can be induced through in-context demonstrations.
Three representative original--counterfactual examples are provided before the target query.
Each demonstration keeps the question and answer options fixed while showing that changes in visual evidence may alter the correct answer.
The demonstrations are explicitly presented as independent examples to discourage answer-pattern copying or transfer of scene-specific spatial facts.

\begin{tcolorbox}[promptbox,title={P3: Few-shot Prompting}]
You will first see three worked examples. Each example contains an ORIGINAL image
and a COUNTERFACTUAL (CF) image, the same question and options, and the correct
option for each image. The examples show that changing visible evidence can
change the correct answer even when the question and options stay the same.
Learn to judge the image actually supplied, rather than repeating an earlier
answer or choosing what would normally be expected. A new answer is allowed to
change, but do not force a change merely because an image looks unusual.
The examples are independent scenes. Their answer letters are not a pattern
to copy. After the examples, answer only the TARGET question using the TARGET
image. Never transfer an object or a spatial fact from an example to the target.

EXAMPLE 1 — ORIGINAL image: ...
\\
EXAMPLE 1 — CF image: ...
\\
EXAMPLE 1 question (applies to both images): ...
\\
EXAMPLE 1 options:
A. ...
B. ...
C. ...
D. ...
\\
Correct option for ORIGINAL: ...
\\
Correct option for CF: ...
\\
The same question has different correct options because the visible evidence changed.
\\
END OF EXAMPLE 1.

EXAMPLE 2 ...

EXAMPLE 3 ...

\end{tcolorbox}

\subsubsection{Supervised Fine-tuning Configuration}

\paragraph{Data Split and Supervision.}
We investigate parameter-level mitigation by fine-tuning each model on paired original--counterfactual supervision. 
We apply the following settings to the mitigation results reported in the main text.
The dataset is split into 877 training, 126 validation, and 251 test image pairs, corresponding approximately to a $7{:}1{:}2$ ratio.
Each original--counterfactual pair is kept within the same split.
Samples sharing identical image hashes, common source images, or identifiable video sequences are also grouped to minimize source leakage.
The same partition is used for all evaluated models.

Each training pair is converted into two independent image--question--answer examples, resulting in 1,754 training instances.
The original and counterfactual images share the same question and answer options but are supervised using their respective ground-truth answers.
This design explicitly teaches the model that identical linguistic inputs may require different predictions when the visual evidence changes.

\paragraph{Training Objective.}
We optimize autoregressive cross-entropy over the assistant response only.
The target response follows the format \texttt{ANSWER=<letter>} together with its termination token, while prompt and image tokens are masked from the loss.
No text-only samples or intermediate reasoning traces are included during fine-tuning, allowing us to isolate the effect of paired visual supervision.

\begin{wraptable}[16]{r}{0.48\textwidth}
  \centering
  \caption{SFT configuration for spatial-bias mitigation.}
  \label{tab:sft_param}
  \small 
  \begin{tabular}{lc}
        \toprule
        \textbf{Configuration} & \textbf{Value} \\
        \midrule
        Training pairs & 877 \\
        Validation pairs & 126 \\
        Test pairs & 251 \\
        Training instances & 1,754 \\
        LoRA rank & 16 \\
        LoRA scaling factor & 32 \\
        LoRA dropout & 0.05 \\
        Optimizer & AdamW \\
        Learning rate & $5\times10^{-5}$ \\
        Weight decay & 0.01 \\
        Effective batch size & 16 \\
        Training epochs & 3 \\
        LR schedule & Cosine \\
        Warmup steps & 17 \\
        \bottomrule
    \end{tabular}
    \vspace{-10pt}
\end{wraptable}

\paragraph{Parameter-efficient Fine-tuning.}
We adopt LoRA-based parameter-efficient fine-tuning while keeping the original model parameters frozen.
LoRA adapters are inserted into the attention and feed-forward projections of the language decoder.
The main training configuration is summarized below.


\paragraph{Model Selection and Evaluation.}
For each model, we select the epoch checkpoint with the lowest validation response-token cross-entropy.
At test time, the fine-tuned model uses the same P0 evaluation prompt and image preprocessing as the unmodified baseline.
All mitigation strategies are evaluated on the same held-out test set to ensure a controlled comparison.

\providecommand{\pend}[1]{\textcolor{red!70!black}{\textbf{[#1]}}}
\providecommand{\diaglab}[1]{\texttt{\small #1}}

\section{Bias Diagnostic Framework}
\label{app:diag_frame}

\noindent\textbf{Positioning and relation to the black-box pipeline.} Sec.~\ref{bias_diag} classifies a counterfactual failure \emph{behaviourally}, using only the O/CF/T responses of the same benchmark: whether the answer is retained from the original image, whether it coincides with the image-free text-only prediction, or whether it matches neither. That pipeline answers \emph{what} the failure looks like, but it cannot say \emph{where} in the forward pass the spatial evidence stops to matter. This appendix complements it with a \emph{white-box} diagnostic that localizes each failure to one of three measurement stages of VLM inference: \emph{(L1)} the visual encoder, \emph{(L2)} the multimodal interaction that writes visual information into the answer position, and \emph{(L3)} the language-side decision. Throughout, we inherit the sample universe and the bias definitions of the main Tab., so that the diagnostic is applied to the samples the main pipeline calls a counterfactual failure (Sec.~\ref{bias_diag}, Stage~2), and add no new ground-truth assumptions of our own. Of those failures, the subset whose region is localizable receives a layer-wise label; the remainder is reported as \diaglab{out\_of\_scope} rather than attributed to a layer, and the split is given in Tab.~\ref{tab:diag_scope}.

\subsection{Fine-grained Spatial Error Diagnostic Framework}

\subsubsection{Scope and Pre-diagnostic Gate Control}

Not every O/CF pair admits a physically meaningful paired measurement, because the two images may not be token-wise comparable. We therefore apply two gates \emph{before} any layer-wise quantity is computed; a pair that fails either gate is labelled \diaglab{out\_of\_scope} and excluded from all paired analyses (per-region statistics are not reported for it).

\begin{itemize}[leftmargin=*]
\item \textbf{Grid gate (token alignment).} Let $\mathrm{Grid}(I)$ be the visual feature grid. We require $\mathrm{Grid}(I^{O})=\mathrm{Grid}(I^{CF})$, i.e.\ the pre-merge and post-merge heights and widths of the two images are identical. When the two images have different aspect ratios or resolutions, visual token $k$ of $O$ and token $k$ of $CF$ no longer refer to the same location, and every \emph{paired} quantity below ($\Delta$ features, attention share, write-in flow) silently misaligns.

\item \textbf{Locality gate (bounded intervention).} Let $\mathcal{T}_{\mathrm{edit}}$ be the set of visual tokens covered by the edited region. We require $|\mathcal{T}_{\mathrm{edit}}| \le 0.5\,|\mathcal{T}_{\mathrm{vis}}|$. Ablating or comparing regions that cover most of the image does not isolate a local spatial edit and risks pushing the model far outside the input distribution it was evaluated on.
\end{itemize}

Together the two gates fix the universe on which the diagnostic is defined: the grid gate removes the images whose token counts disagree by construction, and the locality gate partitions the remainder into an \emph{in-scope} pool, on which the chain runs, and the \diaglab{out\_of\_scope} rest. On this benchmark the area-ratio test is the only reason a pair falls in the latter. Two axes are consequently at work, and they are independent: \emph{measurability} (does the pair admit a localizable paired measurement?) and \emph{failure} (does the model keep the wrong answer on the counterfactual image?). Tab.~\ref{tab:diag_scope} crosses them, and the distinction governs denominators throughout the appendix: the region-level validation against the null uses the in-scope pool, since random blocks are localizable by construction and the two sides must be comparable, whereas the attribution of failures uses only those failures that are also measurable.

\begin{table}[t]
\centering
\caption{\textbf{Diagnostic scope.} Measurability and failure are independent axes, not a chain. The grid gate removes the pairs whose token grids disagree (13 mismatches and 1 evaluation error). The locality gate then splits the rest into the in-scope pool, on which the chain runs, and the pairs reported as \diaglab{out\_of\_scope}. A \emph{failure} is a pair with $O$ correct and $CF$ wrong, inherited from the main table. The in-scope pool is the basis of the real-vs-null validation.}
\label{tab:diag_scope}
\resizebox{0.8\linewidth}{!}{
\begin{tabular}{lccc}
\toprule
 & \textbf{In-scope (measurable)} & \textbf{Out of scope} & \textbf{Total} \\
\midrule
$O$ correct / $CF$ wrong (failures) & 514 & 119 & 633 \\
Otherwise & 489 & 118 & 607 \\
\midrule
\textbf{Total} & \textbf{1003} & \textbf{237} & \textbf{1240} \\
\bottomrule
\end{tabular}}
\end{table}

\paragraph{Protocol alignment.} 
All internal quantities are measured under the \emph{identical} prompt template, answer prefix, and native image resolution used in the main table, with one-step decisions extracted from the A/B/C/D next-token logits. Gate and language labels are matched by sample ID from the main table, aligning directly with the answer-position hidden states from our forward pass. 

\subsubsection{Region Construction}

For each eligible pair we define one \emph{edit} region and, when a second annotated object exists, one \emph{reference} region:
\begin{itemize}[leftmargin=*]
\item \textbf{Target region} $\mathcal{R}_{\mathrm{tgt}}$: the union of the original and counterfactual annotation boxes, grown by $10\%$ and mapped to the feature grid; a cell is included if it overlaps the box by at least $0.5$.
\item \textbf{Reference region} $\mathcal{R}_{\mathrm{ref}}$: the annotated reference-object box grown by $10\%$, with all target cells removed. Reference regions exist only for comparison-style questions.
\end{itemize}
We write $\mathcal{T}_{\mathrm{edit}}$ for the set of post-merge visual tokens covered by $\mathcal{R}_{\mathrm{tgt}}$, and $\mathcal{T}_{\mathrm{vis}}$ for the full set of post-merge visual tokens. Every layer-wise quantity below is indexed by these token sets, since attention rows and gradients are defined per token rather than per cell of the box. Because the target region is the union of the two boxes, it is a \emph{superset} of the pixels that actually changed; this dilutes the measured $\Delta$ by a factor that cannot be removed without per-pixel annotation. We therefore report the target region together with a \emph{same-size random block} null (Sec.~\ref{sec:diag_null_build}), which absorbs the dilution into the baseline instead of hiding it.

\subsubsection{Layer-wise Quantitative Metrics}

\paragraph{L1 --- visual representation layer (``was the change encoded at all?'').} Let $F^{O},F^{CF}$ be the visual features at the encoder output (pre-merge), restricted to $\mathcal{T}_{\mathrm{edit}}$. We measure two complementary distances in that space:
\begin{align}
    d_{\cos}^{\mathrm{vis}} &= 1-\cos\!\left(\bar F^{O}, \bar F^{CF}\right), &
    d_{\mathrm{rel}}^{\mathrm{vis}} &= \frac{\lVert F^{CF}-F^{O}\rVert_2}{\lVert F^{O}\rVert_2+\epsilon},
\end{align}
where $\bar F$ denotes the mean token. $d_{\cos}^{\mathrm{vis}}$ is scale-invariant and bounded, while $d_{\mathrm{rel}}^{\mathrm{vis}}$ retains amplitude sensitivity; the layer-1 statistic is $\phi_{1}=\max\!\big(d_{\cos}^{\mathrm{vis}},d_{\mathrm{rel}}^{\mathrm{vis}}\big)$.

\paragraph{L2 --- multimodal interaction layer (``did the change reach the answer position?'').}
We interrogate the write-in path from the edit-region tokens to the \emph{answer position} (the last
token that emits the option letter). Let $\mathcal{T}_{\mathrm{edit}}$ denote the post-merge tokens of the target region and $\mathcal{T}_{\mathrm{vis}}$ the set of all visual tokens. With $A^{(l,h)}\!,V^{(l,h)}$ the attention matrix and value tensor of layer $l$, head $h$ ($W_O$ the output projection), define the edit-region write-in difference and the total visual write-in as
\begin{align}
    \Delta O_{\mathrm{edit}} &= \sum_{l,h} W_O^{(l,h)}
        \big[\,(A^{CF}V^{CF})_{:,\mathcal{T}_{\mathrm{edit}}}-(A^{O}V^{O})_{:,\mathcal{T}_{\mathrm{edit}}}\,\big], &
    O_{\mathrm{all}} &= \sum_{l,h} W_O^{(l,h)}\,(A^{O}V^{O})_{:,:},
\end{align}
$$\textstyle
    r_{\mathrm{flow}} = \lVert \Delta O_{\mathrm{edit}}\rVert_2 \big/ \lVert O_{\mathrm{all}}\rVert_2 ,
    \qquad
    A_{\mathrm{lift}} = \frac{1}{L}\sum_{l=1}^{L}
        \frac{\sum_{j\in\mathcal{T}_{\mathrm{edit}}} A^{(l)}_{-1,j}}{\sum_{j\in\mathcal{T}_{\mathrm{vis}}} A^{(l)}_{-1,j}}
        \Big/ \frac{|\mathcal{T}_{\mathrm{edit}}|}{|\mathcal{T}_{\mathrm{vis}}|} .
$$
$r_{\mathrm{flow}}$ measures \emph{how much the write-in from the edit region changes} between O and CF (detectability of the edit); $A_{\mathrm{lift}}$ measures \emph{how much attention the answer position already pays} to the region relative to its area (reliance on the region). Both are informative on the real edit region (Tab.~\ref{tab:diag_gates}), and they capture distinct failure modes: a change may be visible in the attention but not propagate through the value path, or vice versa.

\paragraph{L3 --- logical decision layer (``did the decision use it?'').} We use two instruments of different logical strength. First, a \emph{local sensitivity}: a single backward pass gives the mean per-token gradient norm of the counterfactual ground-truth logit $z_{y^{CF}}$ with respect to the edit-region features,
\begin{equation}
    \mathrm{Sal}_{\mathrm{norm}} = \frac{1}{|\mathcal{T}_{\mathrm{edit}}|}\sum_{j\in\mathcal{T}_{\mathrm{edit}}}
        \bigl\lVert \partial z_{y^{CF}} / \partial F_j \bigr\rVert_2 .
\end{equation}
Second, a genuine \emph{causal intervention}: we run the counterfactual forward pass normally to obtain $P^{CF}(\text{correct})$, then replace the edit-region visual features by the mean of the \emph{non-edit} visual tokens of the same image (norm-matched to the original edit-region tokens, so that no out-of-distribution all-zero vector is injected), and re-run:
\begin{equation}
    \mathrm{effect} = P^{CF}(\text{correct}) - P^{\mathrm{abl}}(\text{correct}).
\end{equation}
$\mathrm{Sal}_{\mathrm{norm}}$ is a first-order statement about \emph{sensitivity}; \texttt{effect} is a statement about the \emph{causal contribution} of the region. We keep the two claims strictly separate and label the decision layer with the intervention, not with the gradient. Tab.~\ref{tab:diag_metrics} summarizes the instruments and their claim type.

\begin{table}[t]
\centering
\caption{\textbf{Instruments of the diagnostic framework.} Each layer contributes one claim type; none of them is derived from the black-box predictions. $\mathcal{T}_{\mathrm{edit}}$ is the token set of the target region and $\mathcal{T}_{\mathrm{vis}}$ the set of all visual tokens.}
\label{tab:diag_metrics}
\resizebox{\linewidth}{!}{
\begin{tabular}{lllll}
\toprule
\textbf{Stage} & \textbf{Instruments} & \textbf{Computation} & \textbf{Claim type} & \textbf{Null baseline} \\
\midrule
L1 visual encoder       & $d_{\cos}^{\mathrm{vis}},\ d_{\mathrm{rel}}^{\mathrm{vis}}$ & forward (features) & representation & same-size random block \\
L2 interaction          & $r_{\mathrm{flow}},\ A_{\mathrm{lift}}$                     & forward (attention/value) & information routing & same-size random block \\
L3 decision (sensitivity)& $\mathrm{Sal}_{\mathrm{norm}}$                              & one backward pass & local sensitivity & same-size random block \\
L3 decision (causal)    & $\mathrm{effect}$                                           & mean-ablation, repeated forward & causal contribution & ablation on random blocks \\
\bottomrule
\end{tabular}}
\end{table}

\subsubsection{Null Baseline Construction}
\label{sec:diag_null_build}

To provide a parameter-free baseline without introducing global distributional assumptions, each target edit region $\mathcal{R}_{\mathrm{tgt}}$ is paired with $K$ \emph{same-size} random background blocks drawn from the same image. We maintain two distinct null budgets according to computational cost:
\begin{itemize}[leftmargin=*]
\item \textbf{Metric nulls} ($K_{\mathrm{metric}}$ blocks) supply the sample-specific baseline for representation, routing, and sensitivity metrics (Gates 1--4). They are read directly off the \emph{same} feature, attention, and gradient tensors by permuting token indices, incurring zero additional forward passes.
\item \textbf{Ablation nulls} ($K_{\mathrm{abl}} \ll K_{\mathrm{metric}}$ blocks) supply the null distribution for the intervention \texttt{effect}. Because each block requires an independent forward pass, only a small budget is evaluated for the null self-check. Importantly, this budget affects \emph{only} the baseline verification (Sec.~\ref{sec:diag_null_check}) and does not alter the deterministic real-region gates, where Gate 5 relies solely on $\mathrm{sign}(\mathrm{effect})$.
\end{itemize}

\subsubsection{Deterministic Decision Chain}
\label{sec:diag_chain}

Using the sample-wise null baselines constructed above, the diagnostic applies a deterministic, parameter-free rule independently to each counterfactual failure. Specifically, every metric calculated on the real edit region $\mathcal{R}_{\mathrm{tgt}}$ is compared directly against the empirical mean of the $K_{\mathrm{metric}}$ random background blocks from the \emph{same sample} (denoted as $\overline{(\cdot)}^{\,\mathrm{null}}$).

The decision chain operates as a strict sequence of elimination steps, given as Algorithm~\ref{alg:diag_chain}.

\begin{algorithm}[t]
\caption{Deterministic layer-wise attribution of a counterfactual failure.}
\label{alg:diag_chain}
\begin{algorithmic}[1]
\Require eligible pair $(I^{O},I^{CF},q)$; target region $\mathcal{R}_{\mathrm{tgt}}$ with token set $\mathcal{T}_{\mathrm{edit}}$; budgets $K_{\mathrm{metric}},K_{\mathrm{abl}}$
\Ensure one label from Tab.~\ref{tab:diag_labels}
\If{$\mathrm{Grid}(I^{O})\neq\mathrm{Grid}(I^{CF})$ or $|\mathcal{T}_{\mathrm{edit}}|>0.5\,|\mathcal{T}_{\mathrm{vis}}|$}
    \State \textbf{return} \diaglab{out\_of\_scope} \Comment{Step 0: region not localizable}
\EndIf
\State draw $K_{\mathrm{metric}}$ same-size random blocks; compute their per-metric means $\overline{(\cdot)}^{\,\mathrm{null}}$
\State compute $\phi_{1},r_{\mathrm{flow}},A_{\mathrm{lift}},\mathrm{Sal}_{\mathrm{norm}}$ on $\mathcal{T}_{\mathrm{edit}}$
\If{$\phi_{1}$ or $\overline{\phi_{1}}^{\,\mathrm{null}}$ unavailable}
    \State \textbf{return} \diaglab{undetermined} \Comment{Gate 1 cannot be evaluated}
\EndIf
\If{$\phi_{1} \le \overline{\phi_{1}}^{\,\mathrm{null}}$}
    \State \textbf{return} \diaglab{feature\_collapse} \Comment{the encoder did not encode the edit}
\EndIf
\If{$r_{\mathrm{flow}}, \overline{r_{\mathrm{flow}}}^{\,\mathrm{null}}$ available \textbf{and} $r_{\mathrm{flow}} \le \overline{r_{\mathrm{flow}}}^{\,\mathrm{null}}$}
    \State \textbf{return} \diaglab{alignment\_fail} \Comment{the edit is not routed to the answer position}
\EndIf
\If{$A_{\mathrm{lift}}, \overline{A_{\mathrm{lift}}}^{\,\mathrm{null}}$ available \textbf{and} $A_{\mathrm{lift}} \le \overline{A_{\mathrm{lift}}}^{\,\mathrm{null}}$}
    \State \textbf{return} \diaglab{attention\_failure} \Comment{the answer position ignores the region}
\EndIf
\If{$\mathrm{Sal}_{\mathrm{norm}}$ or $\overline{\mathrm{Sal}_{\mathrm{norm}}}^{\,\mathrm{null}}$ unavailable}
    \State \textbf{return} \diaglab{undetermined} \Comment{Gate 4 cannot be evaluated (no gradient)}
\EndIf
\If{$\mathrm{Sal}_{\mathrm{norm}} \le \overline{\mathrm{Sal}_{\mathrm{norm}}}^{\,\mathrm{null}}$}
    \State \textbf{return} \diaglab{language\_prior} \Comment{the decision does not depend on the region}
\EndIf
\State $\mathrm{effect} \gets P^{CF}(\text{correct})-P^{\mathrm{abl}}(\text{correct})$ \Comment{mean-ablation on $\mathcal{T}_{\mathrm{edit}}$}
\If{$\mathrm{effect}$ unavailable}
    \State \textbf{return} \diaglab{undetermined} \Comment{Gate 5 cannot be evaluated}
\EndIf
\If{$\mathrm{effect}>0$}
    \State \textbf{return} \diaglab{reasoning\_error} \Comment{vision supports the correct answer, but is overridden}
\Else
    \State \textbf{return} \diaglab{understanding\_error} \Comment{vision pushes the decision to the wrong answer}
\EndIf
\Statex
\Statex \textit{Offline self-check:} rerun the chain above on every random block, using the remaining blocks as leave-one-out nulls (Sec.~\ref{sec:diag_null_check}).
\end{algorithmic}
\end{algorithm}

Tab.~\ref{tab:diag_labels} provides a summary of the full label hierarchy. A pair that fails the Step~0 gates is \diaglab{out\_of\_scope} rather than \diaglab{undetermined}, since its region is by construction not localizable. Within the cascade, a missing statistic is handled in one of two ways:
\begin{itemize}[leftmargin=*]
\item \textbf{Strict break.} If the Gate~1 statistic, the Gate~4 gradient, or the Gate~5 intervention is unavailable, the chain cannot be completed and the pair is labelled \diaglab{undetermined}.
\item \textbf{Soft bypass.} If a Gate~2 or Gate~3 statistic is unavailable, that gate is skipped and the chain continues, so a pair can in principle receive an L3 label without the L2 layer having been tested. The asymmetry is deliberate: an unreadable L2 statistic is treated as no evidence against the L2 explanation, whereas a missing gradient or intervention leaves the elimination chain unfinished. Algorithm~\ref{alg:diag_chain} encodes both rules.
\end{itemize}

Note that \diaglab{language\_prior} here is a decision-layer label derived from \emph{local} sensitivity to the edited region; it is distinct from the image-free text-only behavior in the main evaluation, which is recorded as an independent global stratification variable.

\begin{table}[t]
\centering
\caption{\textbf{Diagnostic label set.} Each label names the first elimination step that the sample fails, i.e.\ the earliest stage at which the edited visual evidence stops influencing the answer.}
\label{tab:diag_labels}
\begin{tabular}{lll}
\toprule
\textbf{Label} & \textbf{Stage} & \textbf{Meaning} \\
\midrule
\diaglab{feature\_collapse}     & L1 (Gate 1) & encoder did not encode the edit \\
\diaglab{alignment\_fail}       & L2 (Gate 2) & edit not routed to the answer position \\
\diaglab{attention\_failure}    & L2 (Gate 3) & answer position does not look at the region \\
\diaglab{language\_prior}       & L3 (Gate 4) & decision does not rely on the region \\
\diaglab{reasoning\_error}      & L3 (Gate 5) & vision helps the correct answer; decision overrides it \\
\diaglab{understanding\_error}  & L3 (Gate 5) & vision misleads the decision \\
\diaglab{undetermined}          & fallback    & required statistic unavailable \\
\diaglab{out\_of\_scope}        & Step 0      & grid or locality gate not satisfied \\
\bottomrule
\end{tabular}
\end{table}

\subsubsection{Baseline Self-Inspection}
\label{sec:diag_null_check}

Because each metric rule takes the form ``$\le \text{null mean}$'', a random non-informative rule would naturally fire on roughly half of the null blocks. We therefore validate the diagnostic by applying the \emph{identical} decision chain to the random background blocks themselves, using a leave-one-out procedure among the $K$ blocks to establish a null label distribution. The true diagnostic conclusion is defined by the \emph{departure} of real edit regions from this background distribution, rather than raw uncalibrated label counts. The comparison is stated at the level of the \emph{region}, not of the failure: a random block has no notion of being correct or wrong, so the null pool cannot be conditioned on failure, and the real side of the comparison is the in-scope pool. Tab.~\ref{tab:diag_nullcheck} reports both distributions, Tab.~\ref{tab:diag_gates} shows that every threshold gate fires less often on the real edit region than on a random block of the same size (a conditional ratio below one), and Tab.~\ref{tab:diag_decisions} reports the two quantities that describe the chain as a whole. Two cautions apply when reading the first of these. The ratios of Tab.~\ref{tab:diag_nullcheck} are \emph{share} ratios --- the end state of a cascade --- and not gate effectiveness; the cascade removes null blocks and real regions at very different rates, so a late-gate share can point the opposite way from that gate's own conditional ratio, and only Tab.~\ref{tab:diag_gates} answers the question of whether a gate is informative. And for the labels assigned at Gate~5 the comparison must use the \emph{effect-evaluable} null column, since the remaining blocks cannot be scored there and would otherwise enter the denominator.


\begin{table}[t]
\centering
\caption{\textbf{Label distribution: real vs.\ null.} The null columns come from running the \emph{identical} chain on leave-one-out same-size random blocks. Random blocks are localizable by construction, whereas the \emph{all} pool contains the \diaglab{out\_of\_scope} pairs, so the like-for-like comparison is against the \emph{in-scope} columns; the \emph{gated} columns are the failure-conditioned view. The two null columns differ in their denominator: the right one keeps only the blocks whose downstream statistic exists (25\% of the pool has no ablation effect), and for the Gate~5 labels (\diaglab{reasoning\_error}, \diaglab{understanding\_error}) it is the appropriate comparison, since the \emph{all} column would divide by blocks that cannot be scored there. Percentages are of each pool; $n$ is given in the last row.}
\label{tab:diag_nullcheck}
\resizebox{\linewidth}{!}{
\begin{tabular}{lcccccc}
\toprule
\textbf{Label} & Real (all) & Real (in-scope) & Real (gated) & Real (in-scope $\cap$ gated) & \textbf{Null (all)} & \textbf{Null (eff.-eval.)} \\
\midrule
\diaglab{feature\_collapse} & 11.0\% & 13.7\% & 11.7\% & 14.4\% & 52.9\% & 52.9\% \\
\diaglab{alignment\_fail} & 10.8\% & 13.4\% & 12.3\% & 15.2\% & 30.2\% & 29.9\% \\
\diaglab{attention\_failure} & 20.2\% & 25.0\% & 20.5\% & 25.3\% & 7.9\% & 8.3\% \\
\diaglab{language\_prior} & 3.1\% & 3.9\% & 3.6\% & 4.5\% & 3.4\% & 3.2\% \\
\diaglab{reasoning\_error} & 19.5\% & 24.1\% & 15.8\% & 19.5\% & 0.7\% & 2.8\% \\
\diaglab{understanding\_error} & 16.1\% & 19.9\% & 17.2\% & 21.2\% & 0.7\% & 2.9\% \\
\diaglab{out\_of\_scope} & 19.1\% & 0.0\% & 18.8\% & 0.0\% & 0.0\% & 0.0\% \\
\diaglab{undetermined} & 0.0\% & 0.0\% & 0.0\% & 0.0\% & 4.1\% & 0.0\% \\
\midrule
\emph{$n$} & 1240 & 1003 & 633 & 514 & 36617 & 9230 \\
\bottomrule
\end{tabular}}
\end{table}

\begin{table}[t]
\centering
\caption{\textbf{Gate effectiveness.} \emph{Conditional} firing rate of each threshold gate: the fraction of the samples that \emph{reach} the gate for which it fires. Raw (unconditional) rates are not comparable across the gates of a cascade --- each gate runs only on the survivors of the previous ones --- and can even invert the sign of the effect. Real: target region, in-scope $n=1003$; null: leave-one-out same-size random blocks, $n=36617$. A ratio below one means the real edit region triggers the gate less often than a random block, i.e.\ the gate is informative.}
\label{tab:diag_gates}
\begin{tabular}{llccc}
\toprule
\textbf{Gate} & \textbf{Statistic} & \textbf{Real} & \textbf{Null} & \textbf{Real/null} \\
\midrule
1 (L1) & $\max(d_{\cos}^{\mathrm{vis}},d_{\mathrm{rel}}^{\mathrm{vis}})$ & 13.66\% & 52.89\% & 0.258 \\
2 (L2) & $r_{\mathrm{flow}}$ & 15.47\% & 64.16\% & 0.241 \\
3 (L2) & $A_{\mathrm{lift}}$ & 34.29\% & 46.83\% & 0.732 \\
4 (L3) & $\mathrm{Sal}_{\mathrm{norm}}$ & 8.11\% & 38.42\% & 0.211 \\
\bottomrule
\end{tabular}
\end{table}

\begin{table}[t]
\centering
\caption{\textbf{Whole-chain quantities.} Neither row is a gate's own performance. \emph{Reach} is the product of Gates~1--4 and carries no information about Gate~5; \emph{labels} additionally counts the Gate~4 firings. Gate~5 has no null baseline --- its only statistic is the sign split, which is the weakest in the chain and is therefore indicative rather than conclusive.}
\label{tab:diag_decisions}
\resizebox{\linewidth}{!}{
\begin{tabular}{lccc}
\toprule
\textbf{Quantity} & \textbf{Real} & \textbf{Null (all)} & \textbf{Null (eff.-eval.)} \\
\midrule
\multicolumn{4}{l}{\emph{The cascade as a whole (not a gate's own performance)}} \\
Decision-layer \textbf{reach} (passes Gates 1--4) & 44.07\% & 5.53\% & 5.67\% \\
Decision-layer \textbf{labels} (Gate 4 fires \emph{or} Gate 5 labels) & 47.96\% & 4.88\% & 8.89\% \\
\midrule
\multicolumn{4}{l}{\emph{Gate 5: a terminal classifier, not a threshold gate}} \\
Sign split $\mathrm{effect}>0 : \mathrm{effect}<0$ & 242 : 199 & 254 : 264 & --- \\
Odds of a positive effect, $\mathrm{pos}/\mathrm{neg}$ & 1.216 & 0.962 & --- \\
\emph{$n$} with a usable ablation effect & 442 & --- & 523 \\
\bottomrule
\end{tabular}}
\end{table}

\newcommand{\caseIdFeature}{\texttt{VOC\_007}}
\newcommand{\caseIdAlign}{\texttt{SRP\_003}}
\newcommand{\caseIdAttn}{\texttt{DSSC\_001}}
\newcommand{\caseIdLang}{\texttt{SAS\_022}}
\newcommand{\caseIdReason}{\texttt{SRE\_031}}
\newcommand{\caseIdUnderst}{\texttt{EAT\_021}}

\subsection{Diagnostic Examples}

The three cases below are the exhaustive attributions of a single behavioural signature --- a counterfactual failure --- rather than three distinct black-box phenomena; their population shares are in Tab.~\ref{tab:diag_nullcheck}. Tab.~\ref{tab:diag_cases} instantiates every label of the chain on one benchmark pair each, giving the pair's id together with the readings the chain actually judged, so every claim below can be checked against a single row.


\begin{table}[t]
\centering
\caption{\textbf{One representative failure per label.} Each row is a single benchmark pair, identified by its \texttt{id}, and reports the \emph{deciding} statistic of that pair together with its leave-one-out null baseline: ``real'' is the target edit region, ``null'' the mean over the other random blocks of the same pair. The pair shown for a label is the one whose deciding gate fired by the widest margin within the in-scope $\cap$ gated pool, so each row is the clearest instance of that failure mode rather than a typical one. Margins are shortfalls from each gate's own null and are therefore comparable \emph{only within a label}: the deciding statistic has different units in every row (a feature distance for Gate~1, a write-in ratio for Gate~2, an attention share for Gate~3, a gradient norm for Gate~4, a probability difference for Gate~5), so the column is not a ranking. Gate~5 labels by the sign of the intervention effect rather than by a threshold test and has no baseline column (shown as ``--''); its margin is $|\mathrm{effect}|$. All numbers are produced by \texttt{diagnosis/regate.py}; ``real'', ``null'' and ``margin'' are the readings the chain actually judged.}
\label{tab:diag_cases}
\resizebox{\linewidth}{!}{
\begin{tabular}{llllccc}
\toprule
\textbf{Stage} & \textbf{Label} & \textbf{Pair id} & \textbf{Deciding stat.} & \textbf{Real} & \textbf{Null} & \textbf{Margin (within-label)} \\
\midrule
L1 & \diaglab{feature\_collapse} & \texttt{VOC\_007} & $\max(d_{\cos}^{\mathrm{vis}},d_{\mathrm{rel}}^{\mathrm{vis}})$ & 0.125 & 0.714 & 0.589 \\
L2 & \diaglab{alignment\_fail} & \texttt{SRP\_003} & $r_{\mathrm{flow}}$ & 0.107 & 0.477 & 0.370 \\
L2 & \diaglab{attention\_failure} & \texttt{DSSC\_001} & $A_{\mathrm{lift}}$ & 0.318 & 2.85 & 2.54 \\
L3 & \diaglab{language\_prior} & \texttt{SAS\_022} & $\mathrm{Sal}_{\mathrm{norm}}$ & 0.204 & 0.626 & 0.422 \\
L3 & \diaglab{reasoning\_error} & \texttt{SRE\_031} & $\mathrm{effect}$ & 0.833 & -- & 0.833 \\
L3 & \diaglab{understanding\_error} & \texttt{EAT\_021} & $\mathrm{effect}$ & -0.951 & -- & 0.951 \\
\bottomrule
\end{tabular}}
\end{table}

\paragraph{Case 1 --- visual-level failure (\diaglab{feature\_collapse}): the edit is in the pixels, the answer is unchanged.}
The sharpest instance is \caseIdFeature{} (Tab.~\ref{tab:diag_cases}): its region changes less than a same-size random block does. Gate~1 fires on the encoder output, before any language-side computation, so the failure precedes the decision. \textbf{Mechanism.} The change measured over the annotated region does not exceed what a same-size random block yields, i.e.\ the edit is never represented --- as opposed to represented and then unused, which is the attentional failure of Case~2, and which no saliency map can distinguish from it on its own. The claim is representational, not semantic.

\paragraph{Case 2 --- interaction-level bottleneck (\diaglab{alignment\_fail}, \diaglab{attention\_failure}): the edit is encoded, the answer is unchanged.}
The sharpest instances are \caseIdAlign{} and \caseIdAttn{} (Tab.~\ref{tab:diag_cases}), whose region is written into and read out of the answer position, respectively, less than a random block is. Gate~1 passes, and Gate~2 or~3 fires. \textbf{Mechanism.} For a fixed answer triple a perception failure and a routing failure are observationally equivalent, and Gate~1 eliminates the former; a language-prior account is eliminated by position in the chain, since the diagnosis stops before L3 and therefore reports that the region was not written into the answer position rather than that the decision declined to use it. Gate~2 separates the write from the read. The claim concerns the route, not the content of the evidence.

\paragraph{Case 3 --- decision-level misalignment (\diaglab{language\_prior}, \diaglab{reasoning\_error}, \diaglab{understanding\_error}): the edit is encoded and routed, yet the answer is still wrong.}
The sharpest instances are \caseIdLang{}, \caseIdReason{} and \caseIdUnderst{} (Tab.~\ref{tab:diag_cases}). \textbf{Mechanism.} L3 has two sub-cases. Gate~4 fires when the decision's local sensitivity to the region does not exceed the null, i.e.\ the answer was formed without depending on the region at all (\diaglab{language\_prior}). If Gate~4 passes, Gate~5 labels by the sign of the mean-ablation effect: removing the region lowering the probability of the correct answer means the evidence supported it and the aggregation discarded it (\diaglab{reasoning\_error}), while raising it means the evidence pushed the decision away (\diaglab{understanding\_error}). The two Gate-5 labels have identical predictions, so only the sign of a causal contribution separates them. Gate~5 is a classifier rather than a threshold and therefore has no null baseline, its margin over the null being the smallest in the chain (Tab.~\ref{tab:diag_decisions}).

\paragraph{Caveats.} Two properties of the framework limit the strength of its claims and are reported explicitly. \emph{(i) Region precision.} The target region is the union of the two annotation boxes, hence a superset of the changed pixels; L1 magnitudes are diluted accordingly. This cannot be removed without pixel-level annotation and is absorbed by the same-size null baseline. \emph{(ii) Resolution dependence.} Attention- and gradient-based quantities degrade when the region contains only a handful of tokens, so all measurements are taken at the native image resolution (matching the main Tab.); conclusions are stated per model/resolution operating point and are not transferred across resolutions.

\section{Benchmark Details}
\label{app:benchmark}

We construct a paired visual question-answering benchmark to examine whether spatial judgments follow image evidence when familiar spatial regularities are disrupted. Each instance contains an original aerial image $I_i^{\mathrm{O}}$, an edited counterfactual image $I_i^{\mathrm{CF}}$, a shared question $q_i$, four shared answer options $\mathcal{A}_i$, and image-specific ground-truth labels $y_i^{\mathrm{O}}$ and $y_i^{\mathrm{CF}}$:
\begin{equation}
    d_i = \bigl(I_i^{\mathrm{O}}, I_i^{\mathrm{CF}}, q_i,
    \mathcal{A}_i, y_i^{\mathrm{O}}, y_i^{\mathrm{CF}}, t_i, b_i\bigr),
\end{equation}
where $t_i$ denotes the spatial task and $b_i$ denotes the hypothesized source of bias. Holding the question and answer options fixed supports a direct comparison of predictions across the two visual conditions.

The taxonomy crosses six spatial tasks, including spatial position, spatial relationship, spatial action, geometric structure, depth and scale, and viewpoint and orientation, with four bias-reason labels: cultural and behavioral norms, physical-geometric principles, semantic co-occurrence, and egocentric--allocentric transformation.
These labels specify intended construction contrasts and do not by themselves establish the internal cause of a model's error. The final release contains 1,254 pairs (2,508 image files), with all 24 task--reason combinations represented, as shown in Tab.~\ref{tab:spatial-bias-composition}.

\begin{table}[!ht]
\centering\small
\caption{Final benchmark composition in image pairs. Columns denote cultural/behavioral norms (Cul.), physical-geometric principles (Phy.), semantic co-occurrence (Sem.), and egocentric--allocentric transformation (Ref.).}
\label{tab:spatial-bias-composition}
\begin{tabular}{lrrrrr}\toprule
Spatial task & Cul. & Phy. & Sem. & Ref. & Total\\
\midrule
Spatial Position & 65 & 56 & 59 & 54 & 234 \\
Spatial Relationship & 58 & 52 & 53 & 52 & 215 \\
Spatial Action & 51 & 54 & 51 & 50 & 206 \\
Geometric Structure & 55 & 53 & 46 & 48 & 202 \\
Depth and Scale & 49 & 38 & 49 & 51 & 187 \\
Viewpoint and Orientation & 53 & 55 & 48 & 54 & 210 \\
\midrule Total & 331 & 308 & 306 & 309 & 1,254\\\bottomrule
\end{tabular}
\end{table}

\subsection{Dataset Construction Details}
\label{app:data_cons}

\subsubsection{Source Imagery}
The benchmark is primarily constructed from \texttt{VisDrone}~\citep{du2019visdrone}, \texttt{M3OT}~\citep{nie2025m3ot}, \texttt{UAVDT}~\citep{du2018uavbenchmark}, and \texttt{MDMT}~\citep{liu2023robust}. Existing annotations are used to localize candidate targets and support target identification during counterfactual construction. Target descriptions rely on attributes that remain valid after editing, while spatial cues affected by the intervention are avoided. Annotations are used only for construction and are not rendered as boxes or markers in the released benchmark images.

\subsubsection{Data Construction}

\paragraph{Coverage and target selection.}
Construction is driven by coverage of the six spatial tasks and four prior categories. The recorded expansion stages increase the candidate pool from 457 to 587, 817, and 1,297 pairs, bringing each combination to at least 20, 30, and 50 candidates, respectively. Within a category, source annotations localize an identifiable target and, where needed, reference objects. Task-specific specifications record the spatial variable, target scope, original and counterfactual states, shared question, options, and answer mapping. Target descriptions use attributes that remain identifiable after editing. 

\paragraph{Task-specific selection criteria.}
Position questions require distinguishable image regions; relationship questions require an identifiable target and reference; action questions require observable maneuver geometry; structure questions require an identifiable object group; depth/scale questions require a stated visual comparison criterion; and orientation questions require an explicit coordinate frame and a distinguishable object axis. The prior category specifies the intended conflict, such as a conventional layout, a physical regularity, a semantic association, or a reference-frame transformation. These labels describe construction hypotheses, not experimentally established causes of individual errors.

\textbf{Task specification and counterfactual intervention.}
Each sample is constructed according to a task-specific specification defining the target spatial variable, original and counterfactual states, target scope, question, candidate options, and answer mapping. The counterfactual transformation is formulated as
\begin{equation}
    I_i^{\mathrm{CF}} = \mathcal{T}\!\left(I_i^{\mathrm{O}};
    s_i \leftarrow s_i'\right),
\end{equation}
where $s_i$ denotes the target spatial factor and $\mathcal{T}$ the corresponding editing operation. The intervention may involve a single object, an object--reference relation, or a multi-object configuration. Camera viewpoint, image crop, object identity, and task-irrelevant scene content are preserved as much as possible, with the intervention directed at the spatial factor required by the target task. This specifies the editing objective rather than a guarantee that every non-target pixel is unchanged. The six task categories cover absolute position, pairwise spatial relations, spatial actions, geometric structure, depth/scale, and viewpoint/orientation.

{\color{black}

\paragraph{Counterfactual Image Editing and Quality Control.}
We employ three complementary editing approaches: (1) an instruction-guided generative editor, (2) a deterministic compositor implemented in Pillow and SciPy, and (3) manual editing in Photoshop. Among the 1,254 retained pairs, 848 and 235 samples are produced by the first two automated pipelines, respectively, while the remaining 171 are edited manually.

Generative edits follow sample-specific instructions that specify the target object, intended spatial transformation, and semantic preservation constraints. For deterministic edits, the compositor extracts a target patch from the spatial annotation, fills the vacated region using surrounding valid pixels, and blends the transformed patch using a softened mask. Given target width $w$ and height $h$, the erasure padding is set to $\max(10,\lfloor 0.35\max(w,h)\rfloor)$ pixels, while Gaussian smoothing and mask feathering use radii of $\max(1,0.025\min(w,h))$ and $\max(1,0.035\min(w,h))$, respectively. Scaling uses Lanczos resampling, and rotation uses bicubic resampling.

To suppress unintended artifacts, we verify target identifiability, intervention correctness, preservation of unrelated scene elements, and alignment between the edited spatial layout and ground-truth labels. Because visual degradations directly impair downstream embodied reasoning rather than merely affecting perceptual appearance~\cite{zha2026task}, we also screen out low-clarity scenes and conduct human validation to check visual interpretability and annotation reliability (Appendix~\ref{app:human_perform}). While editing can introduce localized visual changes outside the target region, these checks limit confounding effects without requiring exact pixel-level invariance.
}

\textbf{Question construction.}
Each original--counterfactual pair shares one four-option question. The construction instructions require mutually exclusive distractors that express competing spatial states with parallel wording and a common reference frame. The intended protocol checks the answer mapping against the realized images, rather than relying solely on the planned edit, and avoids wording that reveals which image is counterfactual. Option permutations are applied consistently to both image-specific labels. In the final twenty-pair expansion batches, each answer position occurs five times per image condition. The instructions discourage unsupported assumptions about speed, intent, jurisdiction-specific rules, or metric depth; historical intent-based examples are discussed separately below. 
The intended construction satisfies $y_i^{\mathrm{O}}\neq y_i^{\mathrm{CF}}$. Among the 1,254 retained pairs, 1,251 satisfy this condition, while \texttt{SAC\_007}, \texttt{SAC\_009}, and \texttt{SPE\_004} retain identical answer labels. Answer positions are approximately balanced: the A--D counts are $(308,316,314,316)$ for original images and $(319,316,309,310)$ for counterfactual images.

\subsubsection{Construction Checks and Model-Based Curation}
\label{app:benchmark_curation}

\paragraph{Construction checks.}
The construction protocol checks target identifiability, realization of the requested spatial change, task consistency, answer validity, image dimensions, and file integrity. Prompts request inspection for duplicated objects, residual target footprints, boundary artifacts, and unintended changes to unrelated content. These requirements guide construction. We distinguish these checks from model-conditioned curation and the subsequent human answer audit.


{\color{black}
\subsubsection{Human Annotation Audit}
\label{app:human_answer_audit}

\paragraph{Coverage and procedure.}
Three human reviewers each completed all original--counterfactual pairs, yielding 3,762 pair-level reviews and 7,524 image-level answers. 
Reviewers submit an answer for each image before the interface reveals the stored labels, after which they can assess label validity, flag ambiguity, suggest corrections, and provide comments. Thus, the initial answer-entry and label-review stages are separated. The interface also permits re-answering. Human answers are collected separately from the predictions used for model-based curation.


\paragraph{Analysis and agreement.}
We align the three reviewers' records by sample ID and analyze their recorded human answers. 
Human accuracy reaches 92.88\% on original images and 93.04\% on counterfactual images, with high inter-annotator agreement (Fleiss' $\kappa$ = 0.866/0.858). Only 1.52\% of pairs were flagged as potentially ambiguous by a majority of annotators. These results support the reliability of the annotations and indicate that counterfactual edits remain visually interpretable to humans. 
%
%
%
The similar mean human accuracies between O and CF samples also provide evidence that the edited images remain broadly answerable relative to the stored labels.

\paragraph{Ambiguity and scope of validation.}
Across the three reviewers, 4.39\% pair-level reviews contain an ambiguity flag. At least two flag 19/1,254 (1.52\%), and all three flag 1/1,254 (0.08\%). 
%
The audit supplies human answer-consistency evidence beyond model-based filtering. It does not separately quantify intervention correctness, visual-artifact severity, or preservation of non-target semantics. Consequently, answer agreement should not be interpreted as an artifact-free rate. A dedicated intervention and visual-quality evaluation is needed to measure these dimensions and determine whether model conclusions persist on adjudicated, visually valid subsets. For physical-prior interventions, an intentionally unusual configuration must also be distinguished from an unintended rendering defect.
\FloatBarrier
}

\subsubsection{Construction Prompts}
The construction prompt contains a general system prompt, a per-image user template, task-specific modules, few-shot examples, and successive expansion prompts. We provide a condensed, normalized rendering below to expose the common protocol. These templates summarize the prompt family rather than implying identical wording for every historical sample; sample-specific specifications determine the target and requested spatial states. 

\paragraph{Consolidated construction protocol.}
The protocol below summarizes the common requirements of the prompt family.
\begin{lstlisting}[style=constructionprompt]
ROLE
Construct paired original/counterfactual aerial-image questions that test visible spatial evidence under a specified prior conflict.

INPUT
Source image; optional detection annotations; spatial task; bias reason; target/reference or target-group specification; intended original and counterfactual states; output ID; answer-position schedule; optional few-shot schema examples.

PROCEDURE
1. Inspect the source and annotations. Ground an identifiable target and any unchanged reference. Use a description that remains valid in both images. Reject an ambiguous target.
2. Preserve the original. Edit the supplied image rather than synthesizing a replacement scene. Intervene on the designated spatial factor of the target or target group. Preserve camera, crop, identity and unrelated content as far as the editing process permits.
3. Remove the old target footprint when relocating it. Avoid duplication, ghosting, unintended changes, and annotation overlays. Preserve lighting, support, perspective and occlusion unless the specified factor tests that property.
4. Inspect the realized edit. Determine the after-state from the pixels. Retry or reject if the required distinction is not visible or the target cannot be identified.
5. Write one shared four-option single-choice question. Use parallel, non-overlapping options and an explicit reference frame where required. Do not mention original/counterfactual status. Avoid unsupported claims about motion, intent, legality or metric depth.
6. Assign one answer per image. Require an answer change for newly constructed pairs. Apply the supplied option permutation to the options and both answer labels consistently.
7. Return the core record and available provenance/QC metadata. Report a rejection reason if construction fails; do not claim a nonexistent edited file.

CORE RECORD
id; bias_spatial; bias_reason; O_image_path; CF_image_path;
question; question_type="single-choice"; options={A,B,C,D};
O_answer; CF_answer; metadata.

PROVENANCE WHEN AVAILABLE
Source dataset/frame/annotation; target and reference grounding;
changed factor; original/counterfactual predicates; editor prompt;
editor identity; dimensions; checks; hashes; rejected attempts.
\end{lstlisting}

\paragraph{Task-specific modules.}
This module specifies the question's spatial predicate and guards against conflating adjacent tasks.
\begin{lstlisting}[style=constructionprompt]
SPATIAL POSITION
Assign a target to an absolute image region or a named scene region. Avoid turning the question into a binary target/reference relation.

SPATIAL RELATIONSHIP
Ask for a target's relation to a fixed reference, such as image-plane direction, separation, contact, support or overlap.

SPATIAL ACTION
Ask for visibly depicted maneuver geometry or bodily action. A still image alone does not prove velocity, future intent or a temporal trajectory.

GEOMETRIC STRUCTURE
Classify a multi-object global configuration or a specified structural motif. Do not substitute a single object's position or orientation.

DEPTH AND SCALE
State the comparison rule: projected footprint or visibly supported depth ordering. Do not equate image size with metric distance.

VIEWPOINT AND ORIENTATION
Express the object's physical front or body axis in a declared image/scene frame. Preserve the camera when intervening on object orientation.
\end{lstlisting}

\paragraph{Prior-source modules.}
One module specifies the intended prior contrast. Its label is a construction hypothesis rather than an independently verified cognitive mechanism.
\begin{lstlisting}[style=constructionprompt]
CULTURAL AND BEHAVIORAL NORMS
Contrast a common scene convention with a visibly different state. Ask about spatial evidence, not an unsupported legal or moral judgment. Do not invent jurisdiction-specific rules.

PHYSICAL-GEOMETRIC PRINCIPLES
Manipulate the designated support, contact, scale, perspective or orientation variable. Some historical examples deliberately create strong physical conflicts; later prompts prefer milder anomalies.

SEMANTIC CO-OCCURRENCE
Retain scene/category cues while making the tested object-layout, object-action or object-scale association less typical. The answer must follow the visible state rather than the usual association.

EGOCENTRIC-ALLOCENTRIC TRANSFORMATION
Require a mapping between object/scene coordinates and fixed image coordinates. Name the reference frame explicitly and ground its axes in visible evidence.
\end{lstlisting}

\paragraph{Per-image editing template.}
The abstract template is instantiated with the selected target, spatial states, and preservation constraints.
\begin{lstlisting}[style=constructionprompt]
Edit the supplied image. Target: <stable target/group descriptor and source bounding box or annotation rows>.
Change only <designated spatial variable> from <observed state> to <requested state>.
Keep <fixed reference, camera, crop, identity, count and non-target context> unchanged. Reconstruct the vacated region and any required local contact/shadow details.
Do not add labels, boxes, arrows, text, duplicate targets or unrelated objects. Output one edited image for inspection.
\end{lstlisting}

\paragraph{Exampled editor instructions.}
Following shows several representative editor instructions for different kind of data generation.

\noindent\textbf{Spatial position: \texttt{SPC\_002}.}
\begin{lstlisting}[style=constructionprompt]
Move only the blue upper-left taxi diagonally downward and slightly right, without rotating it, until it clearly straddles both solid yellow centerlines. Reconstruct the old asphalt; keep the other taxi, road markings, cracks, camera, lighting, crop, and aspect ratio unchanged.
\end{lstlisting}
\noindent\textbf{Spatial relationship: \texttt{SRP\_001}.}
\begin{lstlisting}[style=constructionprompt]
Move only annotation row 85 onto the roof of fixed row 86; retain both colors, sizes, headings and visible silhouettes.
\end{lstlisting}
\noindent\textbf{Spatial action: \texttt{SAC\_003}.}
\begin{lstlisting}[style=constructionprompt]
Rotate only the green bus at [933,322,349,197] into an upward left-turn pose within the open junction, translating minimally to avoid all fixed cars, pedestrians, crossings, and the traffic island.
\end{lstlisting}
\noindent\textbf{Geometric structure: \texttt{GSSC\_005}.}
\begin{lstlisting}[style=constructionprompt]
Target only annotation rows 3,4,5,10. Reposition those four cars into a clear compact 2-by-2 rectangular valet-staging array, two rows of two with safe gaps, while preserving colors/types, perspective, shadows, camera and non-target content.
\end{lstlisting}
\noindent\textbf{Depth and scale: \texttt{DSC\_001}.}
\begin{lstlisting}[style=constructionprompt]
Enlarge only the lowest white vehicle in the central chain near [542,217,45,42] to about 170x125 pixels; preserve its center, heading, color, category, ground contact, the silver reference car, and all other content.
\end{lstlisting}
\noindent\textbf{Viewpoint and orientation: \texttt{EAT\_002}.}
\begin{lstlisting}[style=constructionprompt]
Use case: precise-object-edit. Image 1 is the original edit target. Rotate ONLY the large motion-blurred white panel van in the lower-left intersection at source bbox [259,403,268,213] by exactly 180 degrees in yaw around its current image-plane center. Make the cab, windshield and front point unmistakably toward the UPPER-RIGHT, with the cargo rear toward the lower-left. Preserve the same plain white panel-van category, body proportions, projected area, center, scene depth, wheels-on-asphalt support, aerial perspective, natural motion-blur character, daylight and coherent shadow. Preserve every other car, pedestrian, signal, crossing, barrier, tree, storefront, camera and crop. Reconstruct minimal exposed asphalt. Do not translate, resize, deform, duplicate, remove, add logos or readable text, rotate the whole image or globally restyle. Output one photorealistic edited image only.
\end{lstlisting}

\subsection{Dataset Examples}
\label{app:dataset_examples}

\begin{figure}[htbp]
\centering
\begin{minipage}[t]{0.49\linewidth}\centering
\textbf{Original (O)}\\[3pt]
\includegraphics[width=\linewidth,height=0.205\textheight,keepaspectratio]{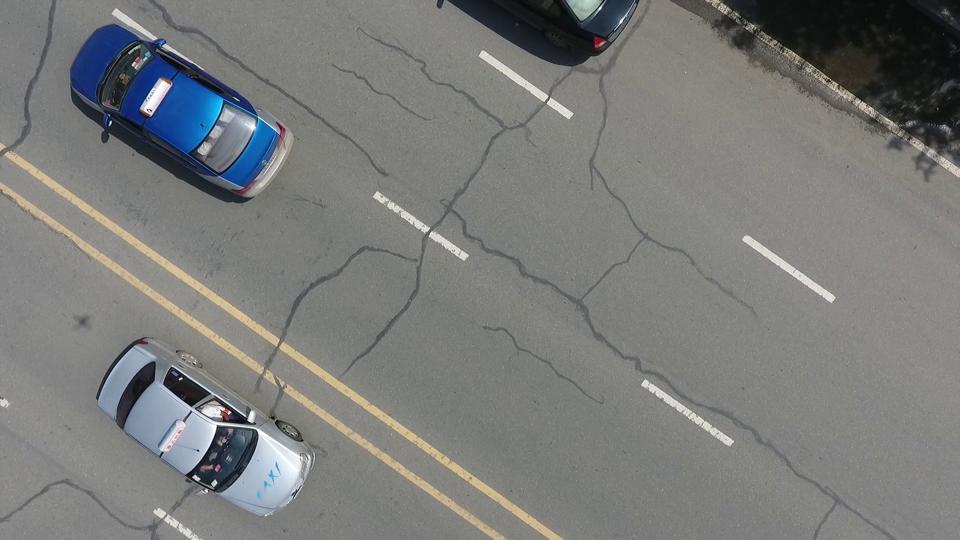}
\end{minipage}\hfill
\begin{minipage}[t]{0.49\linewidth}\centering
\textbf{Counterfactual (CF)}\\[3pt]
\includegraphics[width=\linewidth,height=0.205\textheight,keepaspectratio]{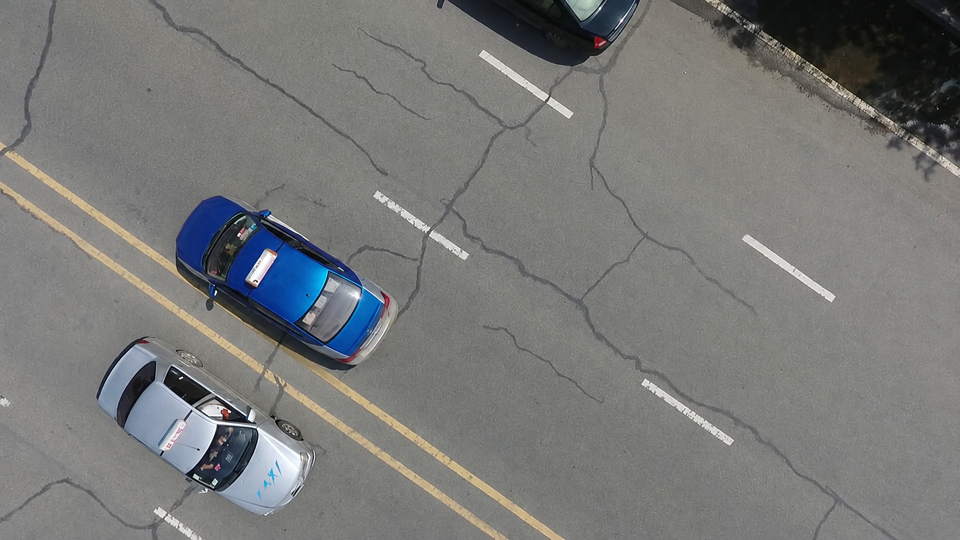}
\end{minipage}
\par\smallskip
\begin{minipage}{\linewidth}\small
\textbf{Question.} Which road region contains the center of the blue taxi with the white roof sign?\par\smallskip
\textbf{A.} The roadside shoulder outside the paved carriageways\par
\textbf{B.} The painted double-yellow centerline zone\par
\textbf{C.} The image-lower-left carriageway\par
\textbf{D.} The image-upper-right carriageway\par
\textbf{Ground truth:} O = \textbf{D}; CF = \textbf{B}.
\end{minipage}
\caption{Example of \textit{spatial position} and \textit{cultural and behavioral norms}.}

\label{fig:example_position}
\end{figure}

\begin{figure}[htbp]
\centering
\begin{minipage}[t]{0.49\linewidth}\centering
\textbf{Original (O)}\\[3pt]
\includegraphics[width=\linewidth,height=0.205\textheight,keepaspectratio]{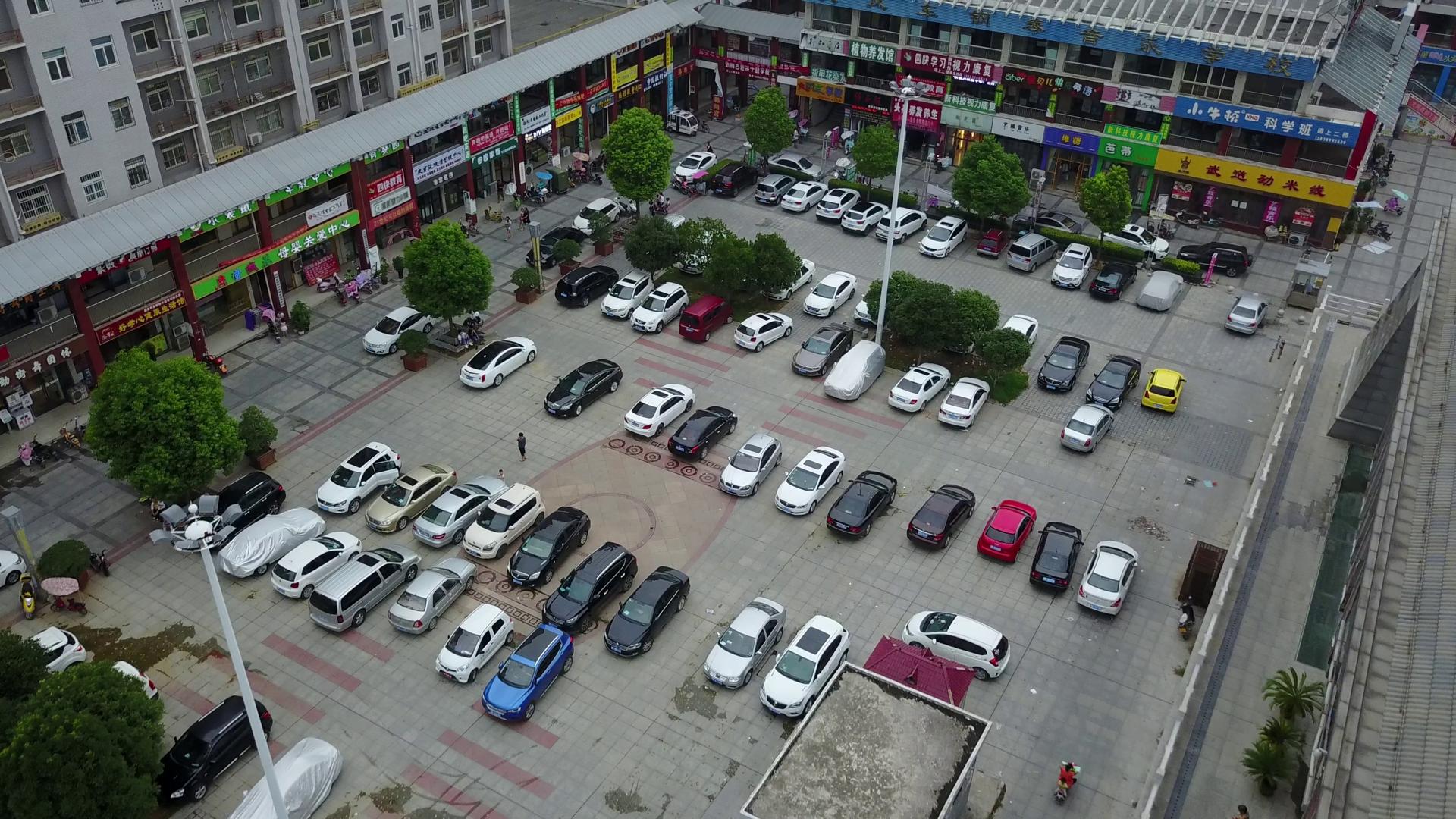}
\end{minipage}\hfill
\begin{minipage}[t]{0.49\linewidth}\centering
\textbf{Counterfactual (CF)}\\[3pt]
\includegraphics[width=\linewidth,height=0.205\textheight,keepaspectratio]{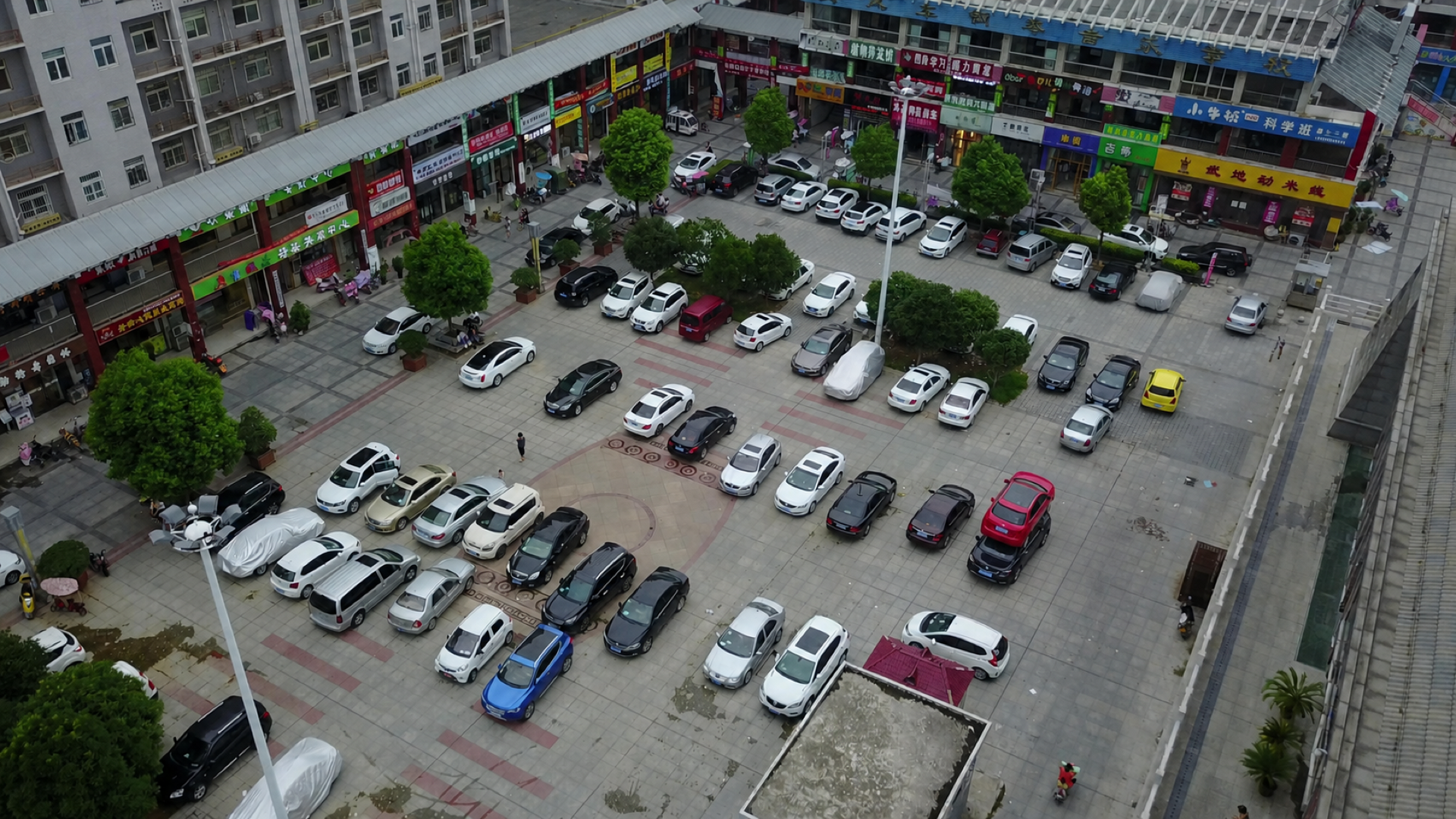}
\end{minipage}
\par\smallskip
\begin{minipage}{\linewidth}\small
\textbf{Question.} What is the physical spatial relationship between the bright red car in the lower-right diagonal row and the adjacent dark car immediately to its right?\par\smallskip
\textbf{A.} The red car is following behind the dark car with a road gap\par
\textbf{B.} One of the two cars is not visible enough to determine the relation\par
\textbf{C.} The red car is resting directly on top of the dark car\par
\textbf{D.} Their vehicle bodies are separate in neighboring diagonal positions\par
\textbf{Ground truth:} O = \textbf{D}; CF = \textbf{C}.
\end{minipage}
\caption{Example of \textit{spatial relationships} and \textit{physical-geometric principles}.}

\label{fig:example_relationship}
\end{figure}

\begin{figure}[t]
\centering
\begin{minipage}[t]{0.49\linewidth}\centering
\textbf{Original (O)}\\[3pt]
\includegraphics[width=\linewidth,height=0.205\textheight,keepaspectratio]{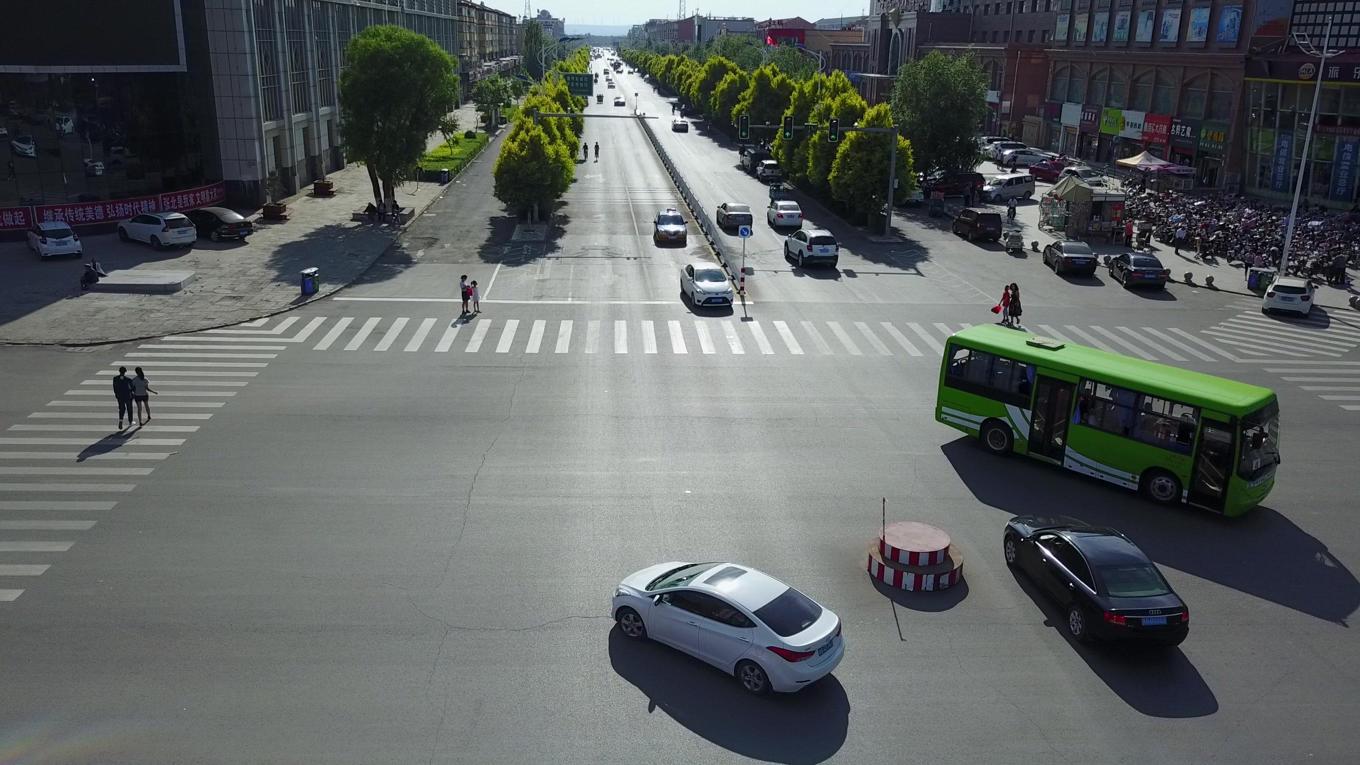}
\end{minipage}\hfill
\begin{minipage}[t]{0.49\linewidth}\centering
\textbf{Counterfactual (CF)}\\[3pt]
\includegraphics[width=\linewidth,height=0.205\textheight,keepaspectratio]{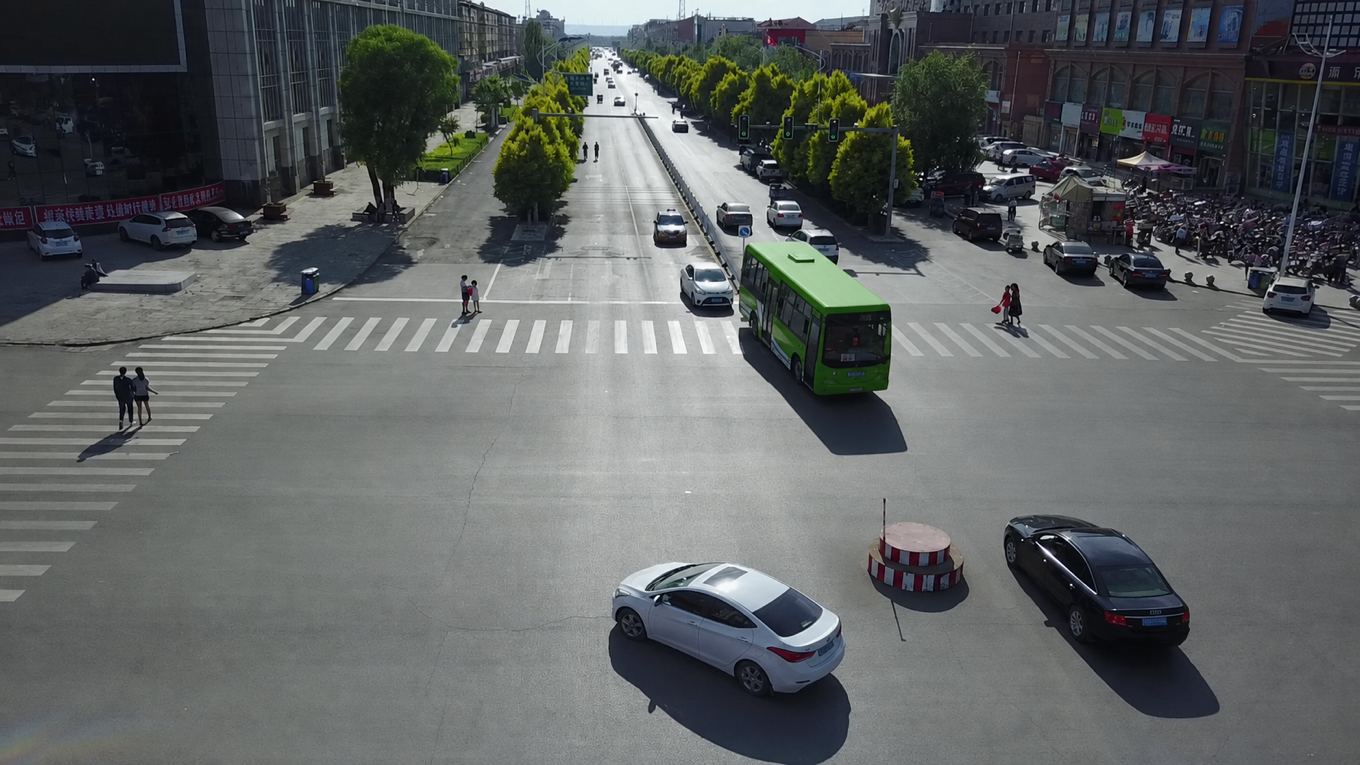}
\end{minipage}
\par\smallskip
\begin{minipage}{\linewidth}\small
\textbf{Question.} Which current maneuver geometry best describes the bright green city bus in the intersection?\par\smallskip
\textbf{A.} Crossing straight from image-left to image-right along the cross street\par
\textbf{B.} Making a U-turn within the horizontal cross street\par
\textbf{C.} Stopped parallel beside the central traffic island\par
\textbf{D.} Turning left from the cross street into the avenue toward the image top\par
\textbf{Ground truth:} O = \textbf{A}; CF = \textbf{D}.
\end{minipage}
\caption{Example of \textit{spatial action} and \textit{cultural and behavioral norms}.}

\label{fig:example_action}
\end{figure}

\begin{figure}[t]
\centering
\begin{minipage}[t]{0.49\linewidth}\centering
\textbf{Original (O)}\\[3pt]
\includegraphics[width=\linewidth,height=0.205\textheight,keepaspectratio]{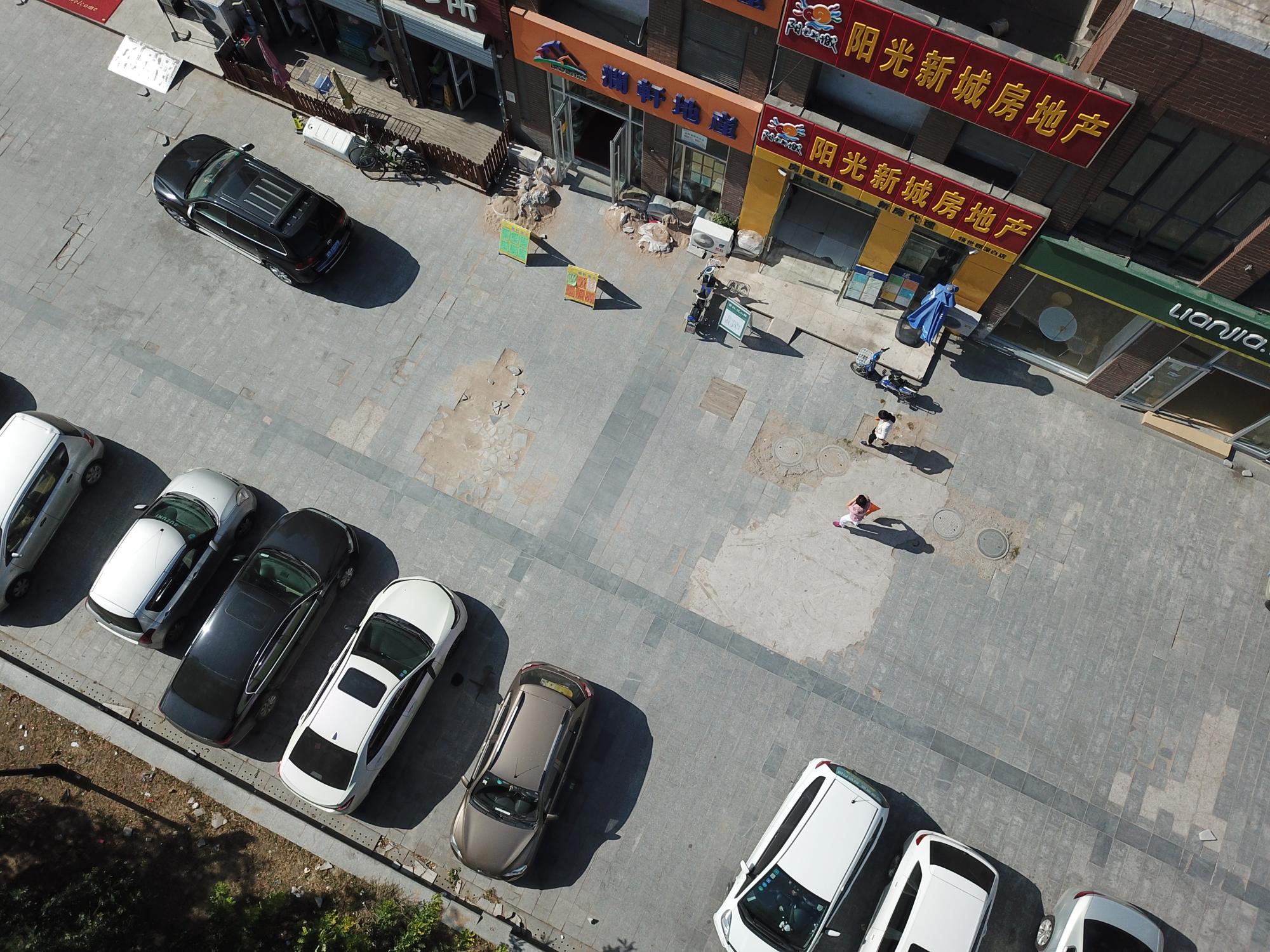}
\end{minipage}\hfill
\begin{minipage}[t]{0.49\linewidth}\centering
\textbf{Counterfactual (CF)}\\[3pt]
\includegraphics[width=\linewidth,height=0.205\textheight,keepaspectratio]{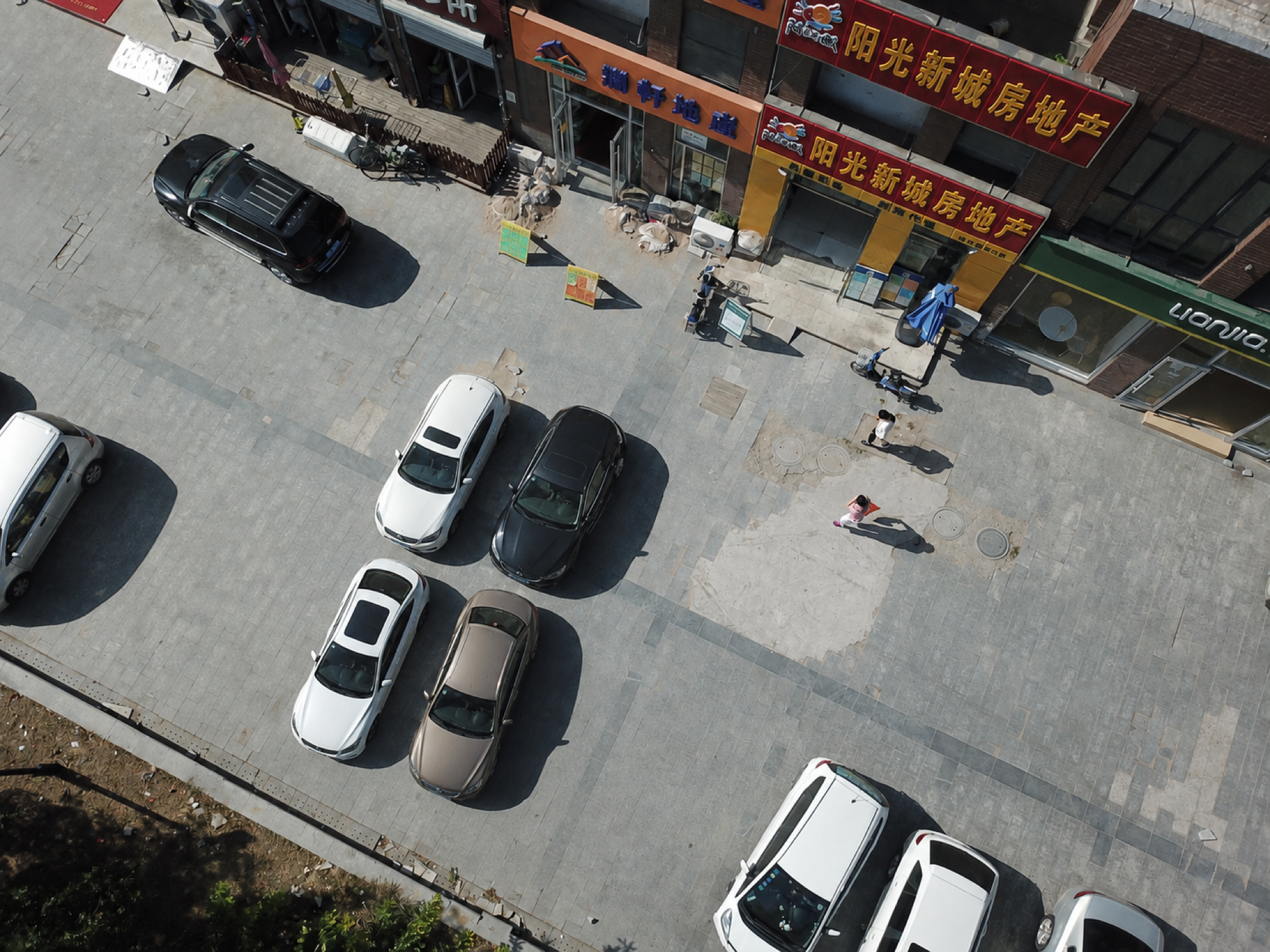}
\end{minipage}
\par\smallskip
\begin{minipage}{\linewidth}\small
\textbf{Question.} In the storefront forecourt, what overall configuration is formed by the selected white, dark, white, and bronze cars?\par\smallskip
\textbf{A.} An irregular scattered cluster\par
\textbf{B.} A two-by-two rectangular array\par
\textbf{C.} An alternating zigzag\par
\textbf{D.} One curb-parallel diagonal row\par
\textbf{Ground truth:} O = \textbf{D}; CF = \textbf{B}.
\end{minipage}
\caption{Example of \textit{geometric structure} and \textit{semantic co-occurrence}.}

\label{fig:example_structure}
\end{figure}

\begin{figure}[t]
\centering
\begin{minipage}[t]{0.49\linewidth}\centering
\textbf{Original (O)}\\[3pt]
\includegraphics[width=\linewidth,height=0.205\textheight,keepaspectratio]{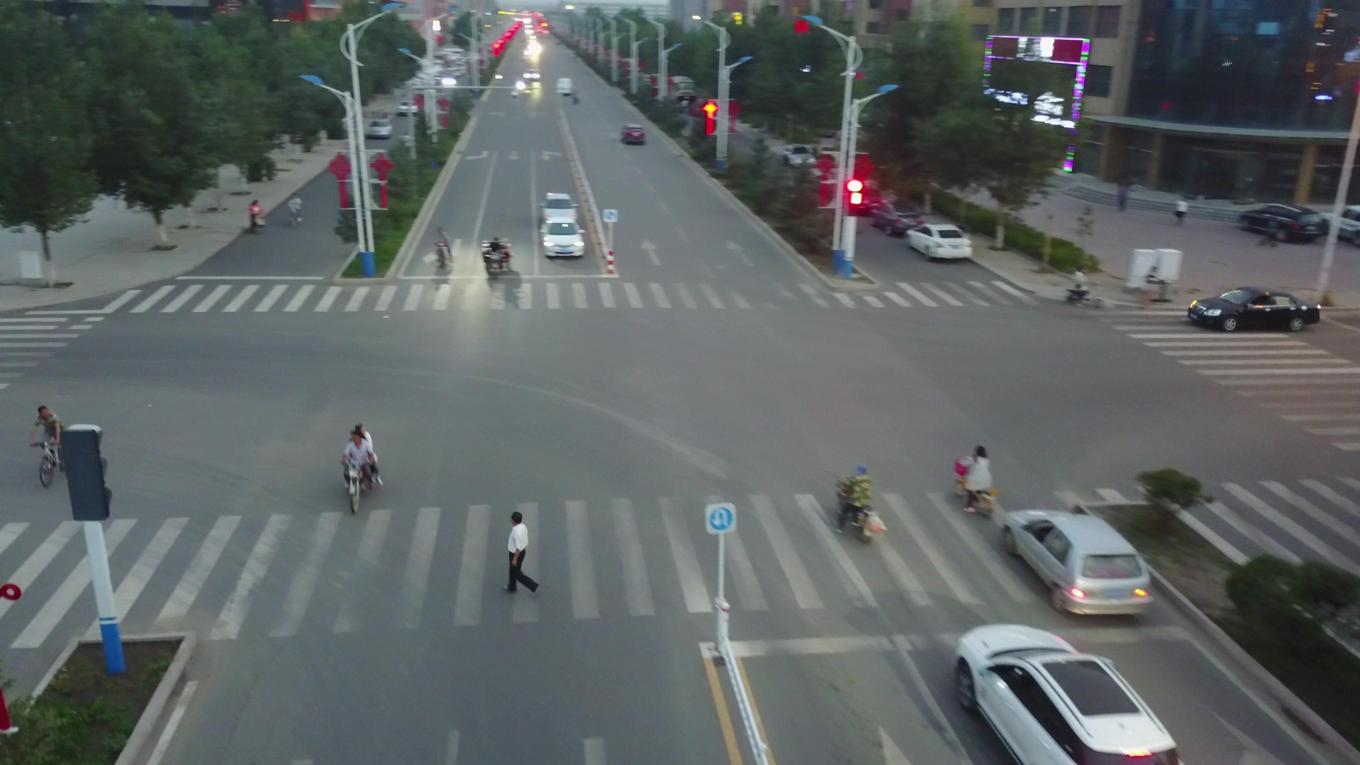}
\end{minipage}\hfill
\begin{minipage}[t]{0.49\linewidth}\centering
\textbf{Counterfactual (CF)}\\[3pt]
\includegraphics[width=\linewidth,height=0.205\textheight,keepaspectratio]{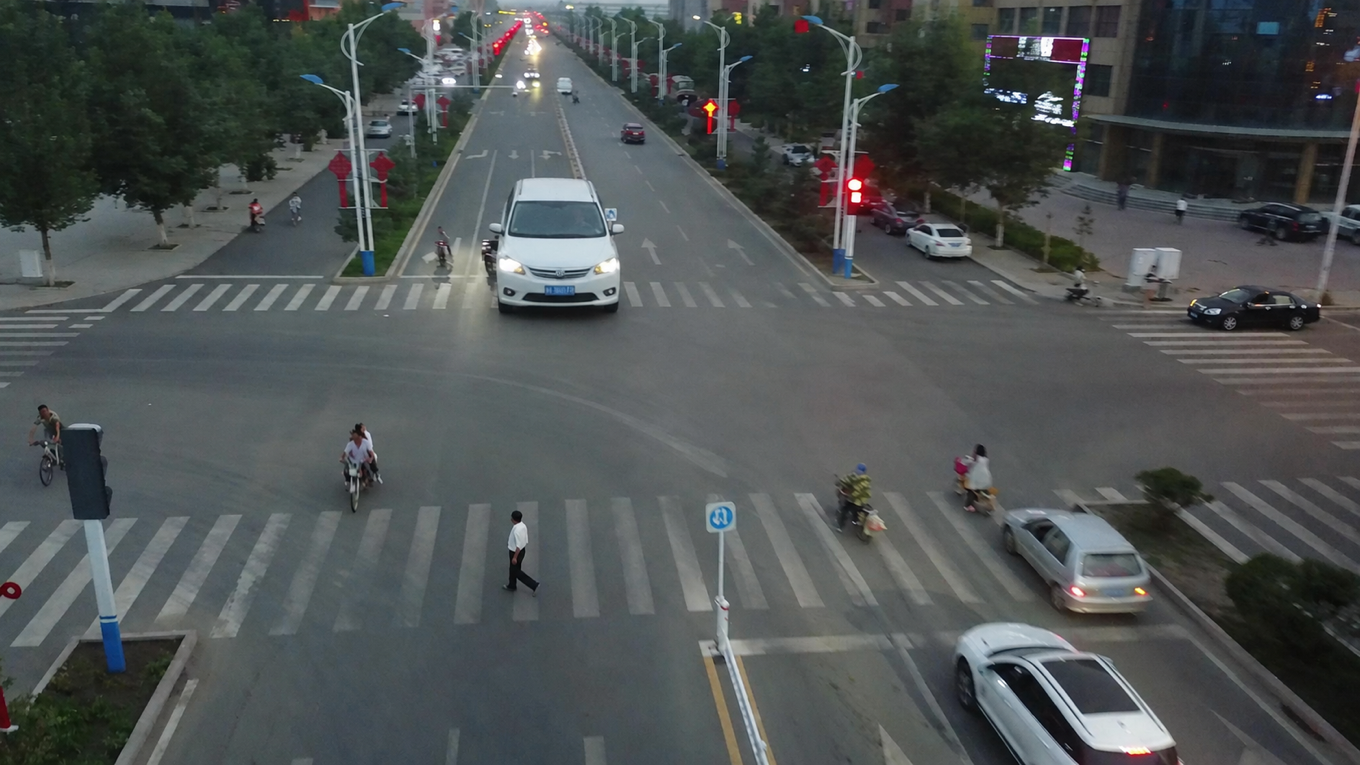}
\end{minipage}
\par\smallskip
\begin{minipage}{\linewidth}\small
\textbf{Question.} Using projected image footprint as the decision rule, how does the lowest white vehicle in the central inbound chain, immediately above the far zebra crossing compare with the large silver hatchback crossing at the lower right?\par\smallskip
\textbf{A.} At least one of the two objects is not visible enough to compare\par
\textbf{B.} Their projected footprints are approximately equal\par
\textbf{C.} The distant central white vehicle has a clearly larger projected footprint than the lower-right silver hatchback\par
\textbf{D.} The distant central white vehicle has a clearly smaller projected footprint than the lower-right silver hatchback\par
\textbf{Ground truth:} O = \textbf{D}; CF = \textbf{C}.
\end{minipage}
\caption{Example of \textit{depth and scale} and \textit{physical-geometric principles}.} 

\label{fig:example_scale}
\end{figure}

\begin{figure}[t]
\centering
\begin{minipage}[t]{0.49\linewidth}\centering
\textbf{Original (O)}\\[3pt]
\includegraphics[width=\linewidth,height=0.205\textheight,keepaspectratio]{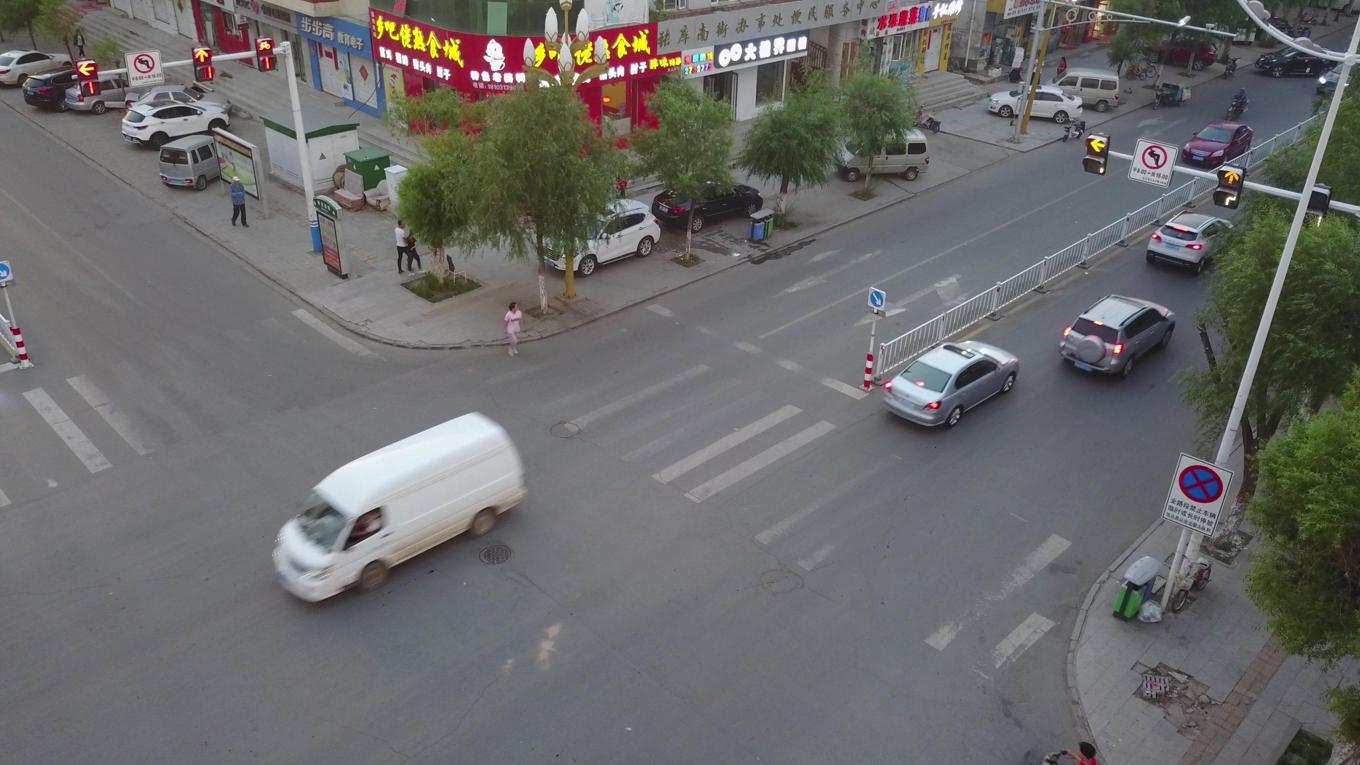}
\end{minipage}\hfill
\begin{minipage}[t]{0.49\linewidth}\centering
\textbf{Counterfactual (CF)}\\[3pt]
\includegraphics[width=\linewidth,height=0.205\textheight,keepaspectratio]{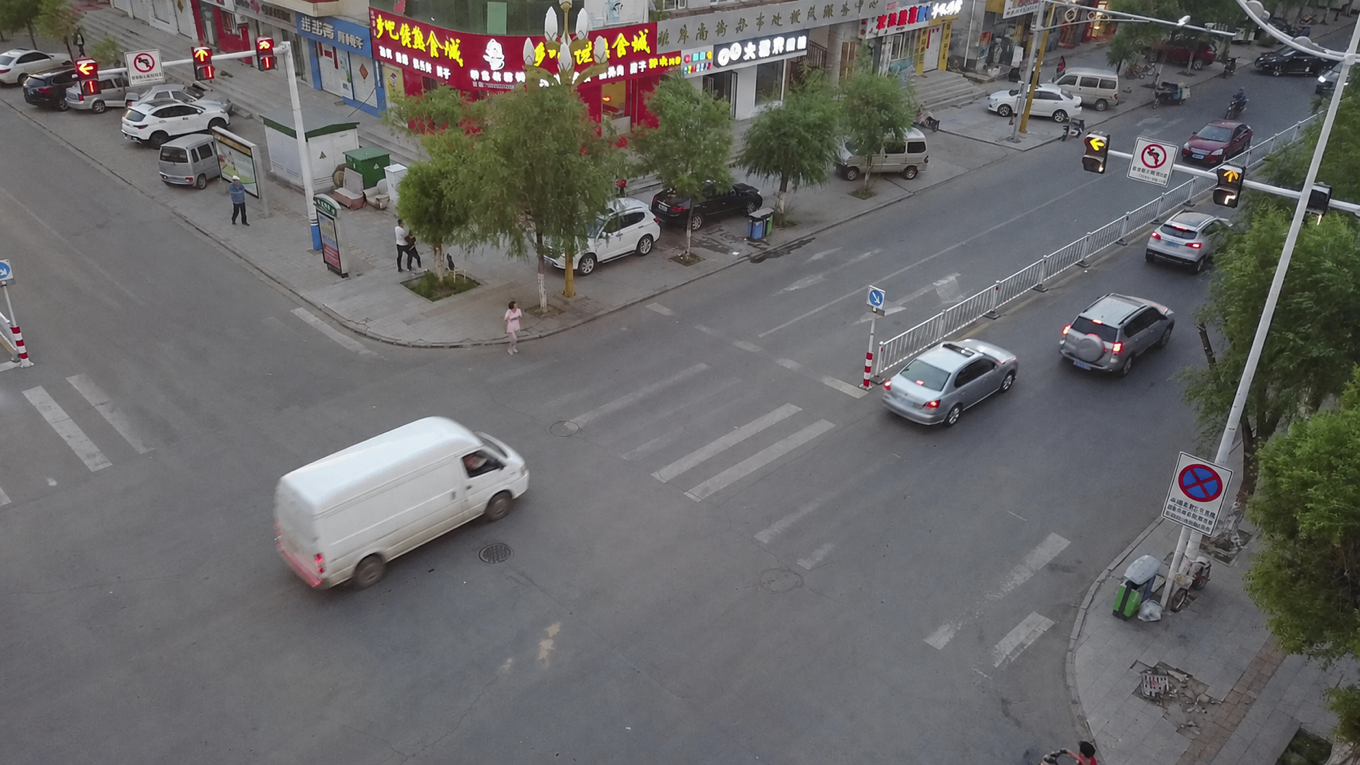}
\end{minipage}
\par\smallskip
\begin{minipage}{\linewidth}\small
\textbf{Question.} When the white panel van's physical forward axis is expressed in the fixed aerial camera frame, which diagonal image direction does it point toward?\par\smallskip
\textbf{A.} Toward the upper-right of the image\par
\textbf{B.} Toward the lower-right of the image\par
\textbf{C.} Toward the lower-left of the image\par
\textbf{D.} Toward the upper-left of the image\par
\textbf{Ground truth:} O = \textbf{C}; CF = \textbf{A}.
\end{minipage}
\caption{Example of \textit{viewpoint and orientation} and \textit{egocentric-allocentric transformation}.} 

\label{fig:example_orientation}
\end{figure}

This section presents several representative data samples from our benchmark. Fig.~\ref{fig:example_position}--\ref{fig:example_orientation} present six illustrative pairs, one for each spatial task and collectively covering all four bias-reason labels. The original and counterfactual images are displayed side by side for exposition, and the shared question can be answered from either image individually. 

The examples distinguish closely related spatial judgments. In Fig.~\ref{fig:example_position}, relocating a taxi changes its region membership, whereas Fig.~\ref{fig:example_relationship} changes the support/contact relation between two cars. Fig.~\ref{fig:example_action} changes a bus's visible maneuver geometry. Fig.~\ref{fig:example_structure} rearranges a group of cars, testing global configuration rather than the position of a single object. Fig.~\ref{fig:example_scale} explicitly compares projected footprints, avoiding an unsupported inference from pixel size to metric depth. Finally, Fig.~\ref{fig:example_orientation} requires expressing a van's physical forward axis in image coordinates. 

\end{document}